\documentclass[conference,a4paper]{IEEEtran} 
\ifCLASSINFOpdf
  \usepackage[pdftex]{graphicx}
  \DeclareGraphicsExtensions{.eps}
\else
\fi
\usepackage[cmex10]{amsmath}
\usepackage{amsfonts}

\newcommand{\argmax}{\operatornamewithlimits{argmax}}

\usepackage{caption}
\usepackage[font=footnotesize]{subfig}
\usepackage{adjustbox}

\begin{document}
%
\title{Diagnosing Convolutional Neural Networks using their Spectral Response}



\author{\IEEEauthorblockN{Victor Stamatescu}
\IEEEauthorblockA{Defence Science and Technology Group\\
Edinburgh, SA 5111, Australia\\
Email: victor.stamatescu@dst.defence.gov.au}
\and
\IEEEauthorblockN{Mark D. McDonnell}
\IEEEauthorblockA{Computational Learning Systems Laboratory (cls-lab.org)\\
School of Information Technology and Mathematical Sciences\\
University of South Australia\\
Mawson Lakes, SA 5095, Australia}}


%


\maketitle

\begin{abstract}
Convolutional Neural Networks (CNNs) are a class of artificial neural networks whose computational blocks use convolution, together with other linear and non-linear operations, to perform classification or regression.
This paper explores the spectral response of CNNs
and its potential use in diagnosing problems with their training.
We measure the \emph{gain} of CNNs trained
for image classification on ImageNet and observe that the best models are also the most sensitive to perturbations of their input. Further, we perform experiments on MNIST and CIFAR-10 to find that the gain rises as the network learns and then saturates 
as the network converges. Moreover, we find that strong gain fluctuations can point to overfitting and learning problems caused by a poor choice of learning rate. We argue that the gain of CNNs can act as a diagnostic tool and potential replacement for the validation loss when hold-out validation data are not available.
\end{abstract}


%
\IEEEpeerreviewmaketitle

\section{Introduction}

Convolutional Neural Networks (CNNs) are computational models made of multiple processing layers
whose parameters are learned from data.
CNNs have been applied to a diverse range of tasks
and continue to set new benchmarks in areas of computer vision, speech recognition and natural language processing~\cite{lecun2015deep}. Despite their widespread use, however,
training on new data sets or with new models remains a complex process.
This is in large part due to the wide search space of model architecture and hyper-parameters~\cite{zoph2017learning}.
Poor choices can cause problems during training,
including high bias (underfitting) or high variance (overfitting).
If left undetected, such issues can severely degrade CNN performance at test time.

In this work our goal is to measure the response of CNNs trained for image classification to different input spatial frequencies. We then use this information as a way to probe and better understand the learning process. Given that some CNN layers include non-linearities, obtaining a transfer function for the entire network requires linearizing it first. We are, to our knowledge, the first to pursue this \emph{linearization} approach, which is analogous to the way in which linear systems are characterized using their transfer functions.

The rest of this paper is organized as follows. Section II outlines key papers related to the spectral properties of CNNs. Section III describes the methodology, including how the spectral response was measured in a number of pre-trained CNNs. Section IV presents our experimental results, which demonstrate how spectral response can serve as a diagnostic tool during training. Finally, Section V concludes with a summary of our findings.

\section{Related Work}

Heravi \emph{et al.}~\cite{heravi2015analyzing} presented 4D visualizations of the frequency response in convolutional filter banks learned by CNNs. Based on this they proposed that the predictions output by CNNs may be sensitive to low amplitude additive noise due to the inability of the filters to suppress high spatial frequencies. Unlike their study, ours is focused on capturing the frequency response of the entire CNN, not just its linear filters.

Recent studies~\cite{rahaman2018spectral,xu2018training,xu2018understanding} have found that neural networks tend to fit low frequency components of input data first during training, while the higher frequencies are modeled more slowly at later stages. This phenomenon was investigated in terms of the regression of target functions in the Fourier domain and is used to explain how neural networks that are capable of overfitting can still generalize well on test data.

A number of other publications relating to CNN spectral properties have focused on improving processing speed and performance through new types of computational units and operators. These include \emph{spectral pooling}~\cite{rippel2015spectral}, concatenated \emph{relu}~\cite{shang2016understanding}, spectral features from convolutional layer activations~\cite{khan2017scene},
and wavelet CNNs~\cite{fujieda2018wavelet}. In particular, the pooling technique in~\cite{rippel2015spectral} performs dimensionality reduction via truncation in the frequency domain, which preserves more information than spatial pooling.

Unlike those studies our focus is on measuring CNN spectral response to a specific test image.
This is similar in some aspects to the approach taken in a number of CNN visualization methods~\cite{simonyan2013deep,zeiler2014visualizing,shrikumar2017learning,zintgraf2017visualizing},
which illustrate why a network has made its prediction for a particular input image.
These methods provide 2D visualizations the input saliency or of features in higher layers.
Our goal, however, is to summarize the CNN spectral response using a single metric,
which can then can be used in a similar way to the training and validation losses
to identify problems during training.

\section{The CNN Transfer Function}

\subsection{Definition}

An artificial neural network is a non-linear function $f$
with parameters $\mathbf{w}$ that are learned during training,
which maps an input tensor $\mathbf{x}$ to an output tensor $\mathbf{z} = f(\mathbf{x};\mathbf{w})$.
This feed-forward operation is known as as a \emph{forward pass} or \emph{inference}.
In the case of CNNs, $f$ is typically composed of multiple computational layers:
\begin{equation}
\begin{aligned}
\mathbf{z} &= f(\mathbf{x};\mathbf{w}_1,\ldots ,\mathbf{w}_L)\\
           &= f_L(f_{L-1}(\ldots f_1(\mathbf{x}_0;\mathbf{w}_1)\ldots;\mathbf{w}_{L-1});\mathbf{w}_L),
\end{aligned}
\end{equation}
where $\mathbf{x}_0=\mathbf{x}$ is the input and each layer $l=1,\ldots,L$ computes an intermediate output $\mathbf{x}_l=f_l(\mathbf{x}_{l-1};\mathbf{w}_l)$. Depending on the task, $\mathbf{z}$ may have the dimensions of a multidimensional tensor, a 2D matrix, a 1D vector or a scalar.

Our aim is to define a transfer function for a CNN.
For a linear system, such as a low-pass filter, the transfer function is found by taking the Fourier transform of its coefficients or, alternatively, by measuring its gain and phase using sinusoidal input signals of different frequencies. CNNs, however, are highly non-linear, so the gain would depend not only on the spatial frequency of the input but also on its amplitude. We therefore need to first \emph{linearize} the network about a certain operating point. Fortunately, linear approximations can be obtained by performing a \emph{backward pass} through the CNN using backpropagation~\cite{simonyan2013deep}.

During training, a \emph{loss} function is added as the final layer
and the output becomes a scalar $z$.
Backpropagation applies the chain rule to obtain its gradient
with respect to the parameters in each layer:
\begin{multline}
\frac{dz}{d(\mathbf{w}_l)^\top} 
= 
\frac{d f_L(\mathbf{x}_{L-1};\mathbf{w}_{L})}{d(\mathbf{x}_{L-1})^\top}
\times
\dots \\
\times
\frac{d f_{l+1}(\mathbf{x}_{l};\mathbf{w}_{l+1})}{d(\mathbf{x}_{l})^\top}
\times
\frac{d f_l(\mathbf{x}_{l-1};\mathbf{w}_{l})}{d(\mathbf{w}_l)^\top},
\label{eqn:chain}
\end{multline}
where each intermediate derivative is evaluated at an operating point set by $\mathbf{x}_0$
and uses the current parameter values.
The resulting \emph{parameter derivatives} are used
by gradient descent optimization algorithms to carry out parameter updates.
In Eqn.~\eqref{eqn:chain} we followed the convention used in~\cite{vedaldi2015matconvnet},
where data and parameter tensors are stacked into vectors.

Backpropagation in Eqn.~\eqref{eqn:chain} works by passing back each layer's \emph{data derivative},
which is a large Jacobian matrix, to the layer $l$ below.
A memory efficient implementation that avoids
direct handling of the Jacobian involves passing back a projected derivative instead~\cite{vedaldi4564matconvnet}:
\begin{equation}
\mathbf{p}_{l} = 
\frac{d\langle\mathbf{p}_{l+1},f_{l+1}(\mathbf{x}_l;\mathbf{w}_{l+1})\rangle}{d\mathbf{x}_{l}}.
\label{eqn:projection}
\end{equation}
Here the inner product $\langle . \, , .\rangle$ projects the layer
along direction  $\mathbf{p}_{l+1}$, which produces a scalar
whose derivative is then evaluated.
When this process is repeated sequentially starting from the output layer,
$\mathbf{p}_l$ becomes the projected derivative of the network down to this layer.

Considering now the entire network, for a given projection tensor $\mathbf{p}$
and output $\mathbf{z} = f(\mathbf{x};\mathbf{w})$,
a single backward pass provides the projected data derivative $d\langle \mathbf{p}, f(\mathbf{x};\mathbf{w}) \rangle / d \mathbf{x}$
of the network output with respect to its input.
This was interpreted by Simonyan \emph{et al.}~\cite{simonyan2013deep} as the derivative in a first order Taylor approximation for the CNN, evaluated in the neighborhood given by the forward pass over an input $\mathbf{x}_0$:
\begin{equation}
 \frac{\mathbf{dz}}{\mathbf{dx}} =  \left .  \frac{d\langle \mathbf{p}, f(\mathbf{x};\mathbf{w}) \rangle}{ d \mathbf{x}} \right|_{x_0}.
 \label{eqn:taylor}
\end{equation}
Following this interpretation, we treat $\mathbf{dz}/\mathbf{dx}$ as a linear system and calculate its \emph{frequency response} by taking the Fourier transform $\mathcal{F}$ and computing the amplitude spectrum (in dB):
\begin{equation}
 \left|H\right| = 20 \times {\rm log}_{10}(\left| \mathcal{F}\left(\frac{\mathbf{dz}}{\mathbf{dx}}\right) \right| ),
 \label{eqn:fft}
\end{equation}
where $H$ is the transfer function.
\subsection{Toy CNN example}

\begin{figure}[ht]
\centering
\subfloat[Color input image]{
\includegraphics[trim={1.cm 1cm 1.cm 1.1cm},clip,width=0.35\columnwidth]{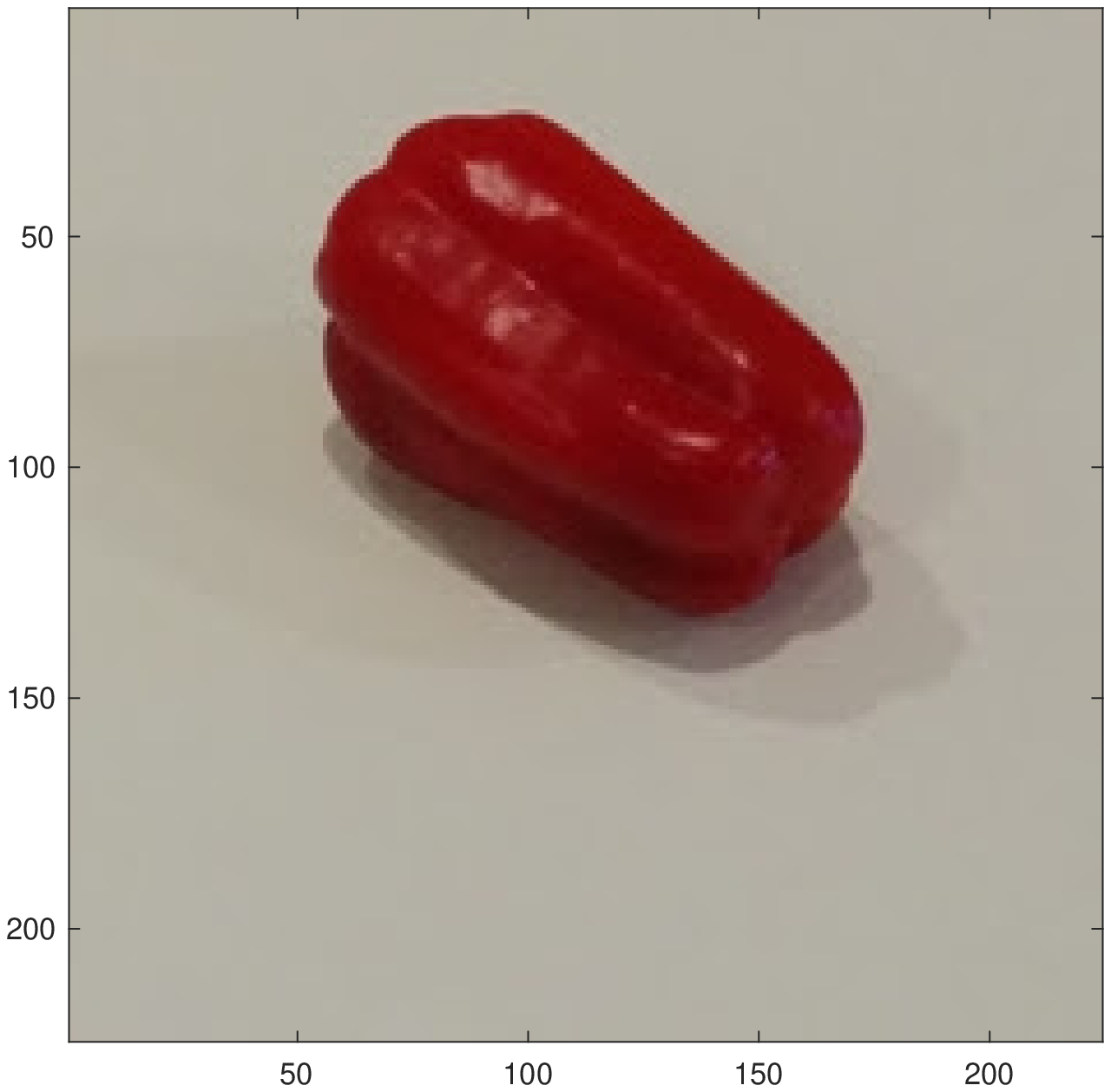}
\label{toy_rgb_input}}
\subfloat[Gaussian filter $\mathbf{w}$]{
\includegraphics[trim={1.cm 0.cm 1.cm 1.1cm},clip,width=0.32\columnwidth]{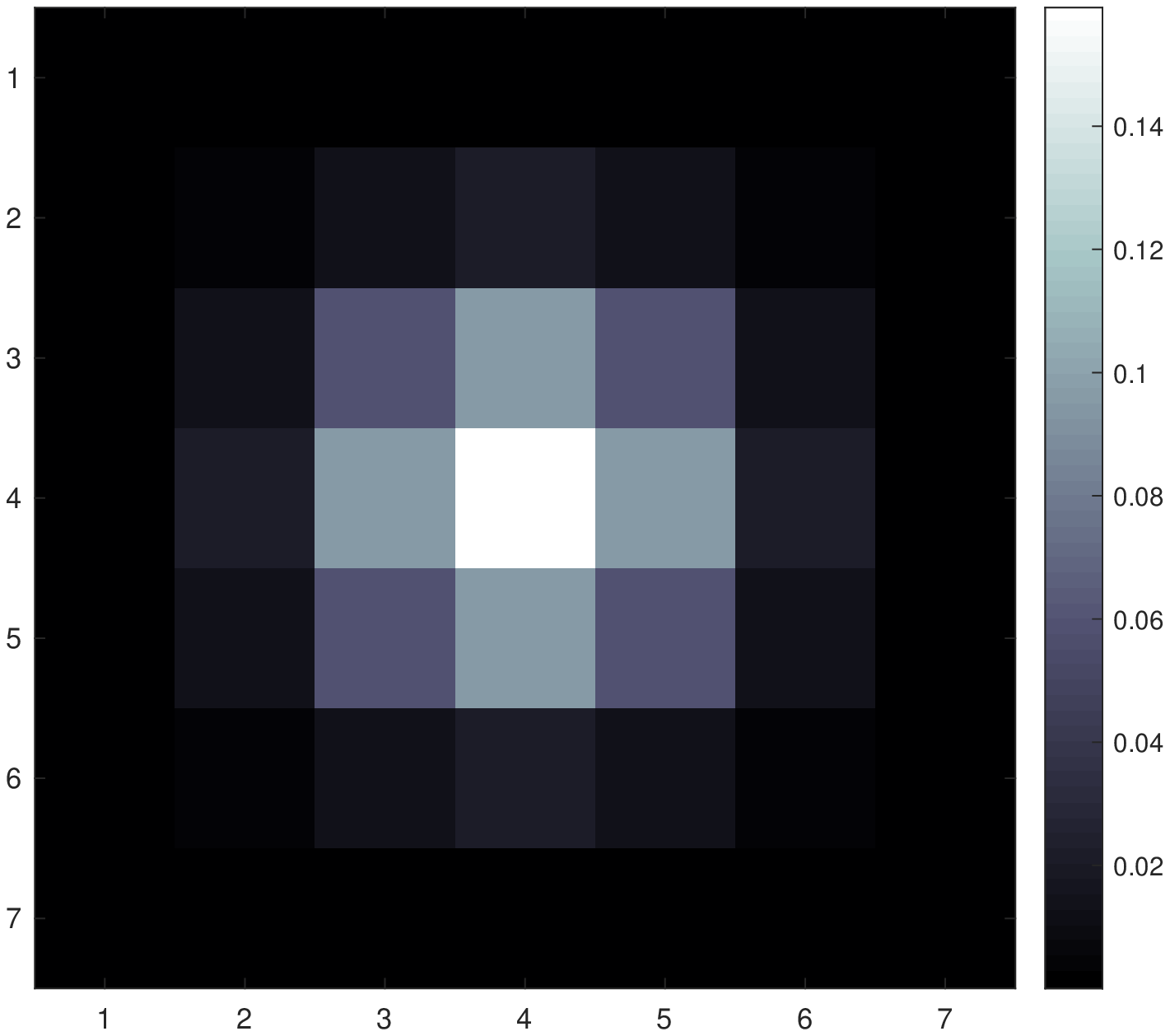}
\label{toy_red_filter}}
\caption{Color image whose red channel is input to a toy CNN consisting of a \emph{conv} layer, which applies a Gaussian filter $\mathbf{w}$, followed by a \emph{relu} layer.}
\end{figure}

\begin{figure}[ht]
\centering
\subfloat[Input $\mathbf{x}$]{
\includegraphics[trim={2.5cm 1cm 2.5cm 1.1cm},clip,width=0.32\columnwidth]{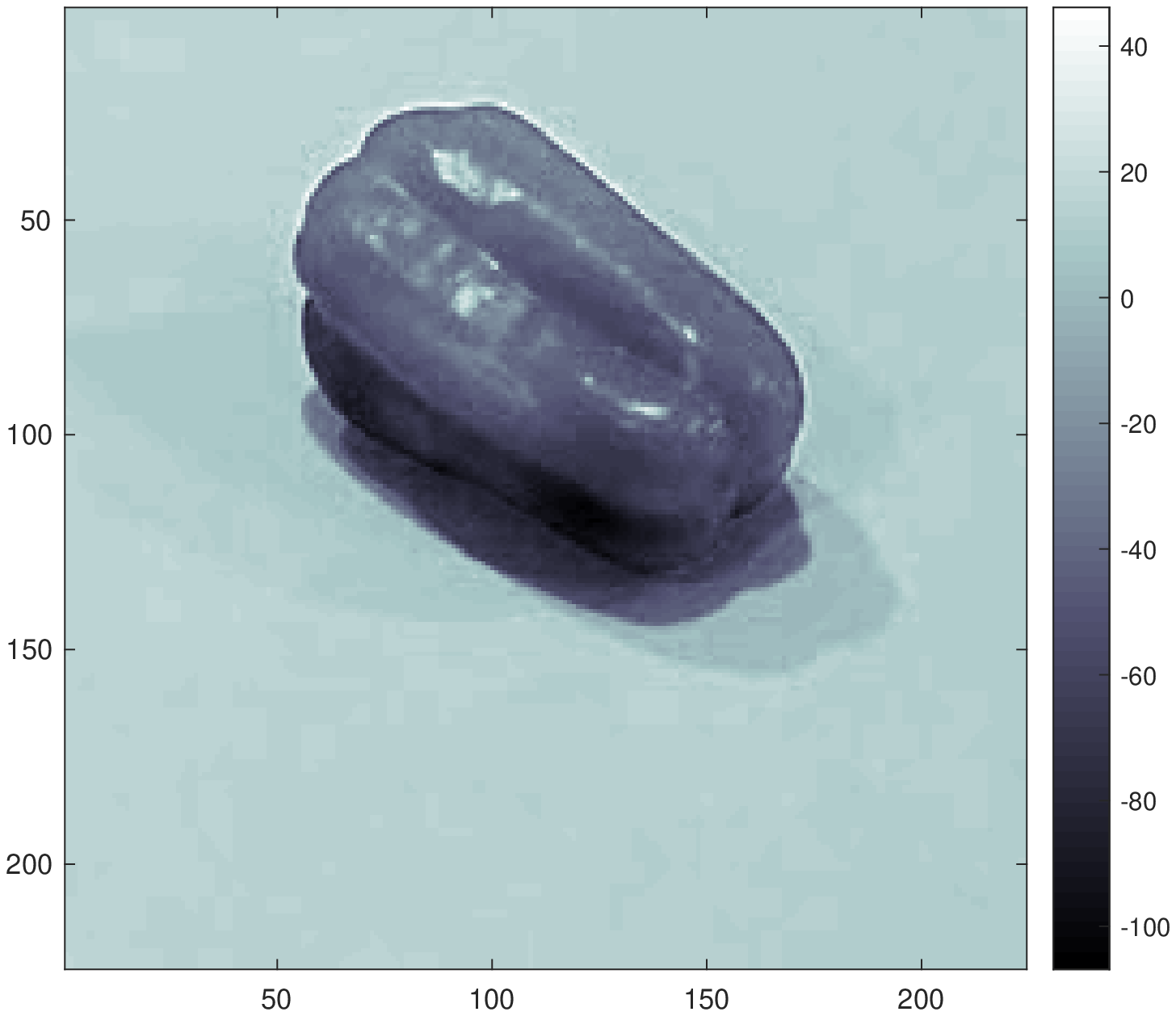}
\label{toy_red_meansub_input}}
\subfloat[\emph{conv} feature map]{
\includegraphics[trim={2.5cm 1cm 2.5cm 1.1cm},clip,width=0.32\columnwidth]{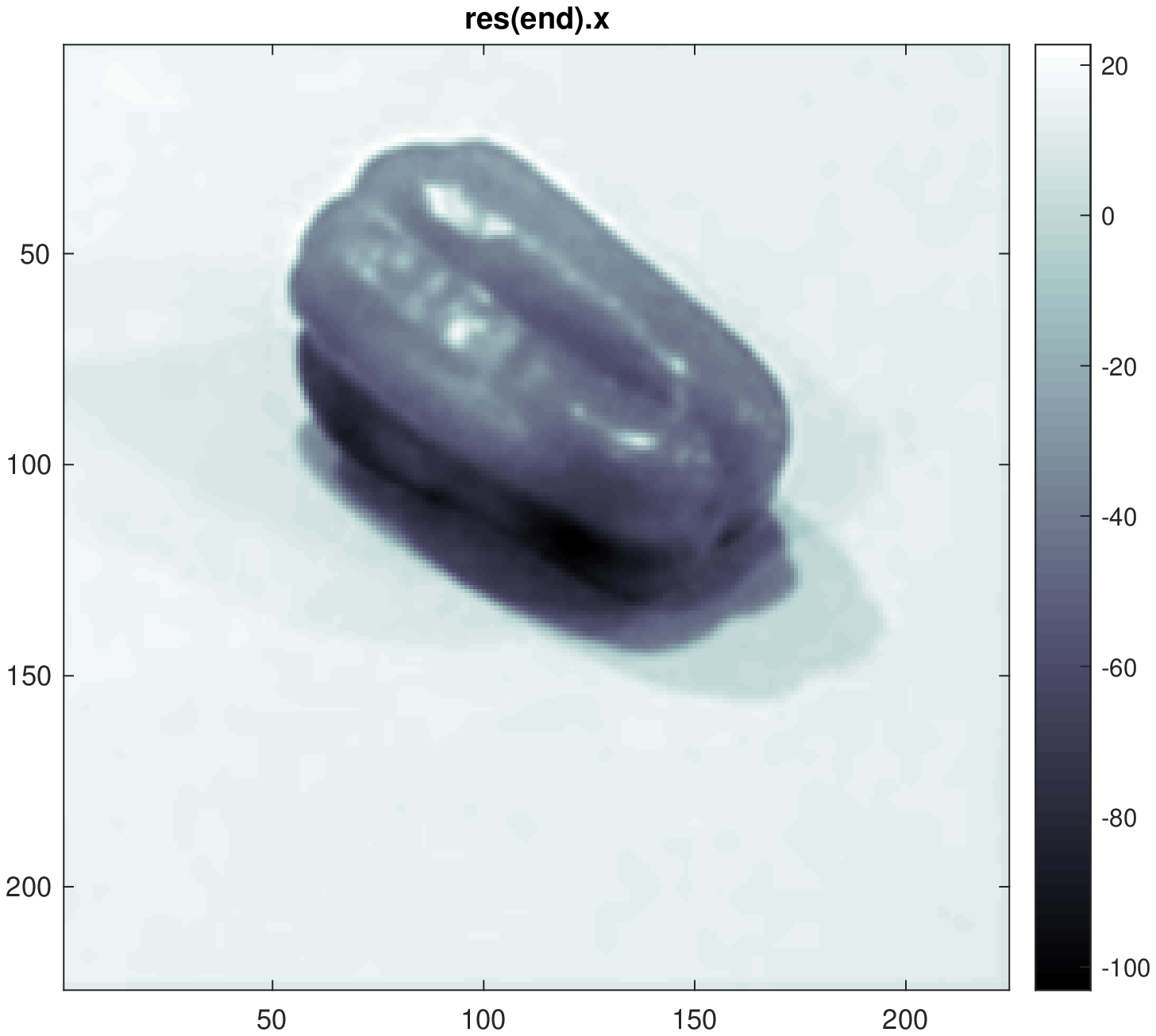}
\label{toy_red_conv_output}}
\subfloat[Output $\mathbf{z}$]{
\includegraphics[trim={2.5cm 1cm 2.5cm 1.1cm},clip,width=0.32\columnwidth]{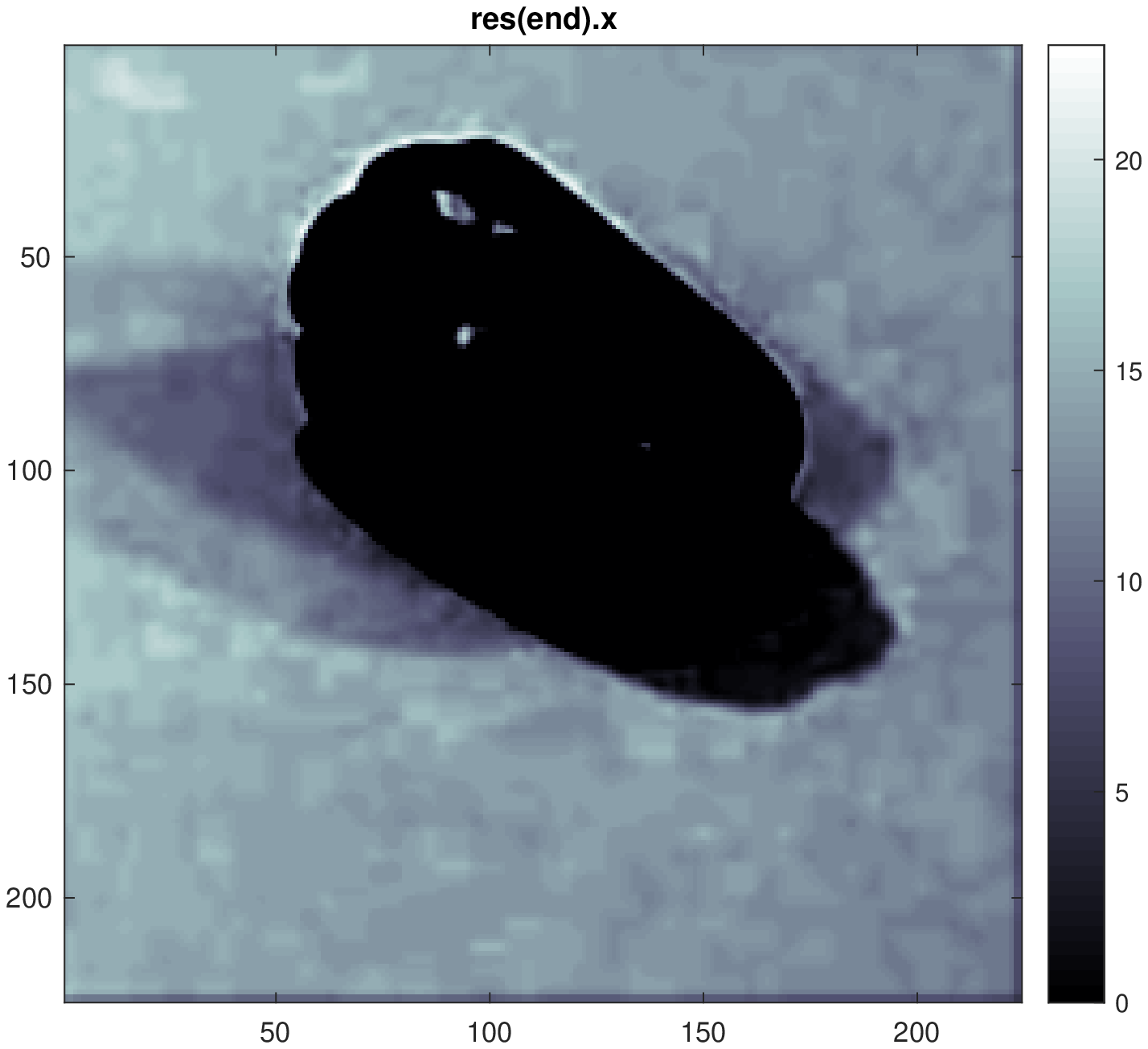}
\label{toy_red_relu_output}}
\caption{Toy CNN forward pass over the red input channel.}
\label{forward_pass}
\end{figure}

We constructed a two-layer toy CNN to illustrate the above discussion,
applying it to the mean-subtracted red channel
of the image in Figure~\ref{toy_rgb_input}.
The model consists of a linear \emph{conv} layer, which convolves the input (Figure~\ref{toy_red_meansub_input}) with a 2D filter $\mathbf{w}$,
followed by a non-linear \emph{relu} layer, which produces an output map $\mathbf{z}$ (Figure~\ref{toy_red_relu_output}) by applying a rectified linear unit
to each pixel in the feature map produced in the previous layer (Figure~\ref{toy_red_conv_output}).
The filter $\mathbf{w}$ is a symmetric single-channel $7\times7$ pixel Gaussian filter with $\sigma=1$ pixels, shown in Figure~\ref{toy_red_filter}.

\begin{figure}[ht]
\centering
\subfloat[Set $\mathbf{p}=1$ at $(150,100)$ and $\mathbf{p}=0$ elsewhere]{
\begin{tabular}[b]{c}%
\includegraphics[trim={2.5cm 1cm 2.5cm 1.1cm},clip,width=0.32\columnwidth]{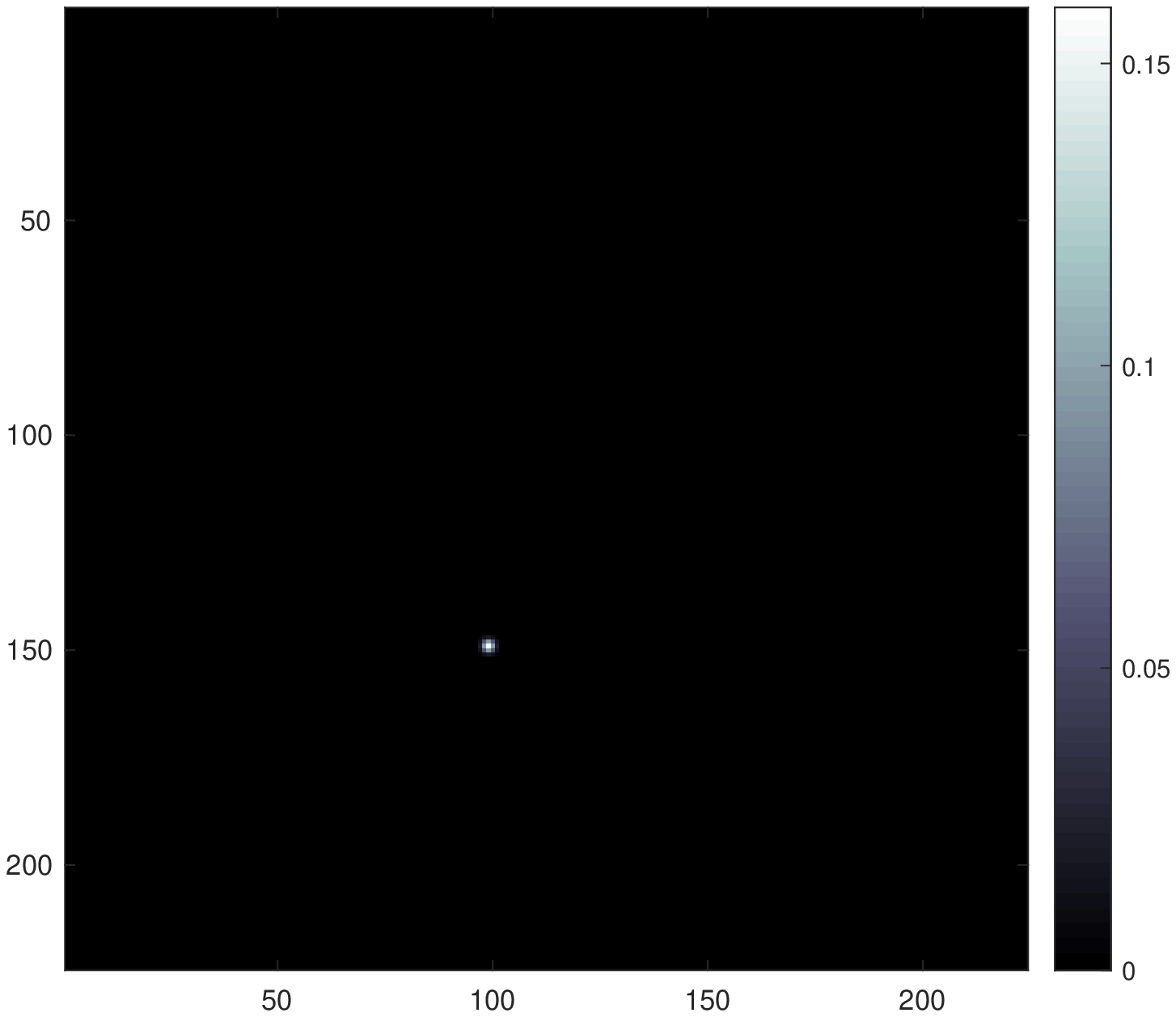}\hspace{0.025\textwidth}
\includegraphics[trim={2.5cm 1cm 2.5cm 1.1cm},clip,width=0.32\columnwidth]{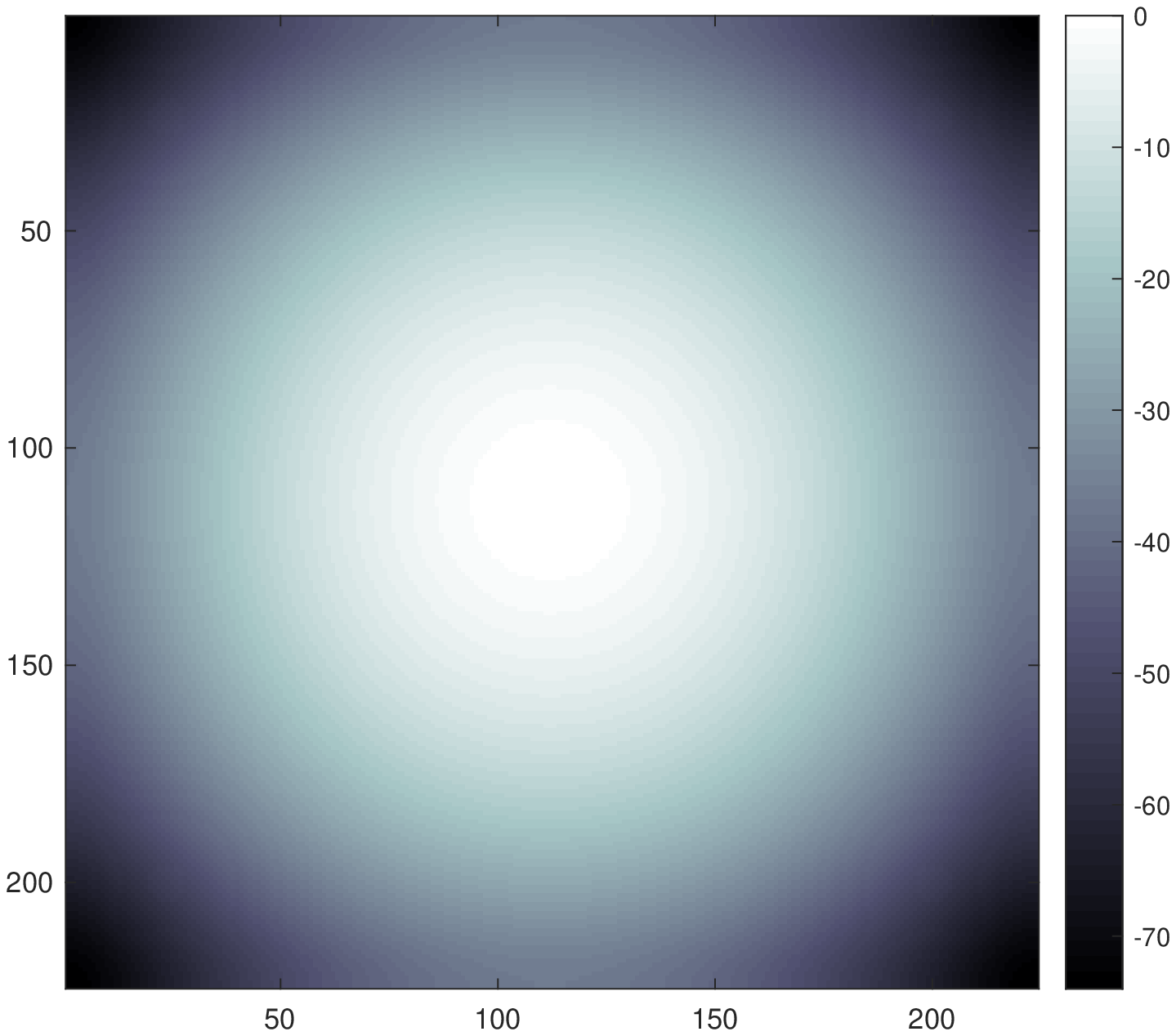}
\end{tabular}
\label{p1_150_100}
}\\
\subfloat[Set $\mathbf{p}=1$ everywhere]{
\begin{tabular}[b]{c}%
\includegraphics[trim={2.5cm 1cm 2.5cm 1.1cm},clip,width=0.32\columnwidth]{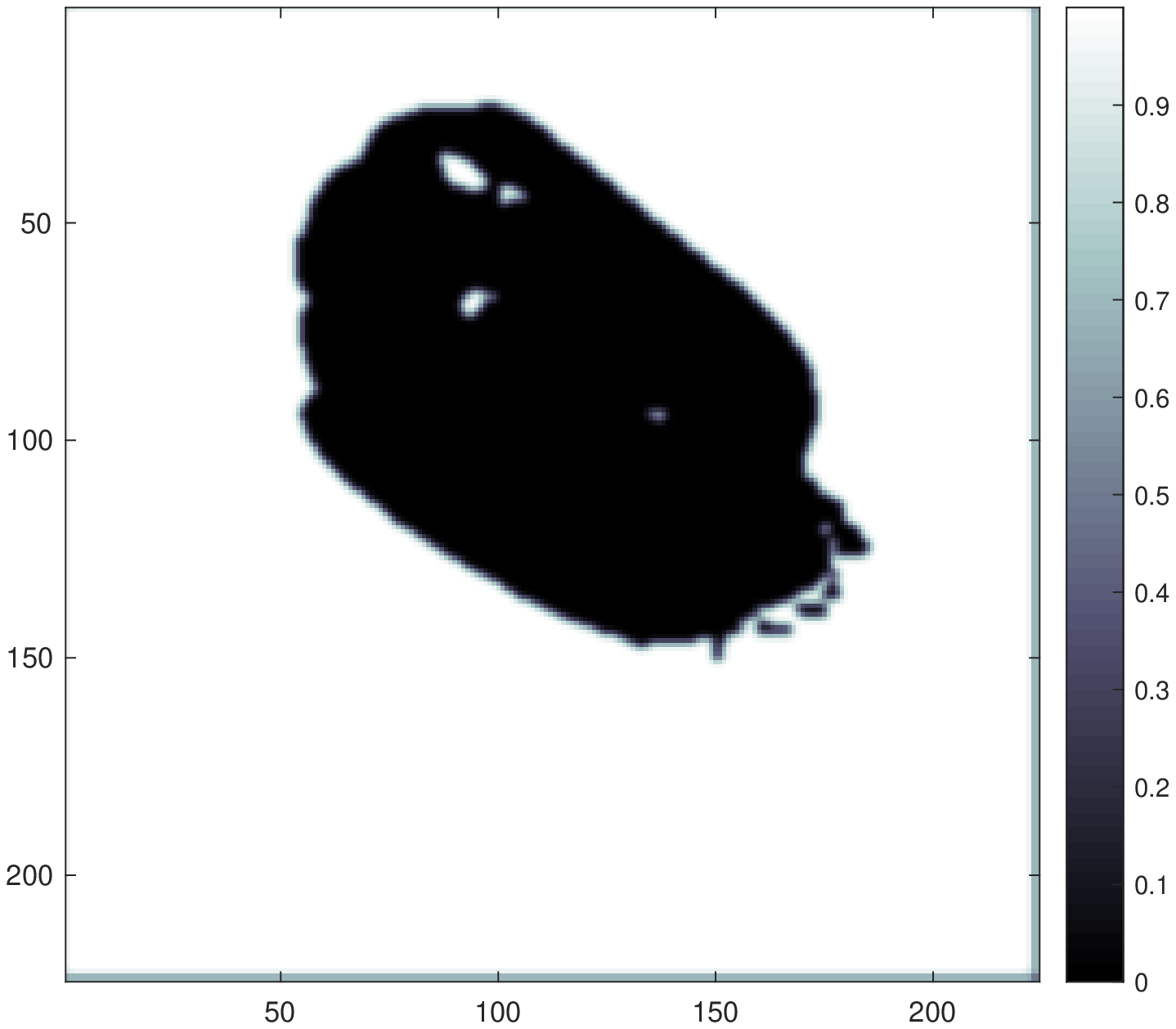}\hspace{0.025\textwidth}
\includegraphics[trim={2.5cm 1cm 2.5cm 1.1cm},clip,width=0.32\columnwidth]{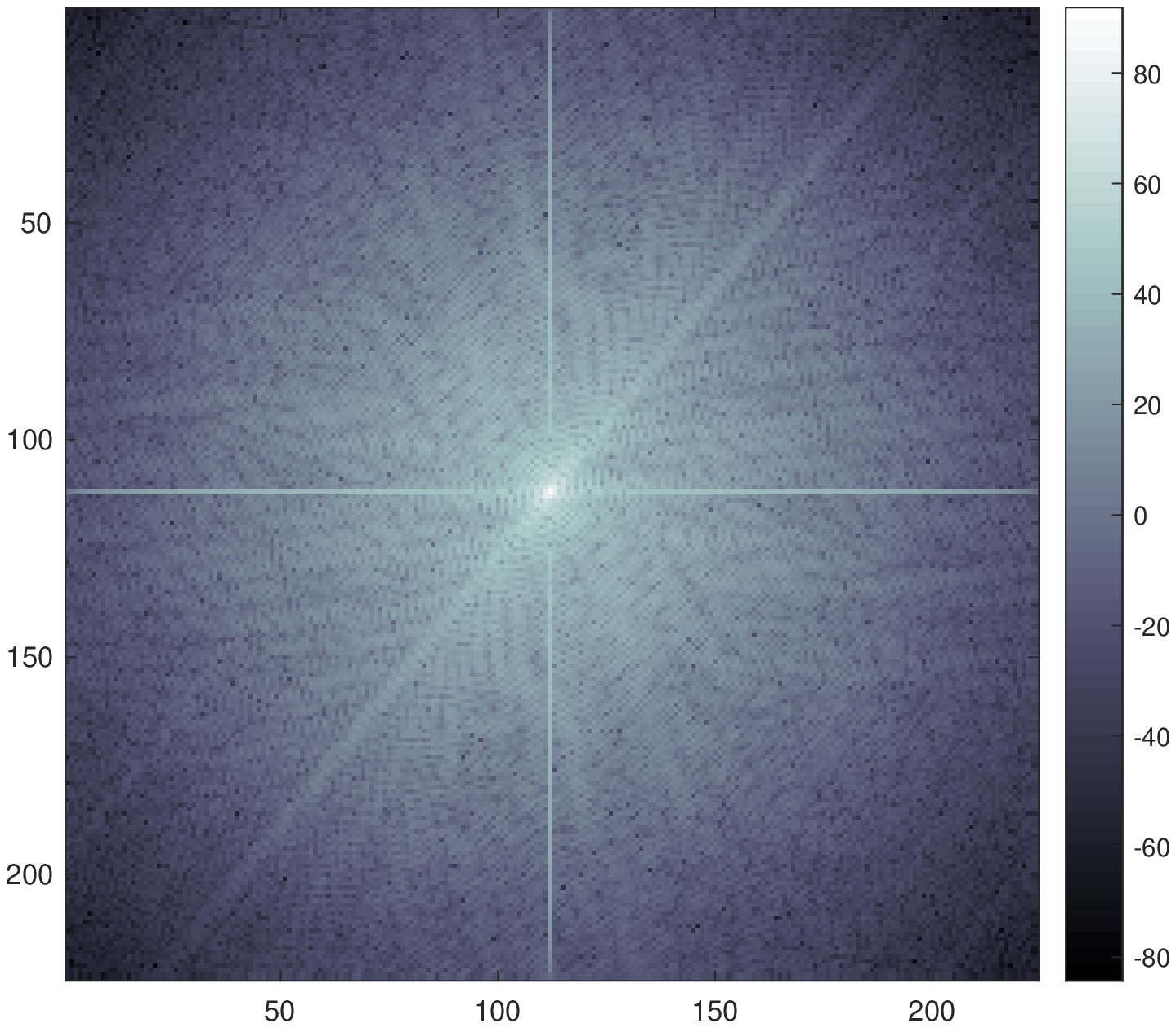}
\label{p1_everywhere}
\end{tabular}
}
\caption{Data derivative $d\mathbf{z}/d\mathbf{x}$ (\emph{left}) obtained from the backward pass and the corresponding frequency response (\emph{right}, in dB) for two choices of projection tensor $\mathbf{p}$.}
\label{backward_pass}
\end{figure}

Following the forward pass, a backward pass was used to compute the data derivative $d\mathbf{z}/d\mathbf{x}$. As shown in Figure~\ref{backward_pass}, this depends on the choice of 2D projection tensor $\mathbf{p}$,  which has the dimensions of $\mathbf{z}$. In Figure~\ref{p1_150_100} we used $\mathbf{p}$ set to $0$ everywhere except $(150,100)$, where it was 1.
These coordinates correspond to a non-zero output unit in $\mathbf{z}$ and any such location would yield the same frequency response, which, barring edge effects, is shift-invariant. Conversely, setting $\mathbf{p}$ to $1$ at a location where $\mathbf{z}$ is $0$ would yield $0$ everywhere in $d\mathbf{z}/d\mathbf{x}$. As detailed in~\cite{shrikumar2017learning}, this effect is due to the \emph{relu} non-linearity and is further illustrated by Figure~\ref{p1_everywhere}, where $\mathbf{p}$ was set to $1$ everywhere.

The data gradient in Figure~\ref{p1_150_100}
is simply the flipped impulse response of the symmetric Gaussian filter
at a particular location in the input image.
Hence the toy CNN frequency response given input $\mathbf{x}$ and and our choice of $\mathbf{p}$
is the amplitude of the Fourier transform pair to a low-pass filter,
which is the linear component of the CNN.

\begin{figure}[ht]
\centering
\subfloat[Input $\mathbf{x}$]{
\includegraphics[trim={2.5cm 1cm 2.5cm 1.1cm},clip,width=0.32\columnwidth]{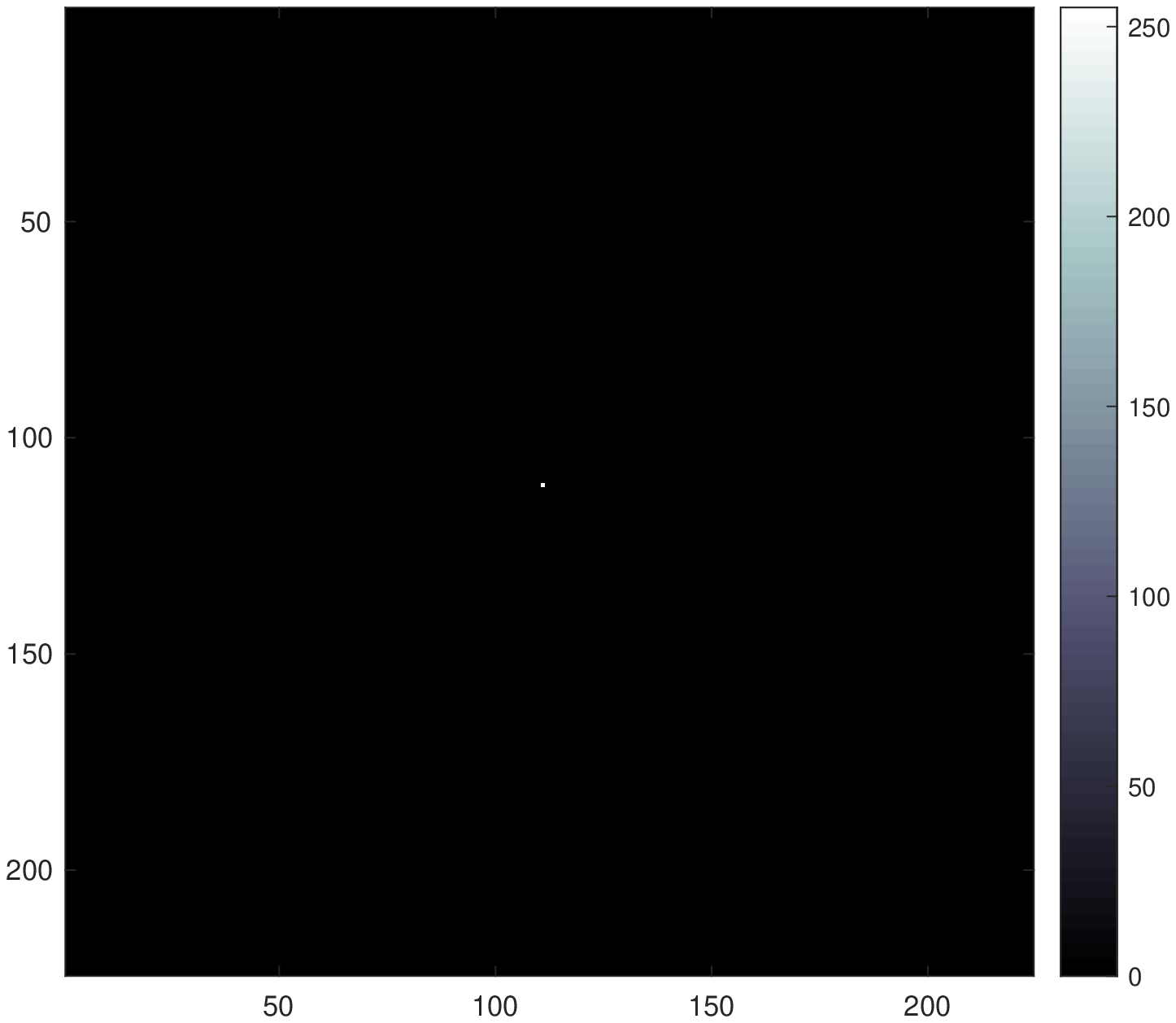}
\label{impulse_input}
}\hspace{0.025\textwidth}
\subfloat[Output $\mathbf{z}$]{
\includegraphics[trim={2.5cm 1cm 2.5cm 1.1cm},clip,width=0.32\columnwidth]{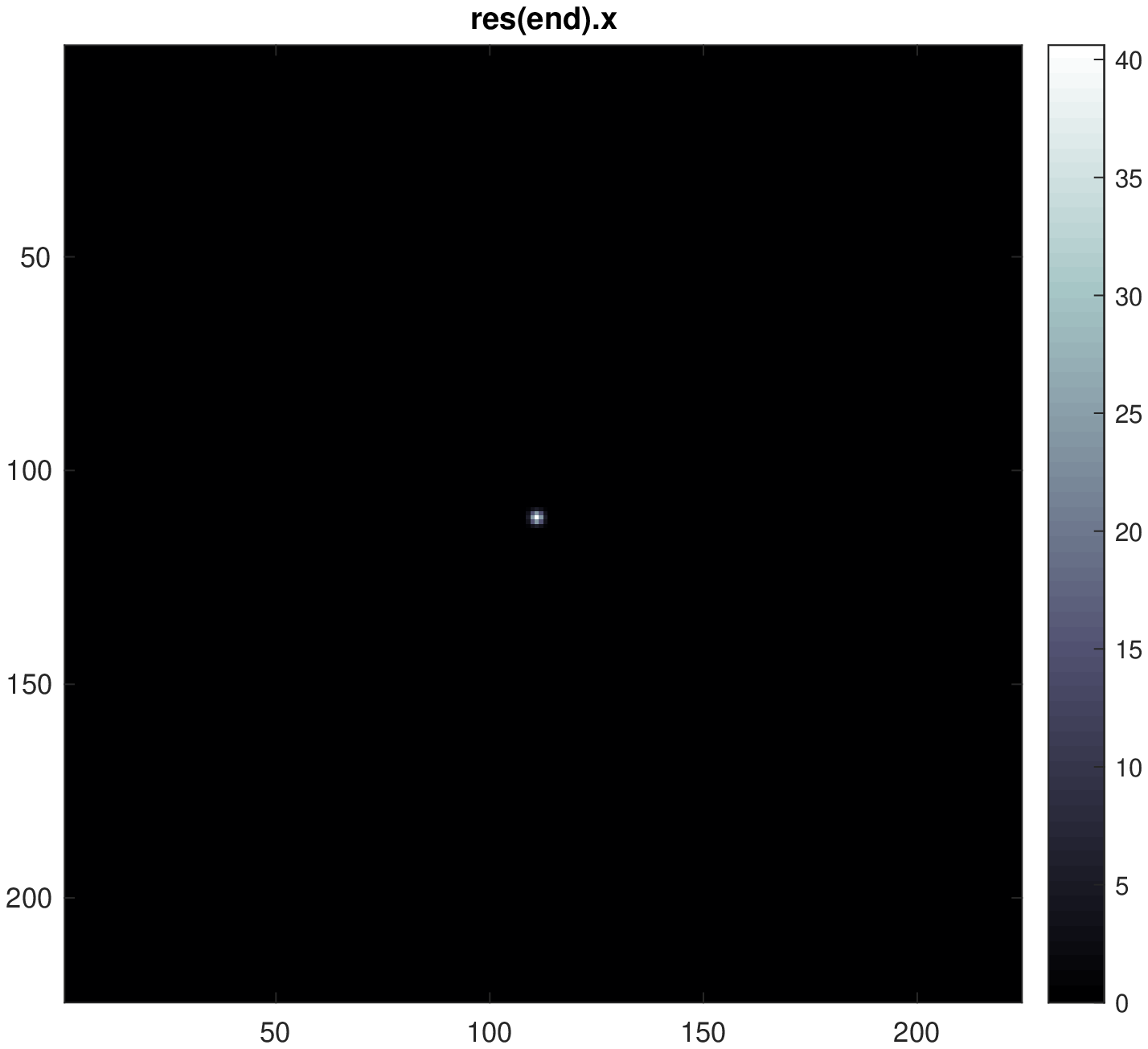}
\label{impulse_output}
}\\
\subfloat[$d\mathbf{z}/d\mathbf{x}$ obtained by setting $\mathbf{p}=1$ everywhere]{
\includegraphics[trim={2.5cm 1cm 2.5cm 1.1cm},clip,width=0.32\columnwidth]{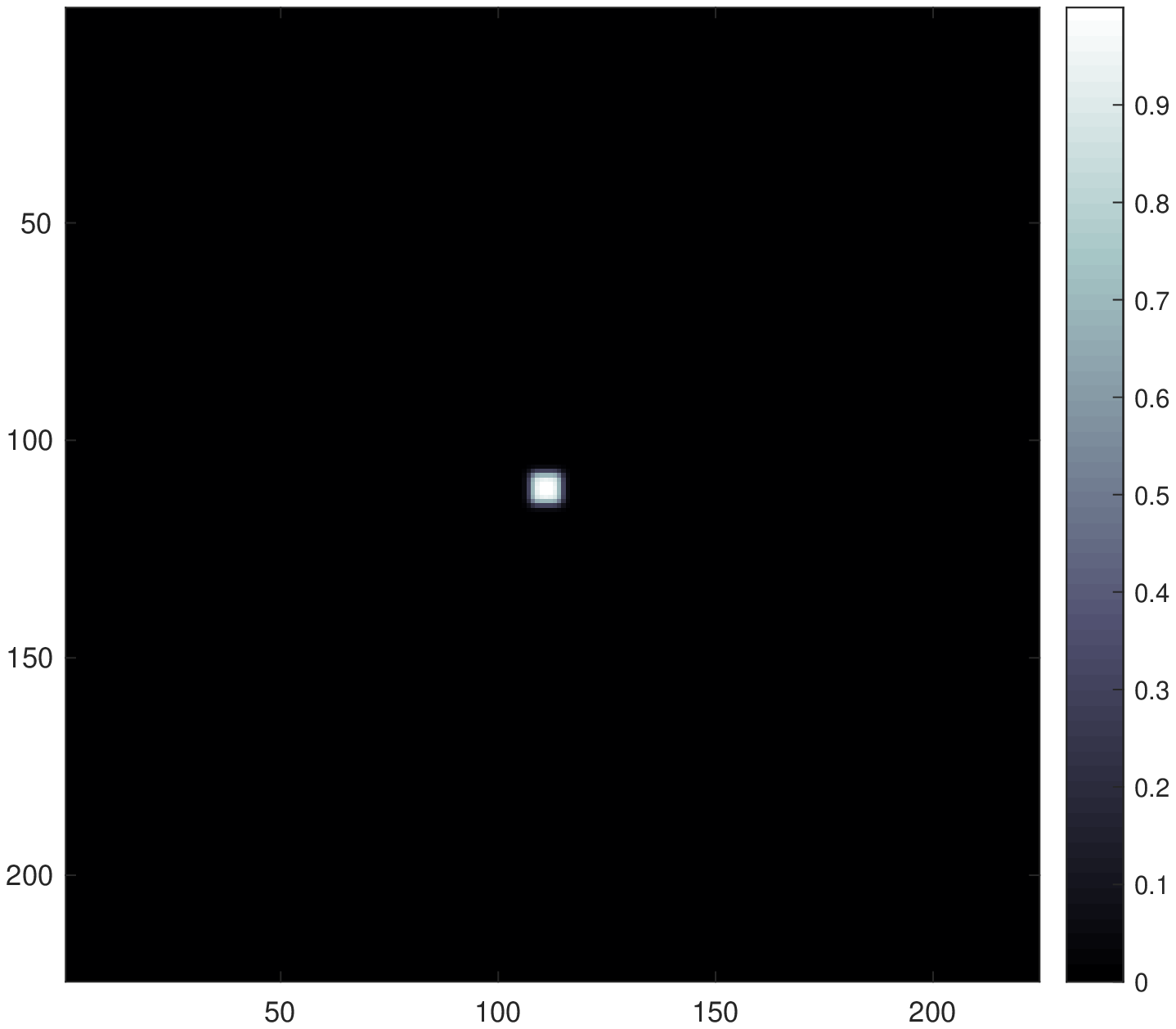}
\label{impulse_dzdx}}\hspace{0.025\textwidth}
\subfloat[Frequency response]{
\includegraphics[trim={2.5cm 1cm 2.5cm 1.1cm},clip,width=0.32\columnwidth]{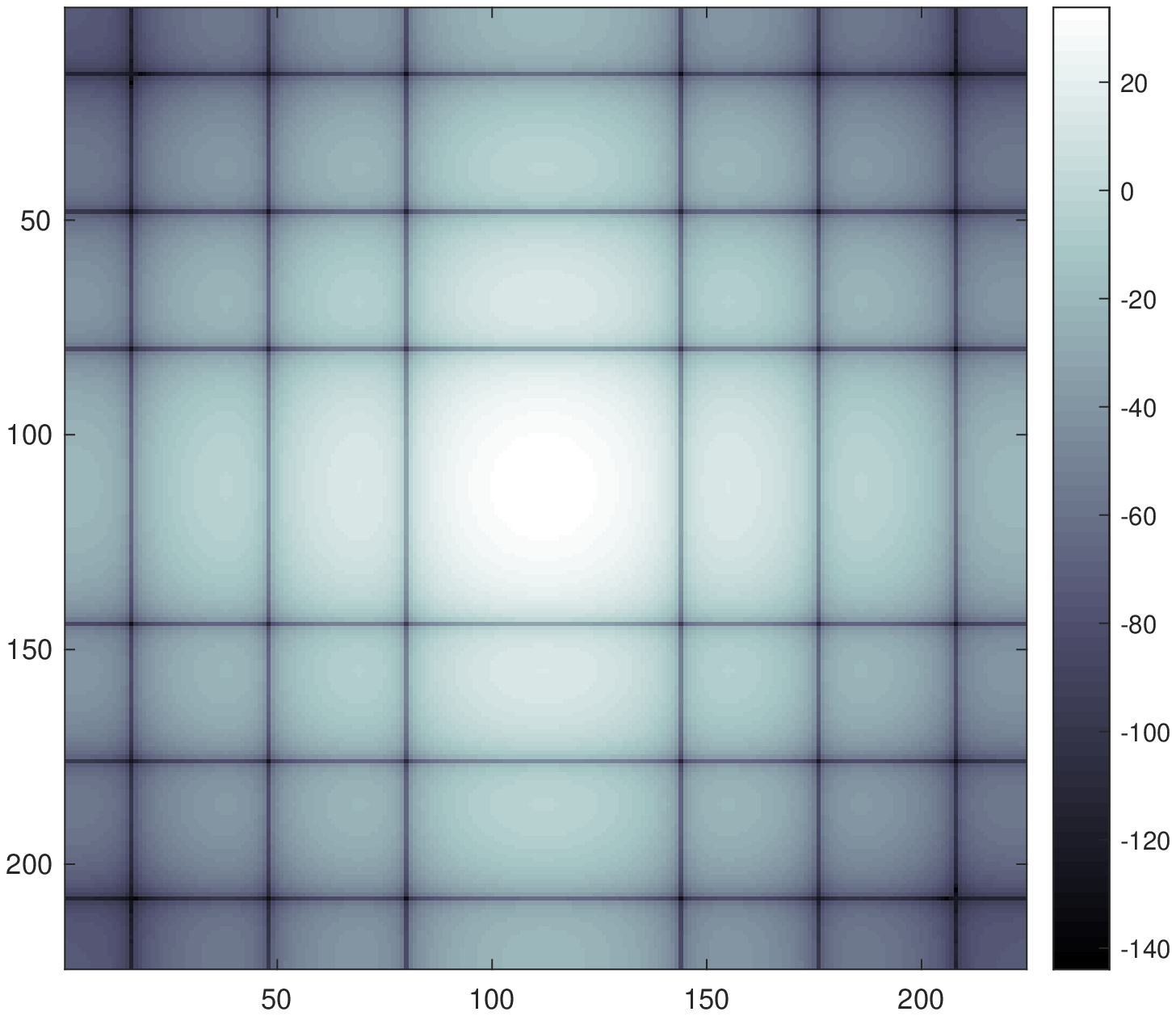}}
\caption{The toy CNN frequency response (in dB) obtained via a forward and backward pass over an impulse image.}
\label{impulse_forward_backward_pass}
\end{figure}

Next we computed the toy CNN frequency response for the synthetic image
of a 2D Dirac-Delta function (Figure~\ref{impulse_input}), whose channels have values of $255$ in the central pixel and are $0$ elsewhere.
Using an \emph{impulse image} input is motivated by its Fourier transform pair, which is a constant at all spatial frequencies.
Performing a forward pass (Figure~\ref{impulse_output}) and
setting $\mathbf{p}$ to $1$ everywhere yields a data gradient
that is a superposition of filter impulse responses
about the non-zero input pixel location (Figure~\ref{impulse_dzdx}).
Before computing the Fast Fourier Transform of $d\mathbf{z}/d\mathbf{x}$ for impulse images, a Hann window was applied to reduce edge effects.
The resulting frequency response is shown in Figure~\ref{impulse_output}.

\subsection{ImageNet CNNs}

\begin{figure*}
\centering
\subfloat[Input Red Channel $x$]{
\begin{tabular}[b]{c}%
\includegraphics[trim={0.7cm 0.7cm 0.7cm 0.2cm},clip,width=0.37\columnwidth]{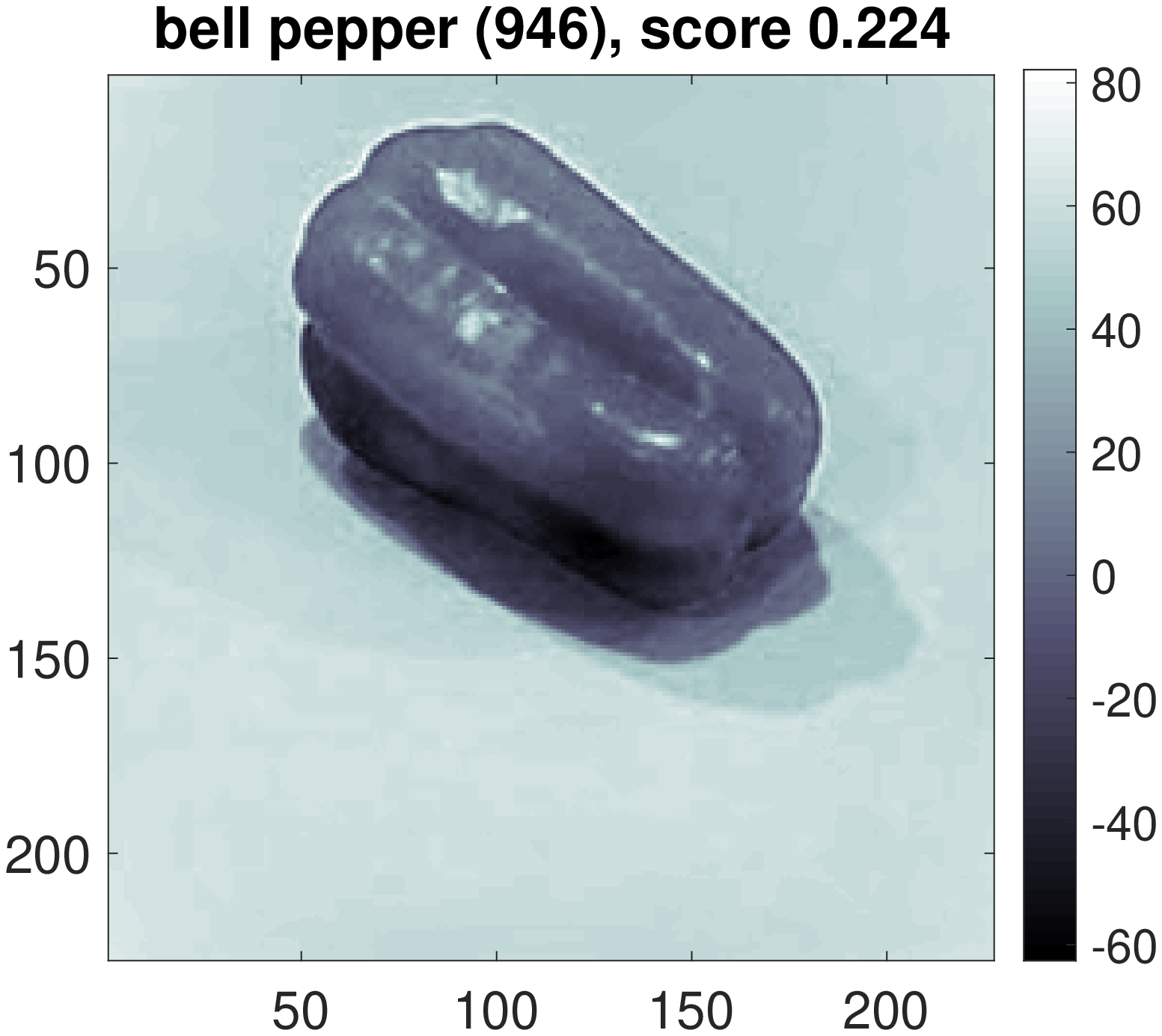}\\
\includegraphics[trim={0.7cm 0.7cm 0.7cm 0.2cm},clip,width=0.37\columnwidth]{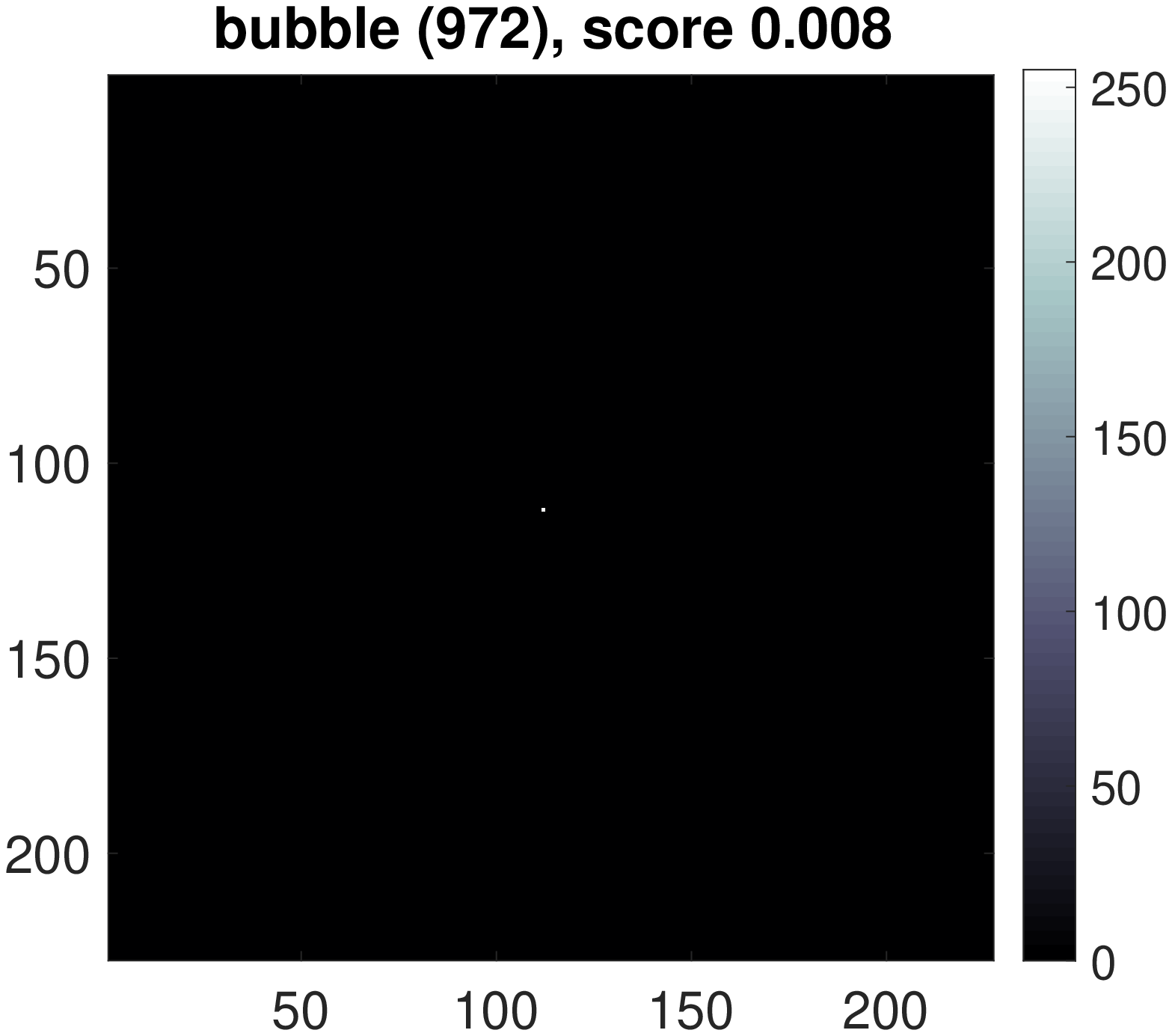}
\end{tabular}
}
\subfloat[Data Derivative $dz/dx$]{
\begin{tabular}[b]{c}%
\includegraphics[trim={0.7cm 0.7cm 0.7cm 0.2cm},clip,width=0.36\columnwidth]{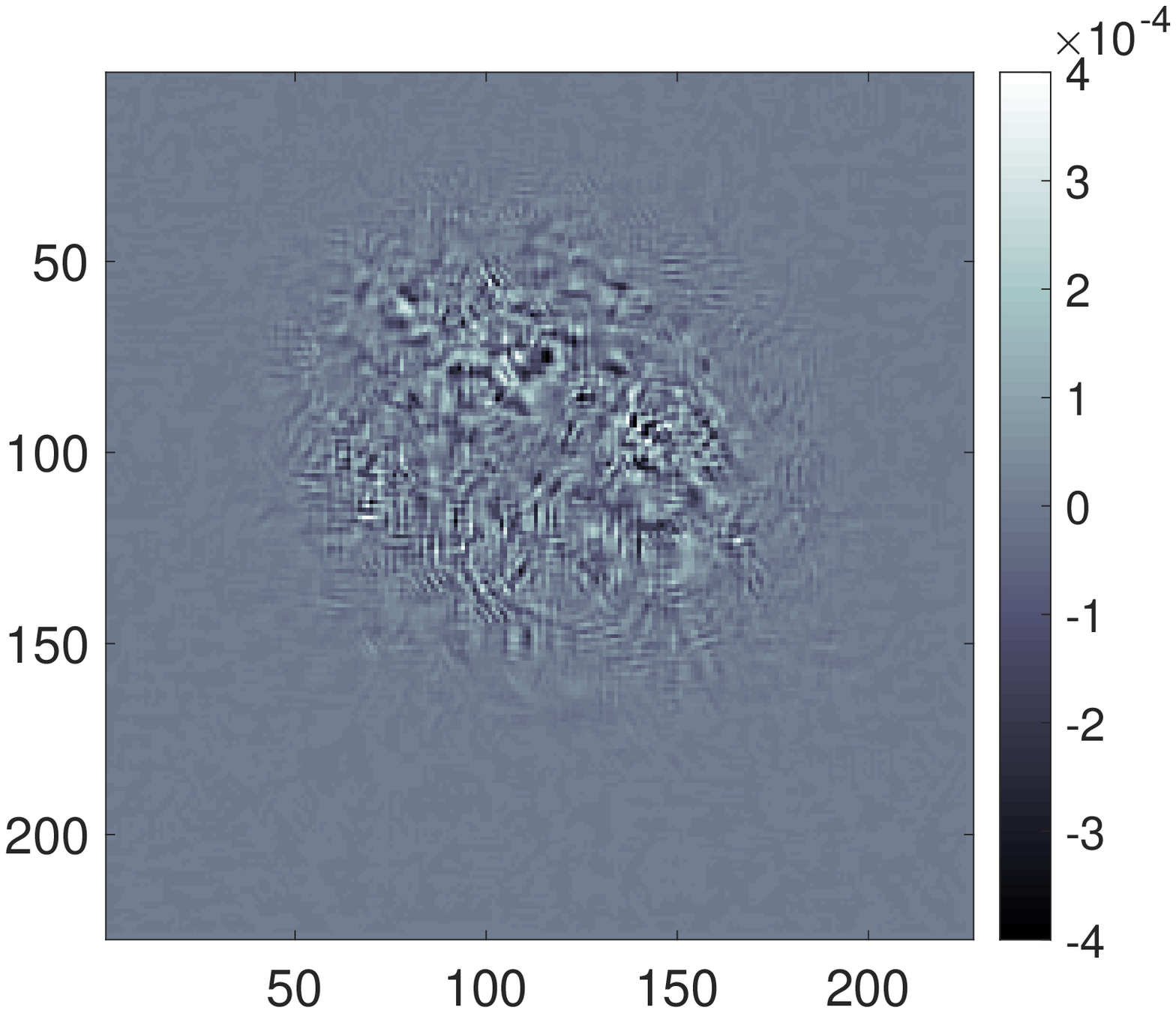}\\
\includegraphics[trim={0.7cm 0.7cm 0.7cm 0.2cm},clip,width=0.36\columnwidth]{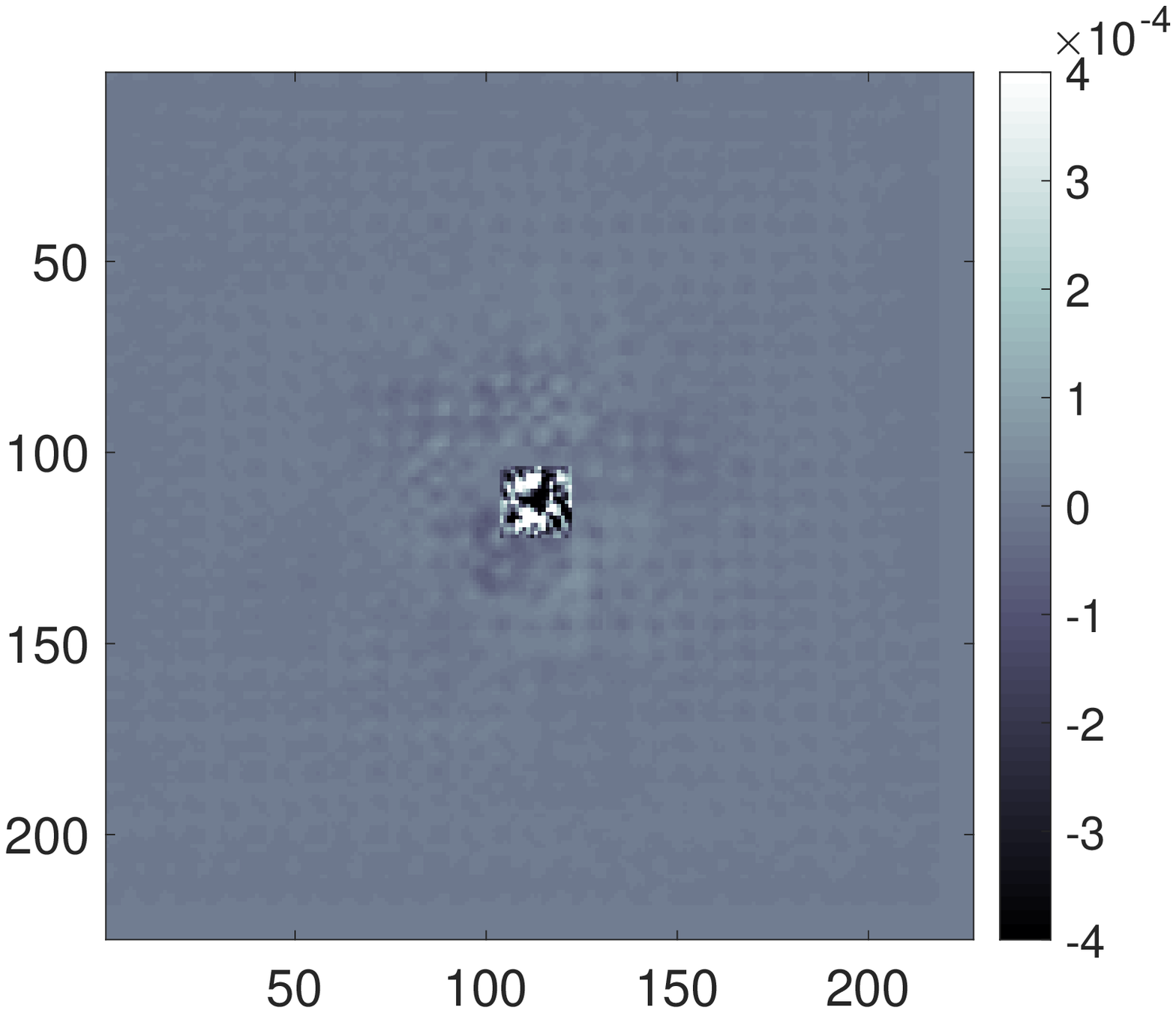}
\label{class-saliency}
\end{tabular}
}
\subfloat[CNN frequency response (dB)]{\begin{tabular}[b]{c}%
\includegraphics[trim={0.7cm 0.7cm 0.7cm 0.2cm},clip,width=0.37\columnwidth]{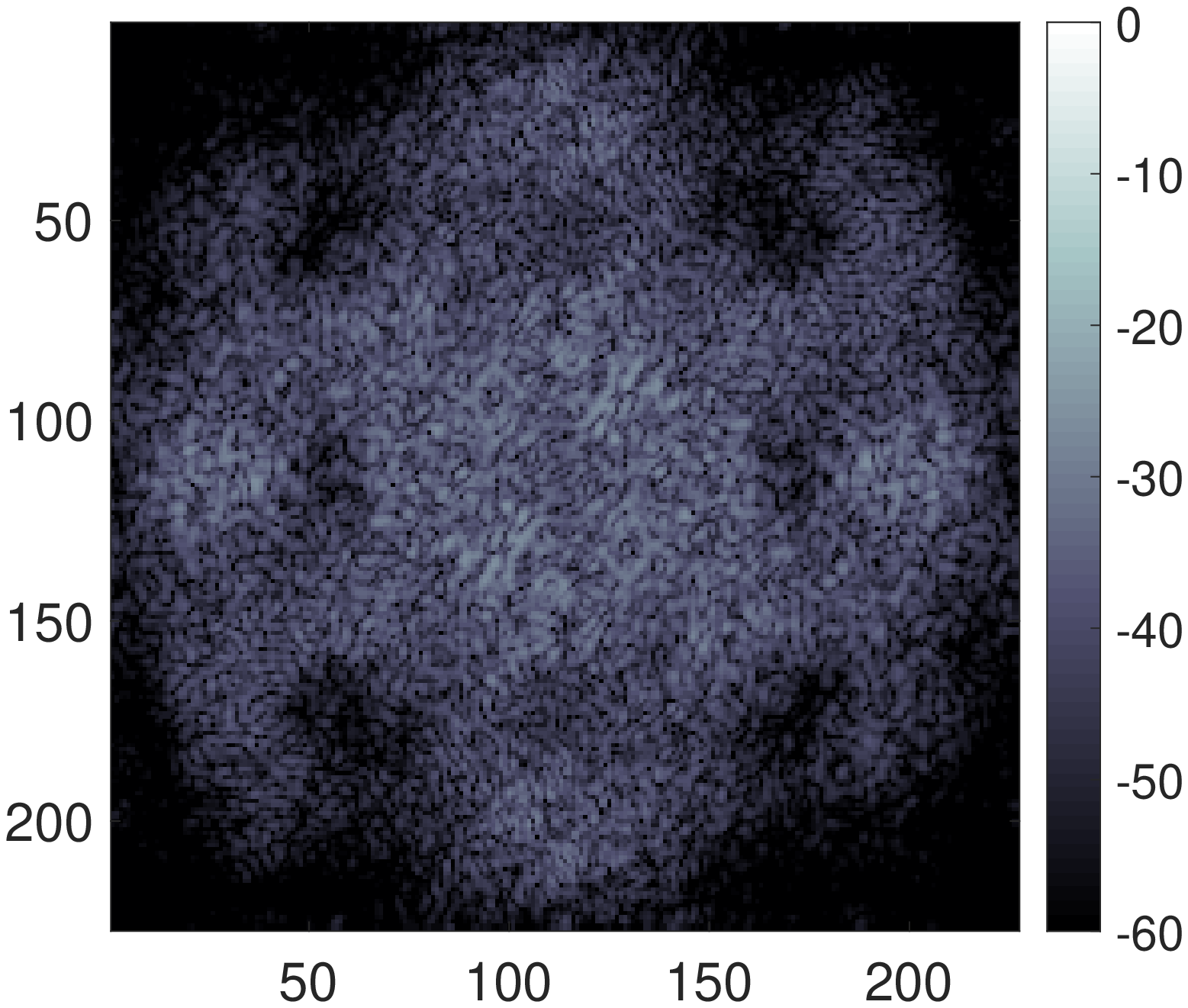}\\
\includegraphics[trim={0.7cm 0.7cm 0.7cm 0.2cm},clip,width=0.37\columnwidth]{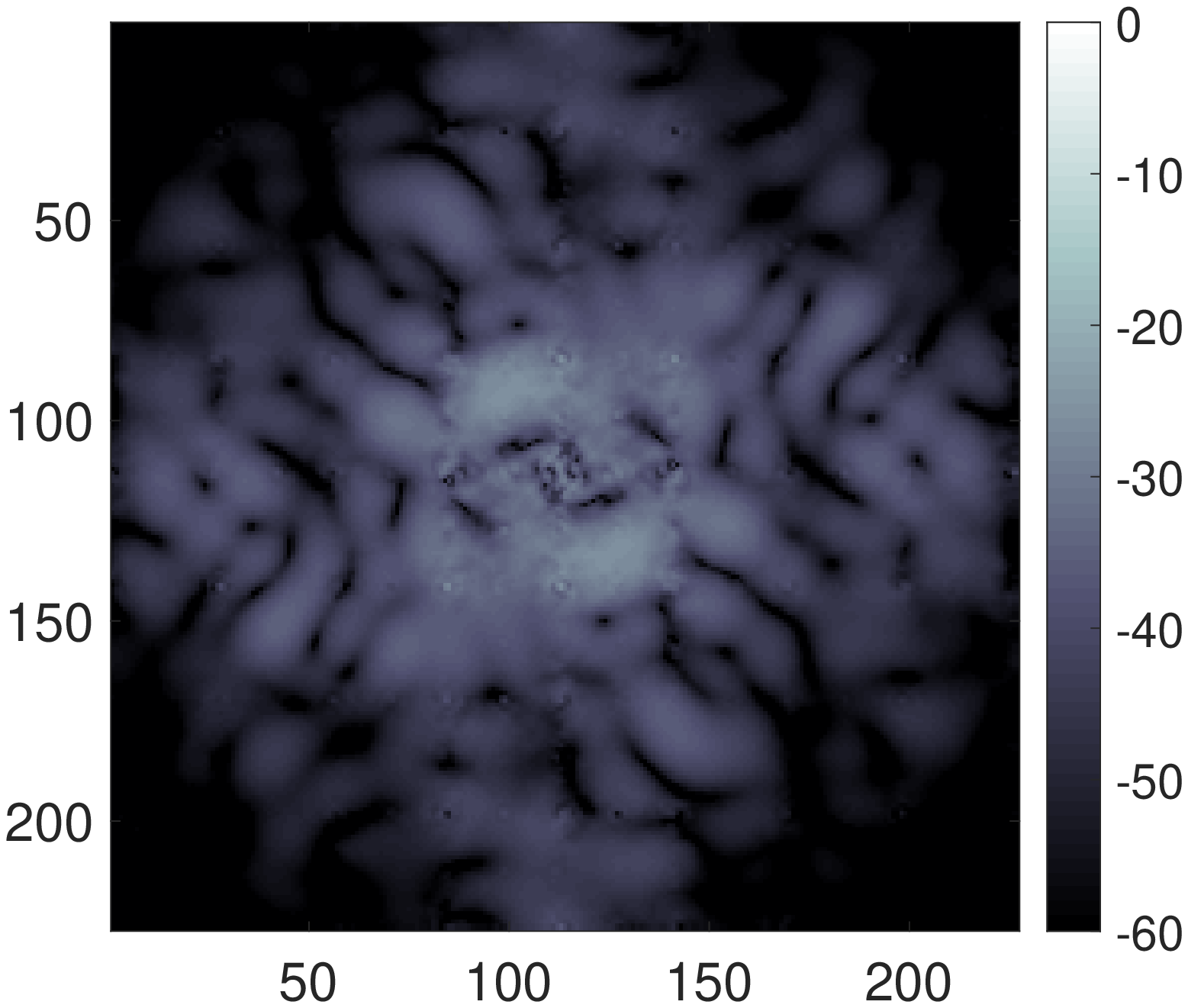}
\label{amplitude-spectrum}
\end{tabular}
}
\caption{The data derivative and frequency response of AlexNet~\cite{krizhevsky2012imagenet}, computed using the pre-processed red channel of a natural image (\emph{top}) and an impulse image (\emph{bottom}). The predicted class, its index in the output vector of probabilities and the resulting confidence \emph{score} obtained following the forward pass are also given for each input image.}
\label{alexnet}
\end{figure*}

In order to measure the frequency response of CNNs such as AlexNet~\cite{krizhevsky2012imagenet}, which was trained to perform image classification on ImageNet~\cite{ILSVRC15}, the tensor $\mathbf{p}$ must match the dimension of the output vector of class probabilities $\mathbf{z}$. This is given by the application of the \emph{softmax} function to the \emph{logit} vector $\mathbf{l}$ of units in the final layer:
\begin{equation}
  z_i = \frac{\exp(l_i)}{\sum_{j=1}^N \exp( l_j)},
\label{eqn:softmax}
\end{equation}
where $i=1,\ldots,N$ is the class index and, for ImageNet, $N=1000$ object classes.
The predicted class index and its confidence score are extracted as
$i = \argmax \: \mathbf{z}$ and $\text{\emph{score}} = \max \: \mathbf{z}$, respectively.

We consider only the frequency response for the winning class by using the projection tensor to extract a data derivative with respect to the output unit $i$ by setting all other elements in $\mathbf{p}$ to zero.
Given that scaling $\mathbf{p}$ also scales $d\mathbf{z}/d\mathbf{x}$ by the same factor (see Eq.~\eqref{eqn:taylor}),
to enable comparison between different CNNs we set its non-zero element to $p_i=1/\text{\emph{score}}$.
This normalizes the frequency response for the different CNN scores obtained from respective forward passes over the same input image.

Figure~\ref{alexnet} shows the data derivative and frequency response
for the red channel of a natural image and an impulse image, both obtained using AlexNet.
The impulse image had the same input repeated across its red, green and blue channels.
Simonyan \emph{et al.}~\cite{simonyan2013deep} refer to examples like those in Figure~\ref{class-saliency}
as \emph{image-specific class saliency} because they indicate which input pixels have most impact on the score.
We found it difficult to characterize the shape of the resulting $2D$ amplitude spectra in Figure~\ref{amplitude-spectrum} with a single metric
(\emph{e.g.} \emph{bandwidth}), however, in terms of the overall gain, taking the surface maximum
provides the \emph{Maximum Gain} (in dB) of the frequency response.

\begin{figure}
\adjustbox{valign=t}{
\begin{minipage}[t]{0.5\linewidth}
  \subfloat[ResNet-152~\cite{he2016deep} 6.7\%]{
  \centering
  \includegraphics[trim={1.5cm 0cm 1.5cm 0cm},clip,width=0.47\columnwidth]{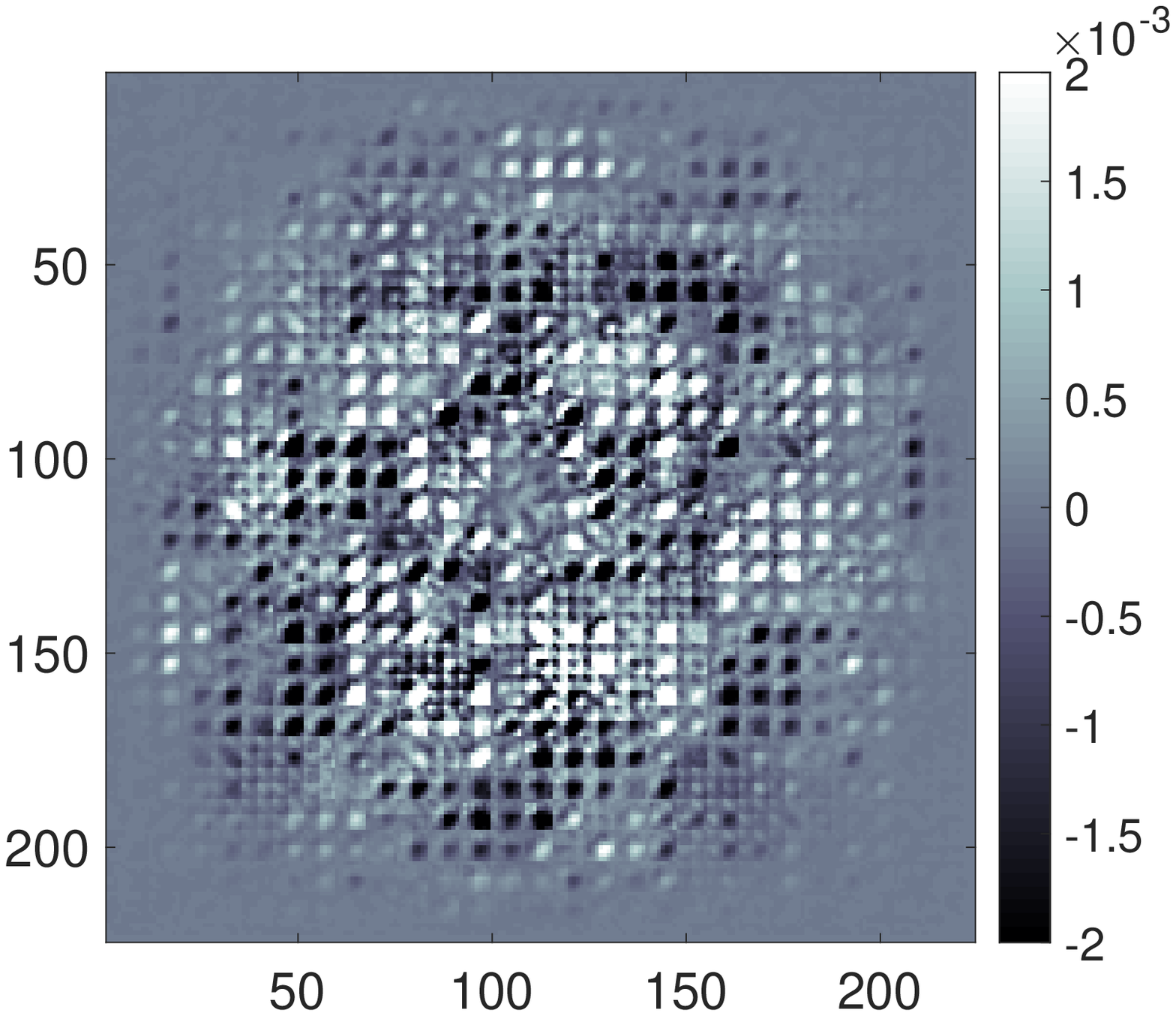}
  \includegraphics[trim={1.5cm 0cm 1.5cm 0cm},clip,width=0.47\columnwidth]{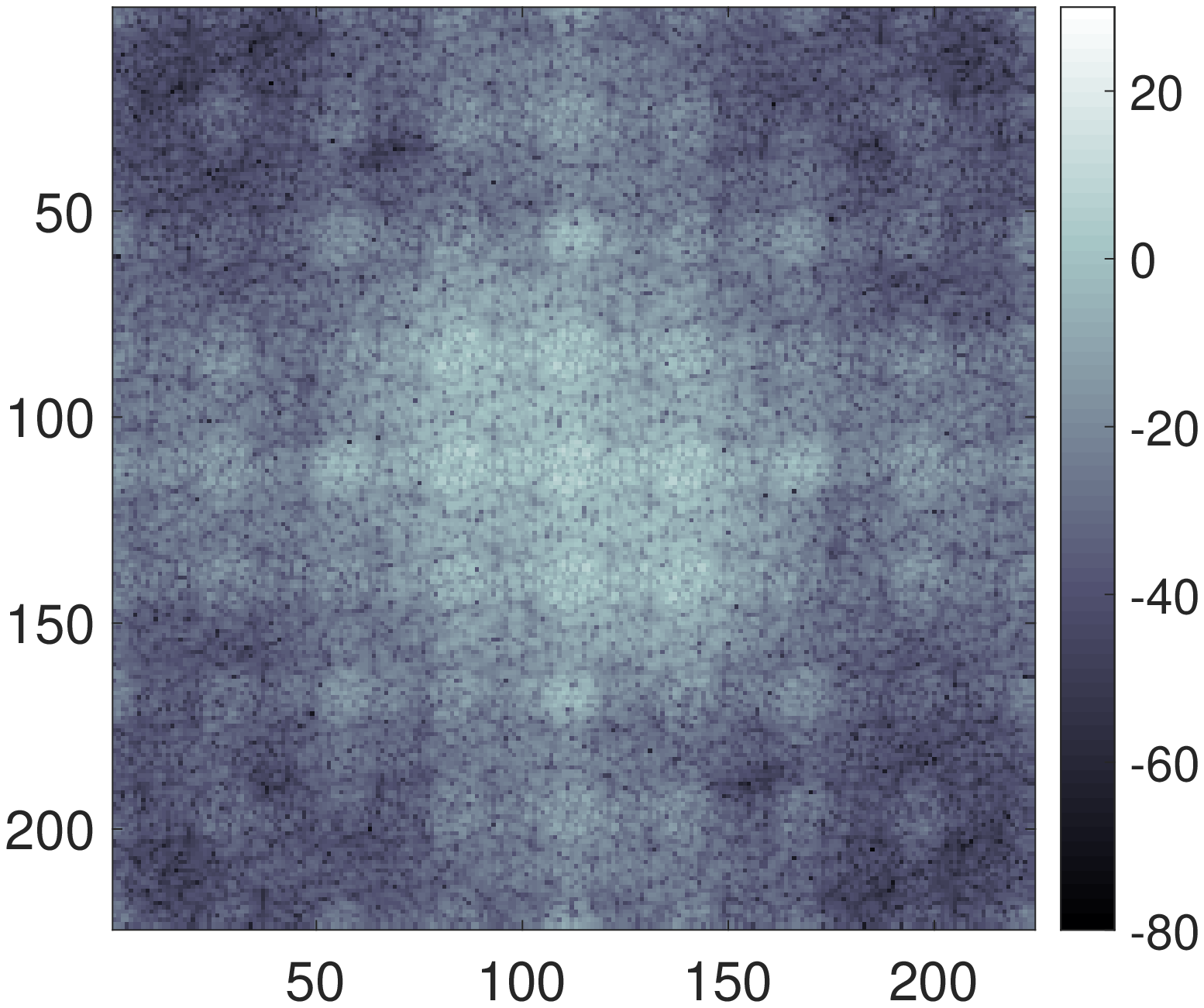}
  }\\
  \vspace*{-0.02mm}
  \subfloat[ResNet-101~\cite{he2016deep} 7.0\%]{
  \centering
  \includegraphics[trim={1.5cm 0cm 1.5cm 0cm},clip,width=0.47\columnwidth]{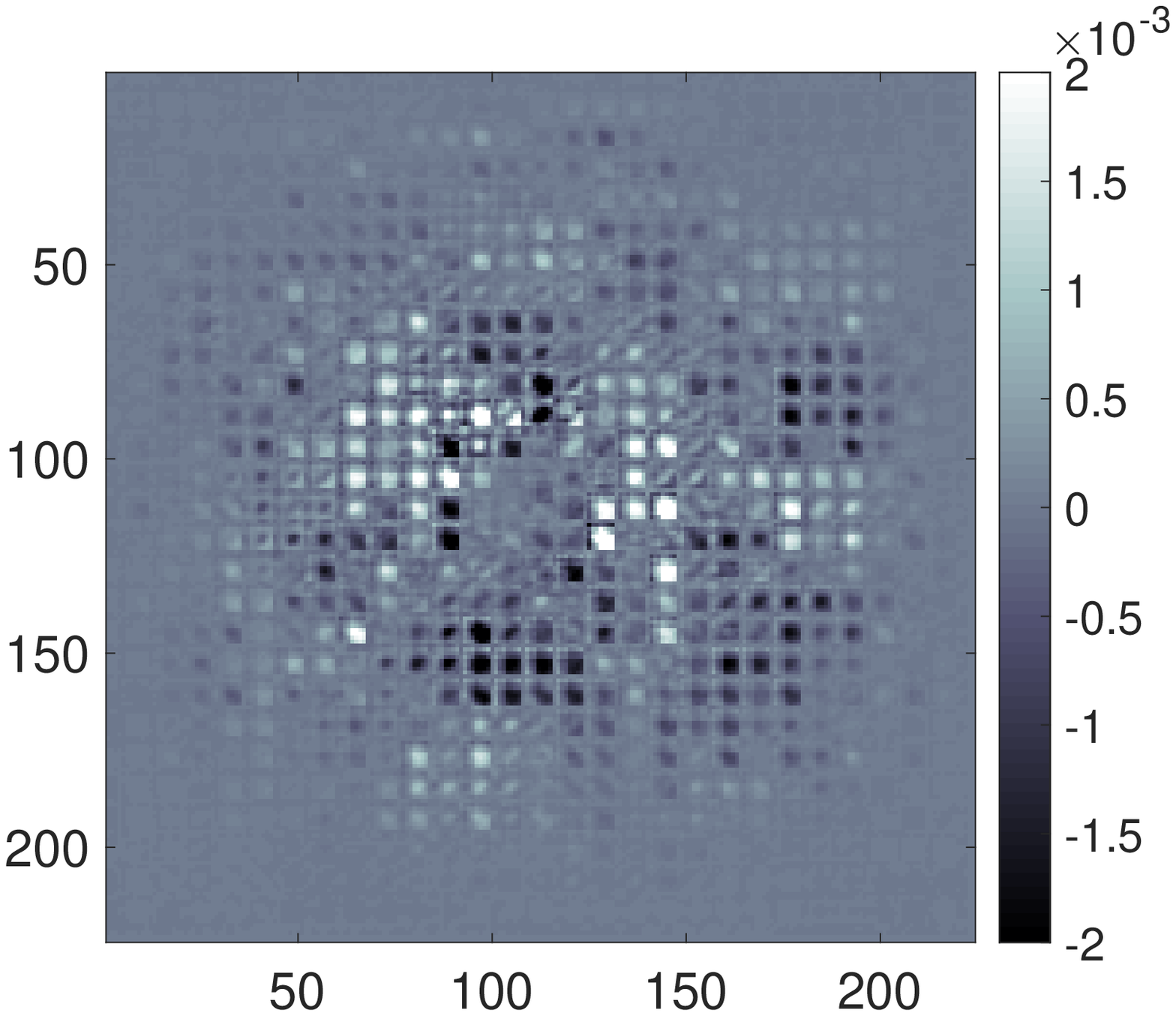}
  \includegraphics[trim={1.5cm 0cm 1.5cm 0cm},clip,width=0.47\columnwidth]{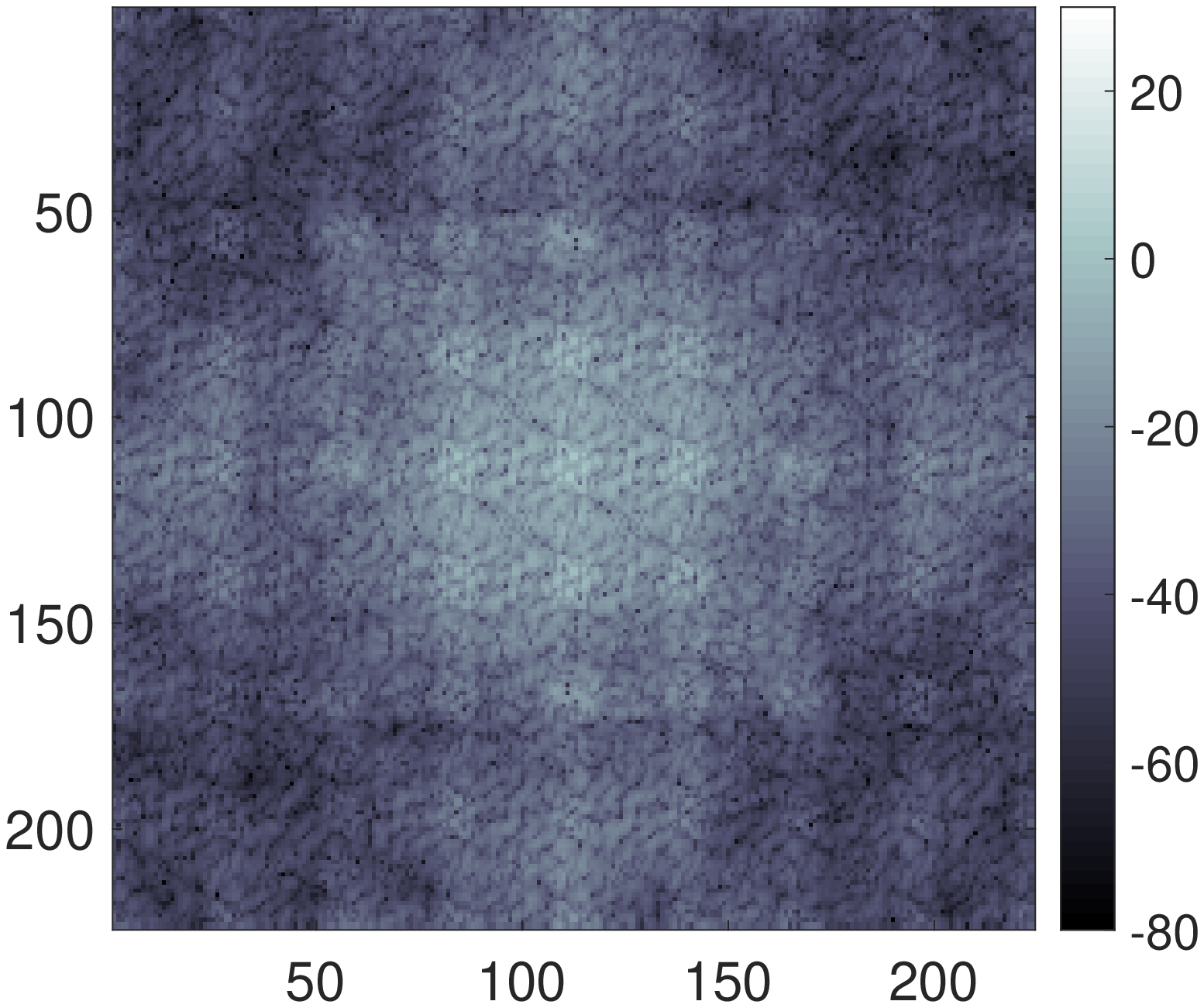}
  }\\
  \vspace*{-0.02mm}
  \subfloat[ResNet-50~\cite{he2016deep} 7.7\%]{
  \centering
  \includegraphics[trim={1.5cm 0cm 1.5cm 0cm},clip,width=0.47\columnwidth]{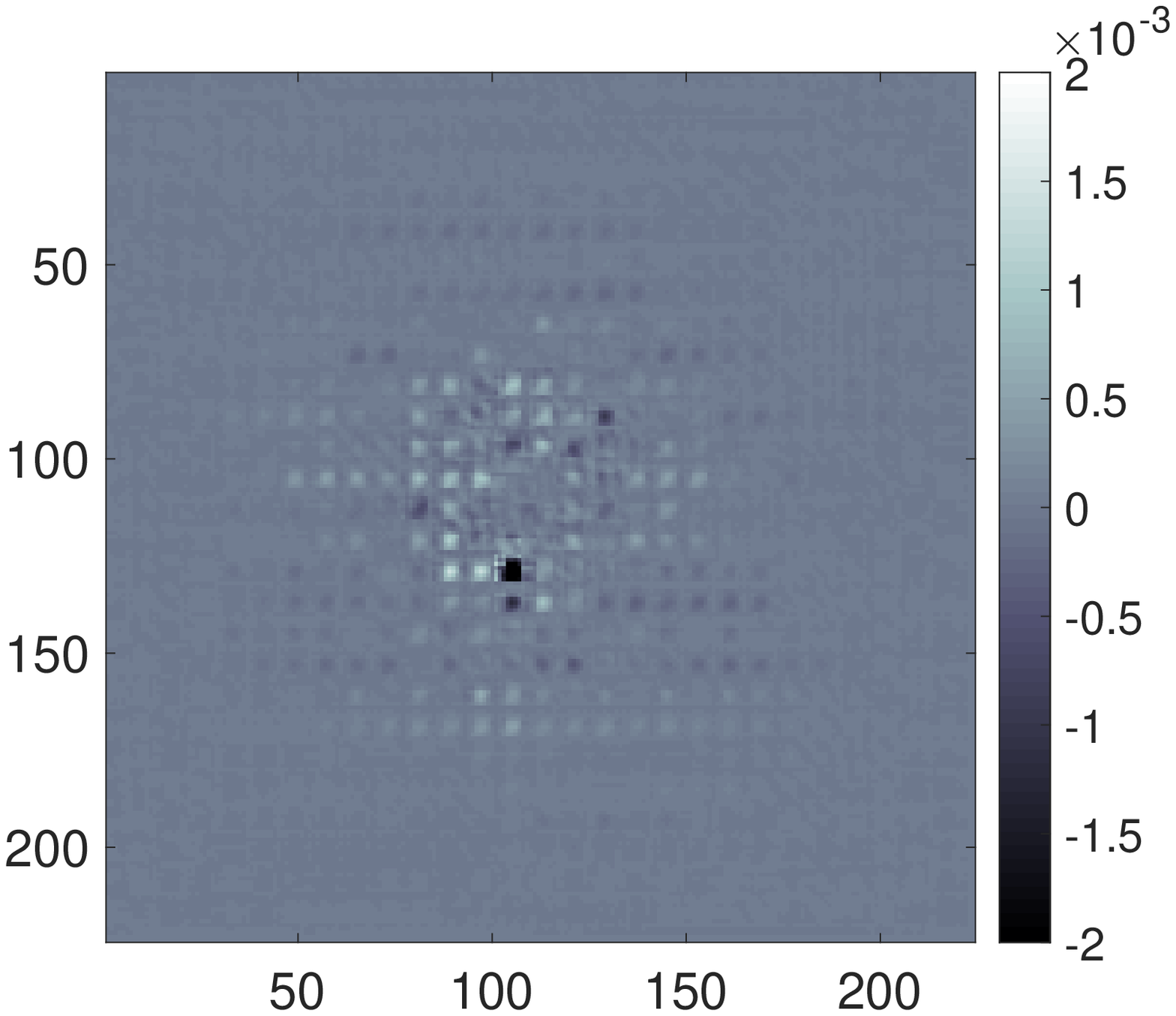}
  \includegraphics[trim={1.5cm 0cm 1.5cm 0cm},clip,width=0.47\columnwidth]{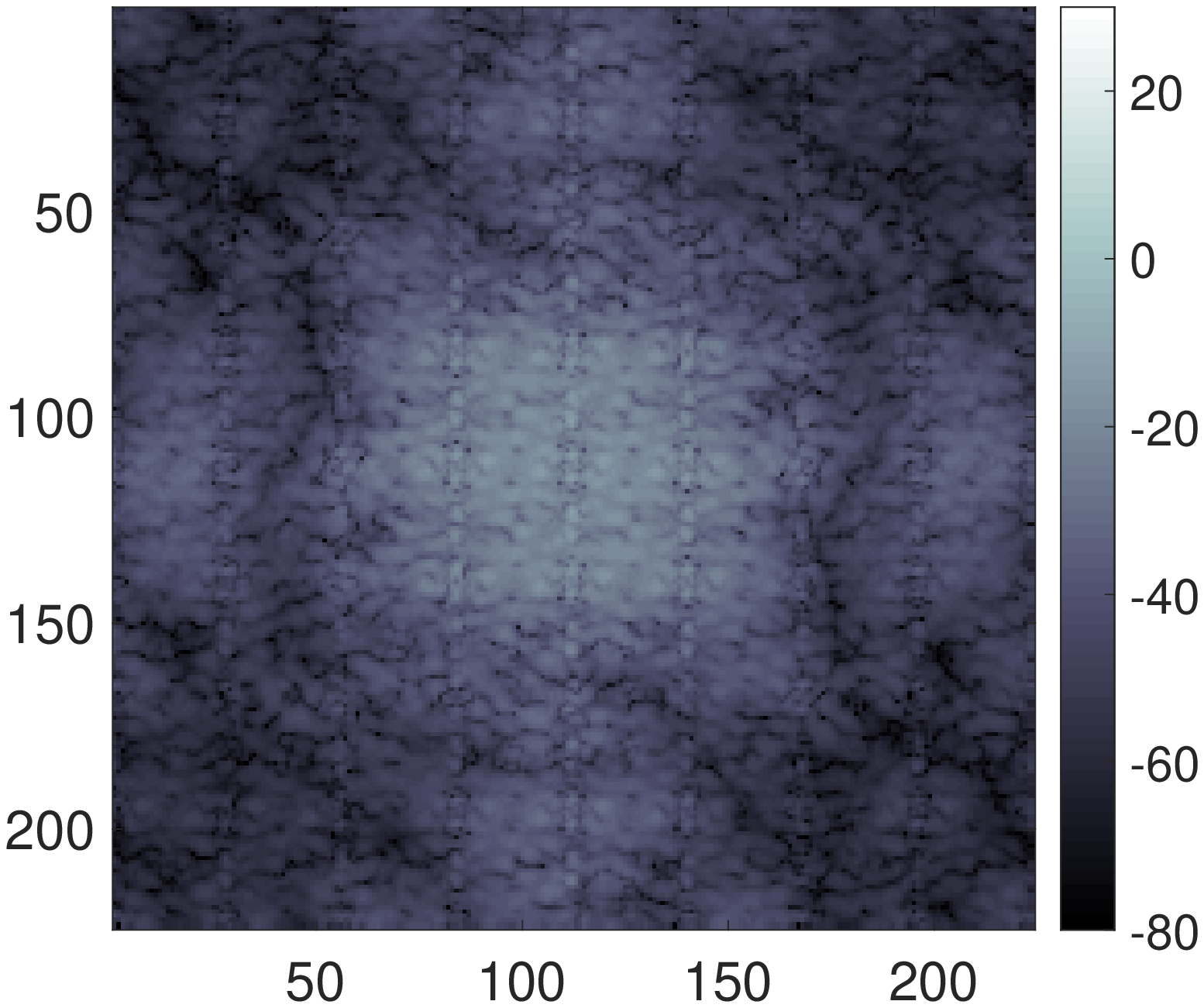} 
  }\\
  \vspace*{-0.02mm}
  \subfloat[MatConvNet-VGG-v.deep-16 9.5\%]{
  \centering
  \includegraphics[trim={1.5cm 0cm 1.5cm 0cm},clip,width=0.47\columnwidth]{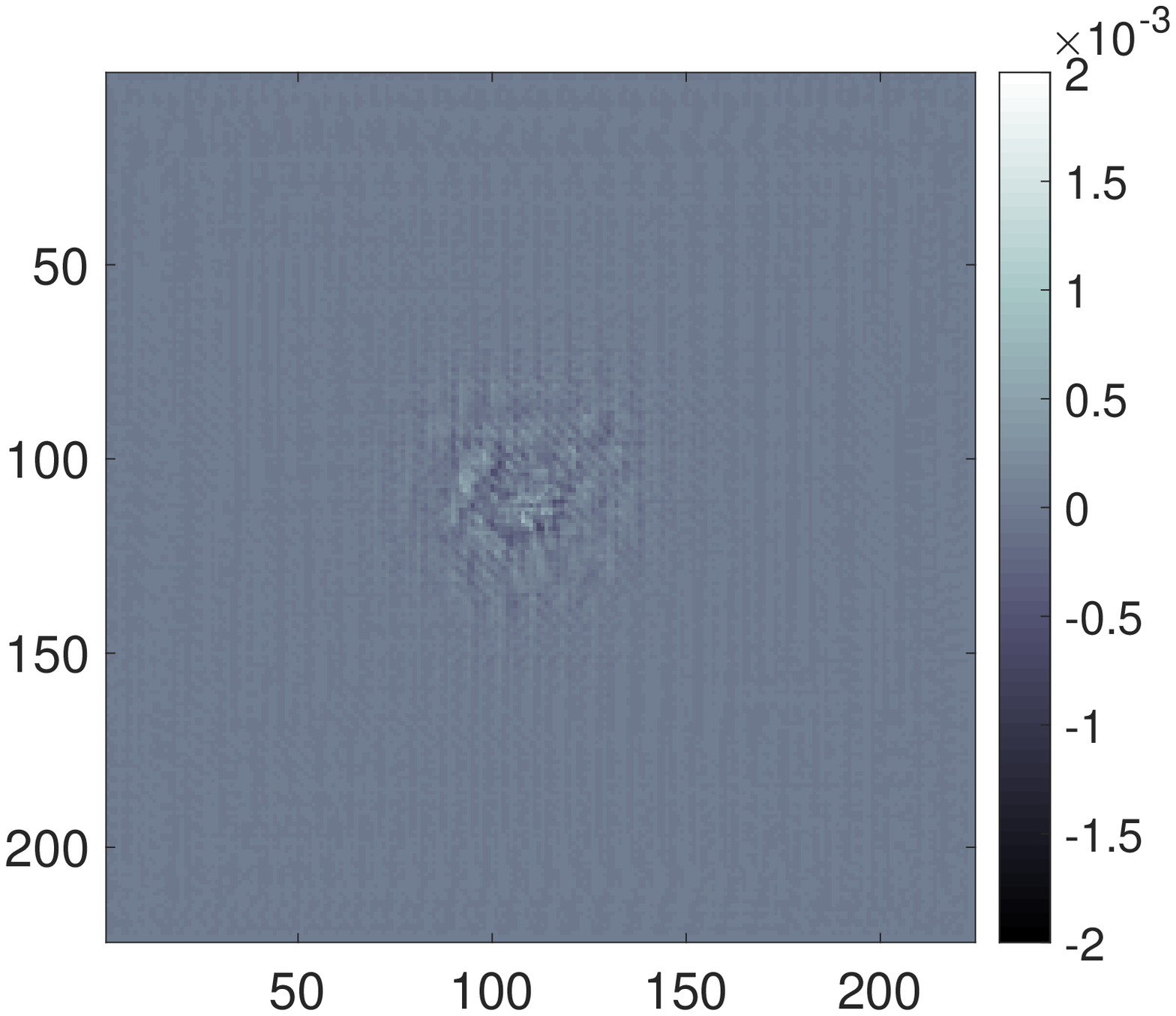}
  \includegraphics[trim={1.5cm 0cm 1.5cm 0cm},clip,width=0.47\columnwidth]{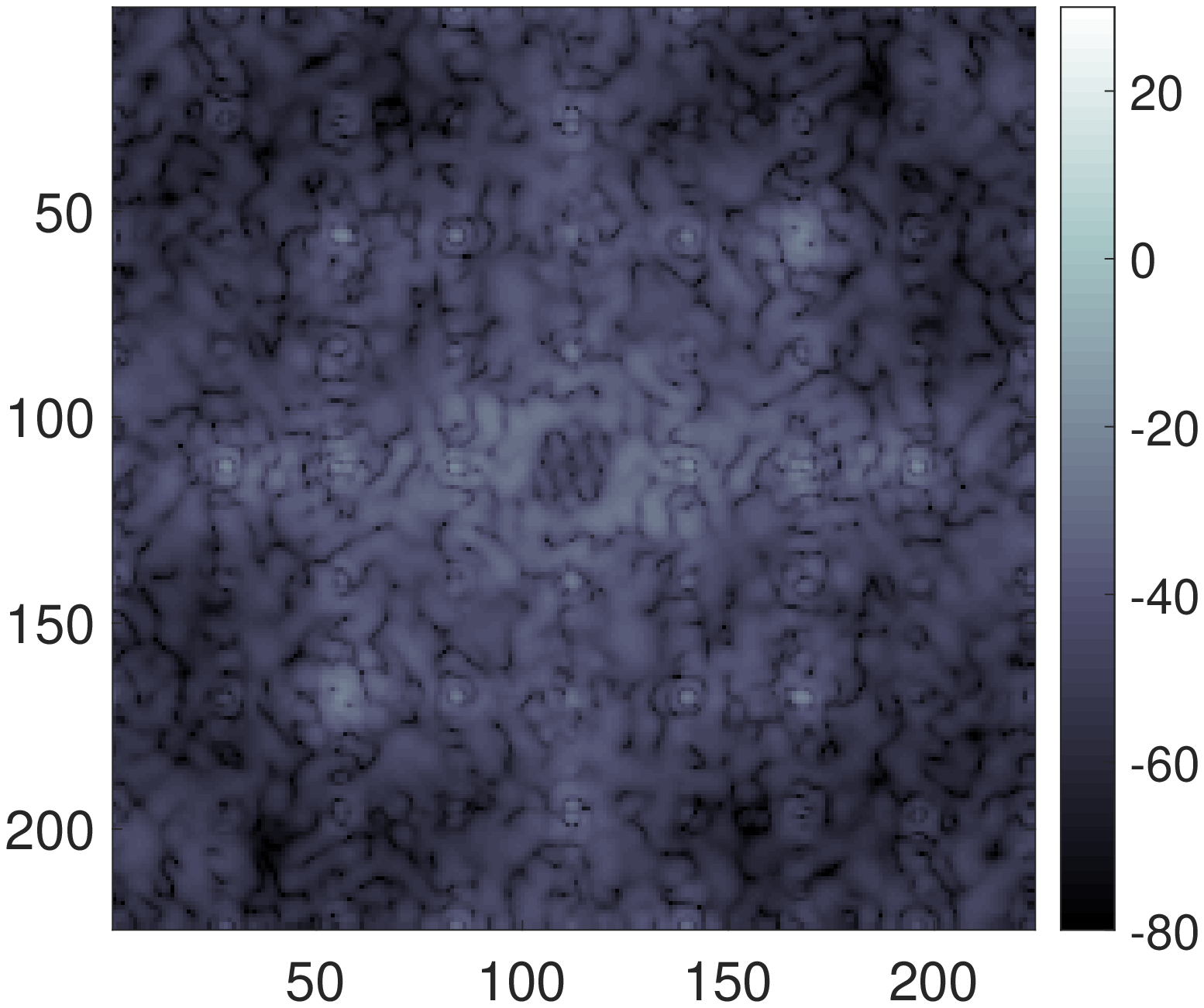} 
  }\\
  \vspace*{-0.02mm}
  \subfloat[VGG-v.deep-19~\cite{simonyan2014very} 9.9\%]{
  \centering
  \includegraphics[trim={1.5cm 0cm 1.5cm 0cm},clip,width=0.47\columnwidth]{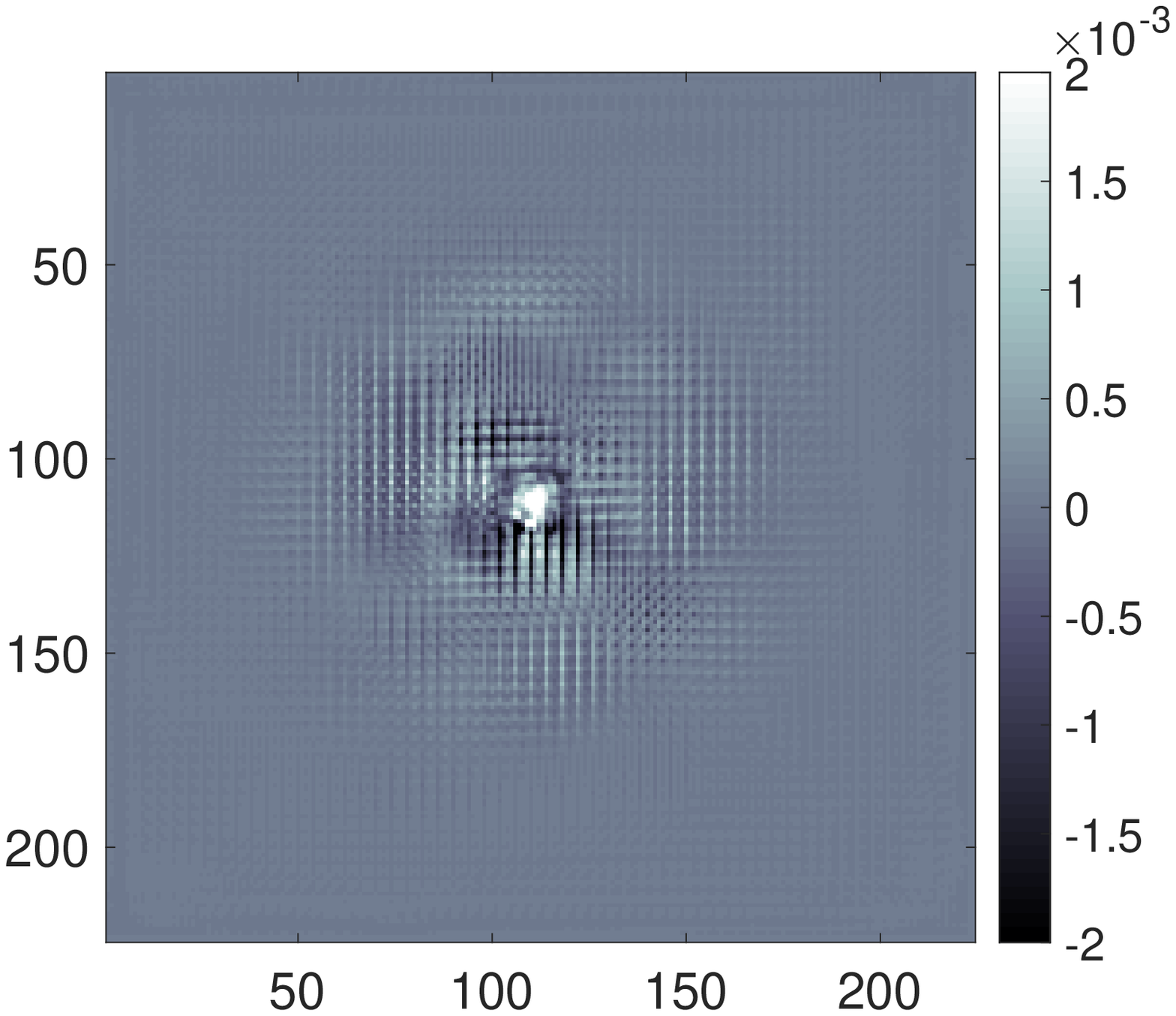}
  \includegraphics[trim={1.5cm 0cm 1.5cm 0cm},clip,width=0.47\columnwidth]{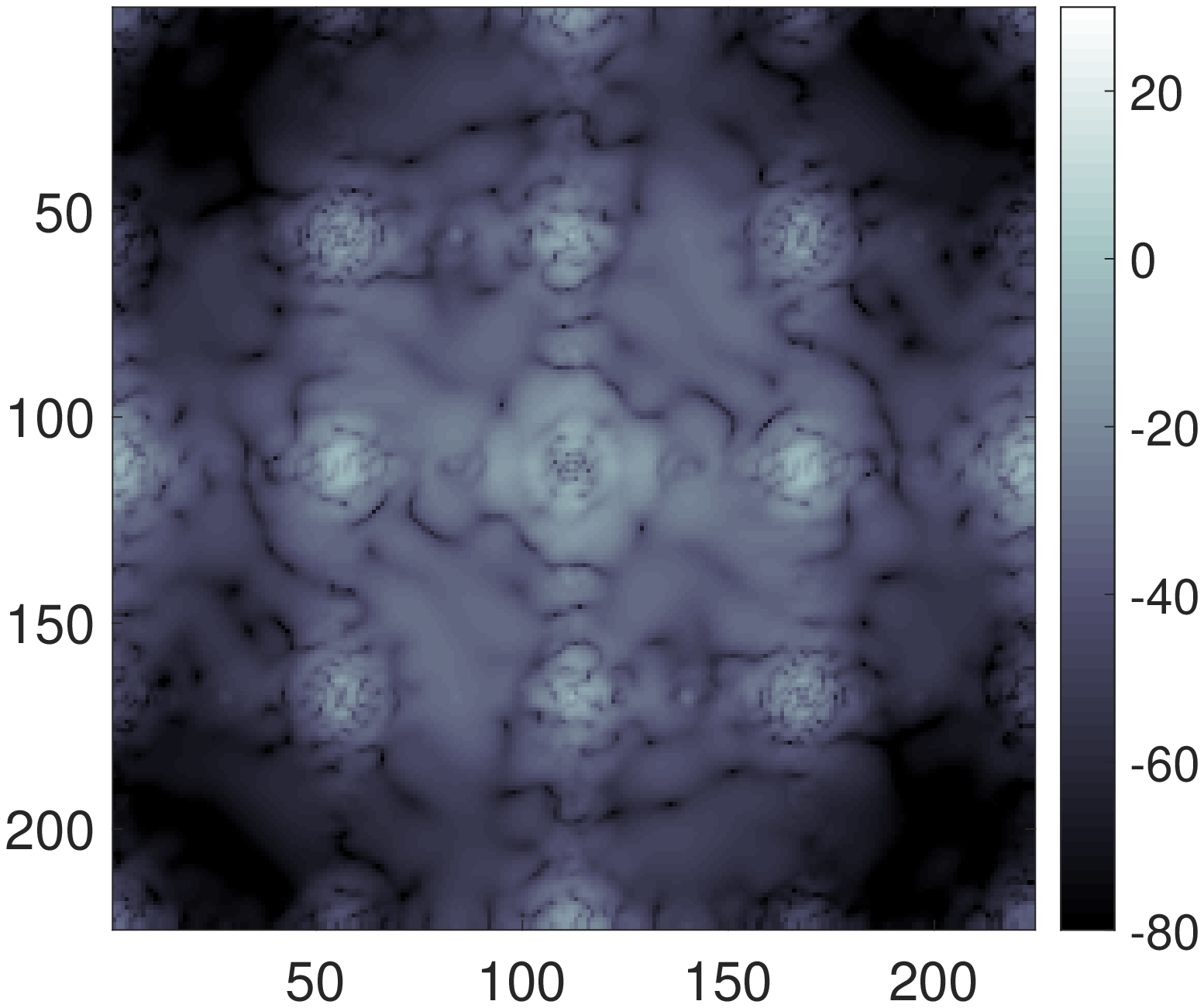} 
  }\\
  \vspace*{-0.02mm}
  \subfloat[VGG-v.deep-16~\cite{simonyan2014very} 9.9\%]{
  \centering
  \includegraphics[trim={1.5cm 0cm 1.5cm 0cm},clip,width=0.47\columnwidth]{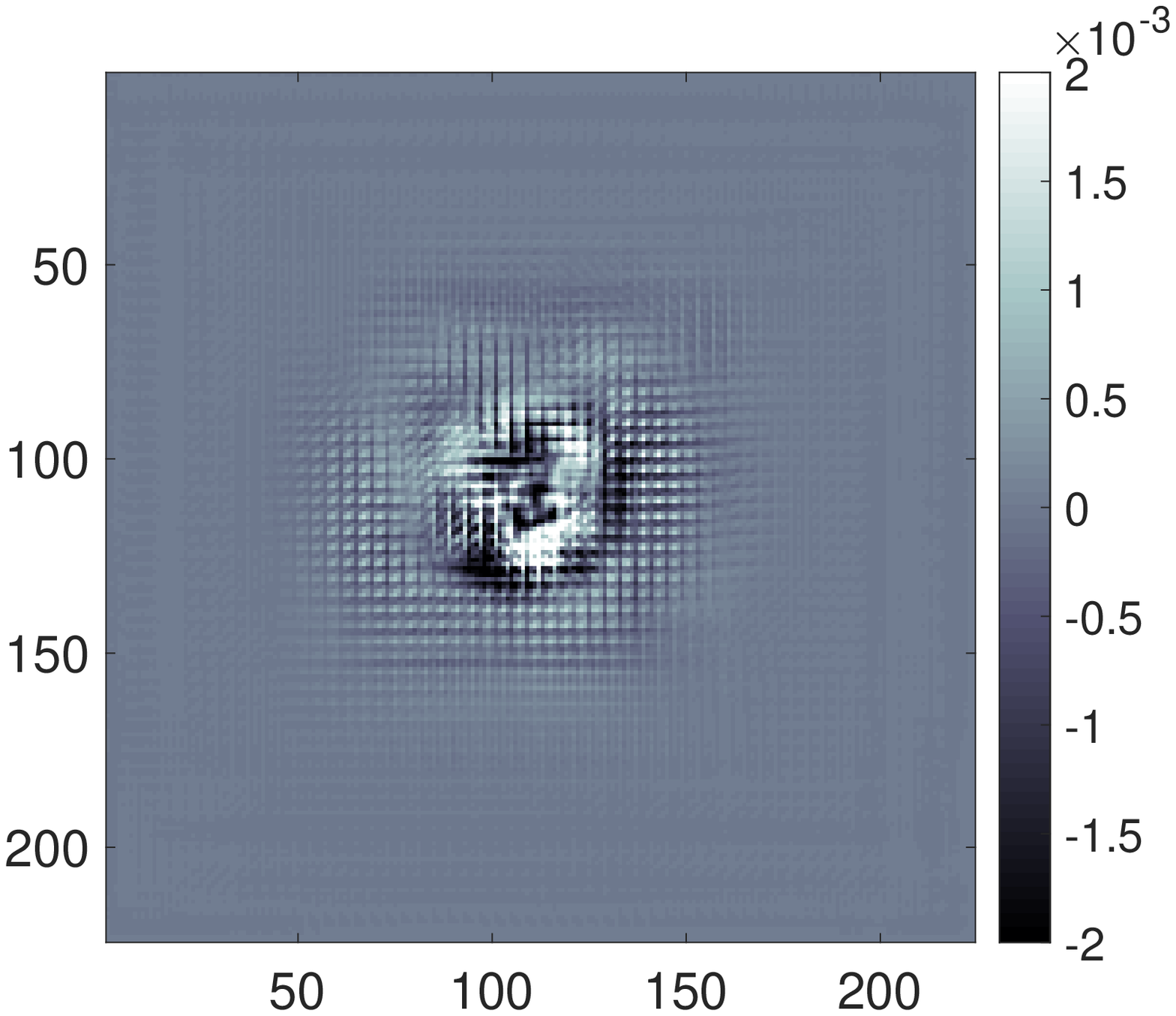}
  \includegraphics[trim={1.5cm 0cm 1.5cm 0cm},clip,width=0.47\columnwidth]{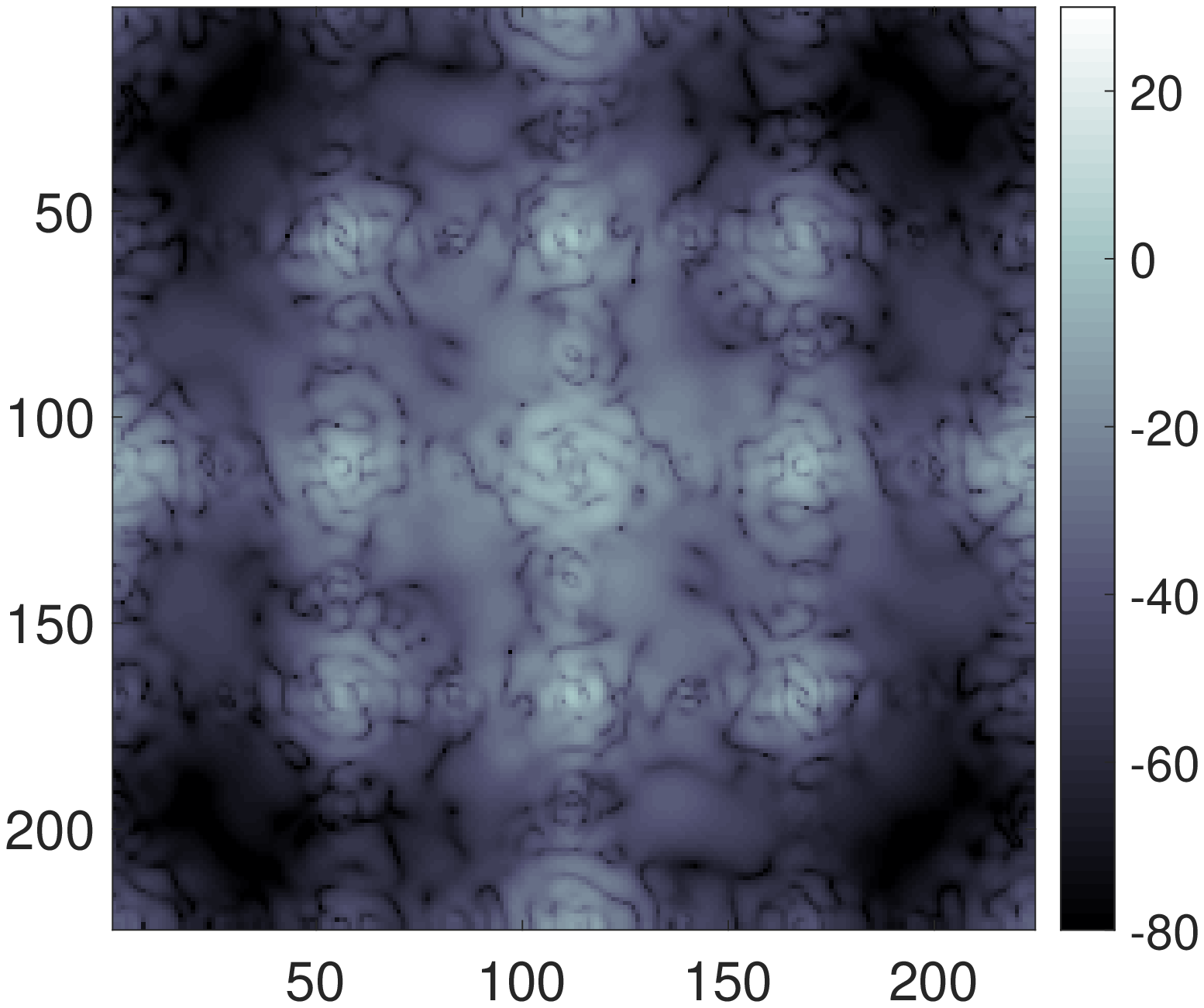} 
  }\\
  \vspace*{-0.02mm}
  \subfloat[GoogleNet~\cite{szegedy2015going} 12.9\%]{
  \centering
  \includegraphics[trim={1.5cm 0cm 1.5cm 0cm},clip,width=0.47\columnwidth]{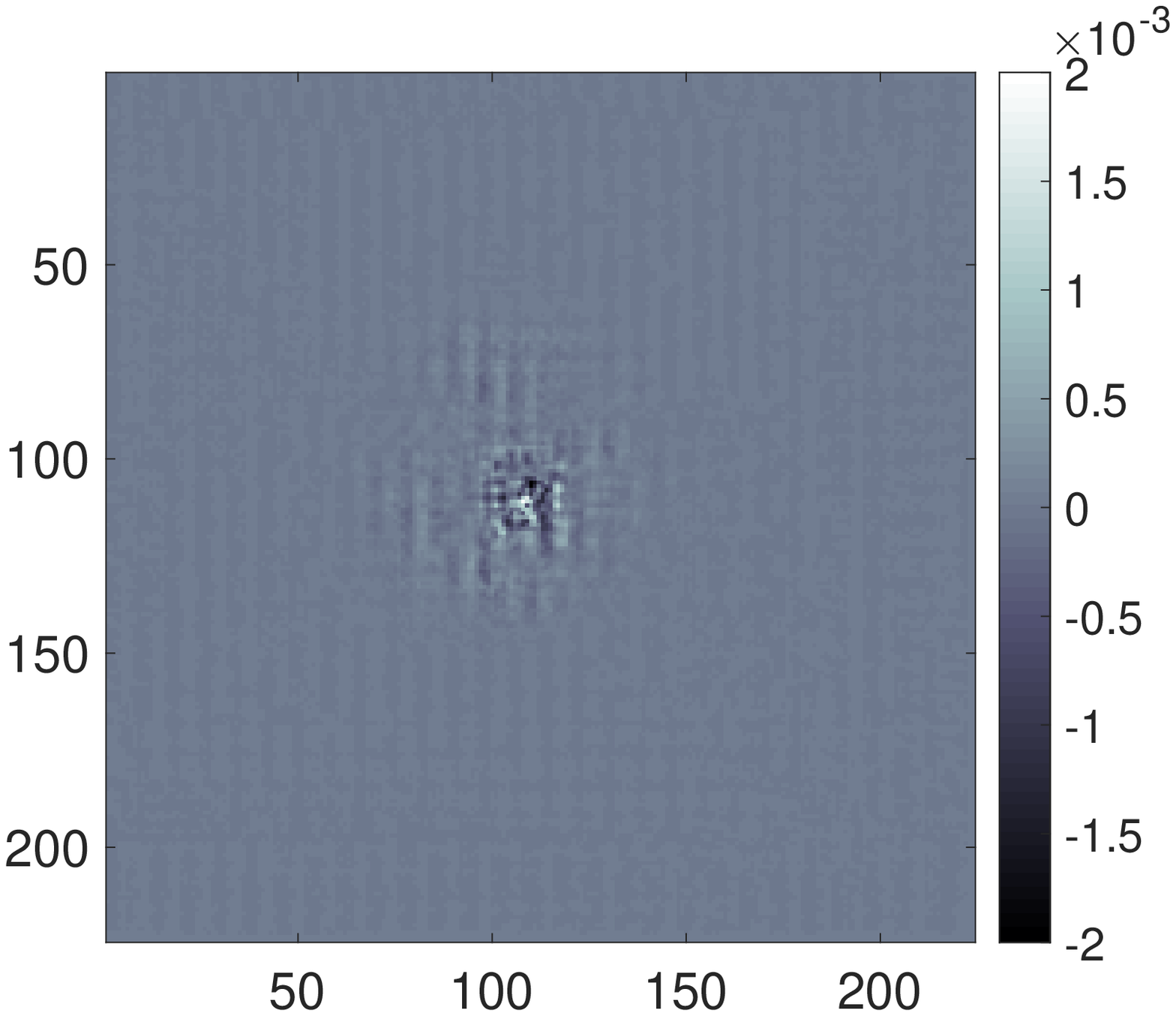}
  \includegraphics[trim={1.5cm 0cm 1.5cm 0cm},clip,width=0.47\columnwidth]{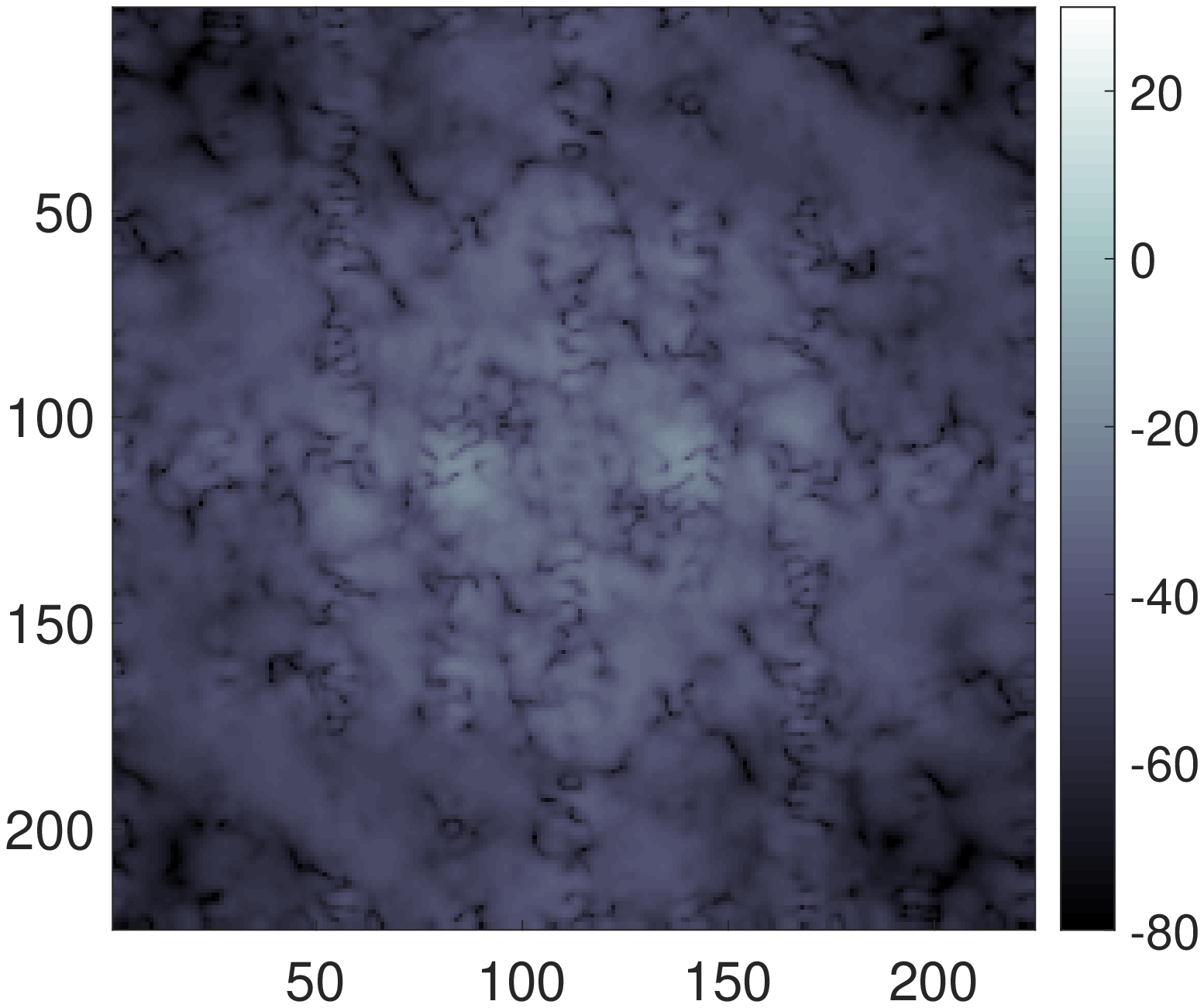} 
  }\\
  \vspace*{-0.02mm}
  \subfloat[VGG-s~\cite{chatfield14} 15.3\%]{
  \centering
  \includegraphics[trim={1.5cm 0cm 1.5cm 0cm},clip,width=0.47\columnwidth]{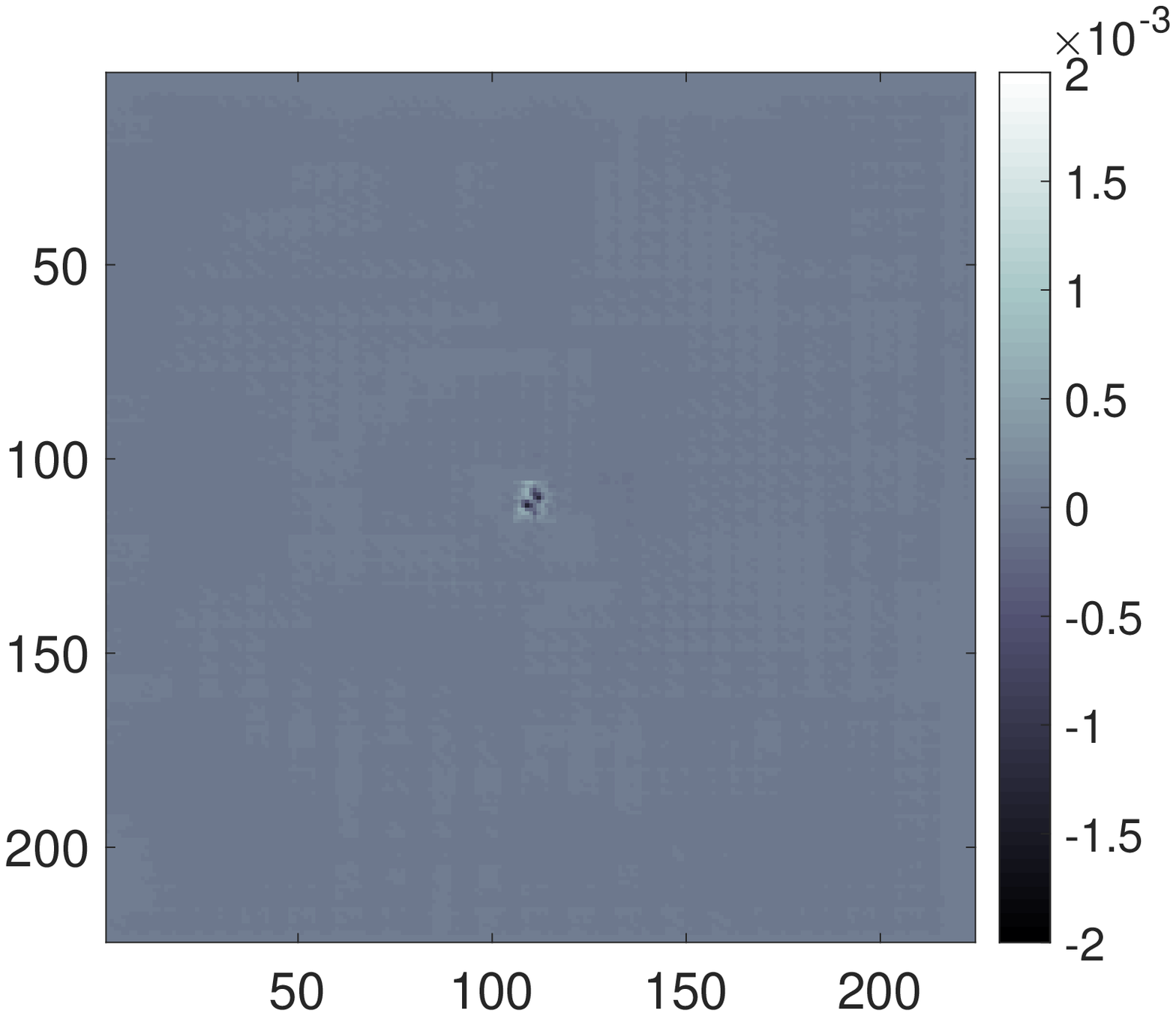}
  \includegraphics[trim={1.5cm 0cm 1.5cm 0cm},clip,width=0.47\columnwidth]{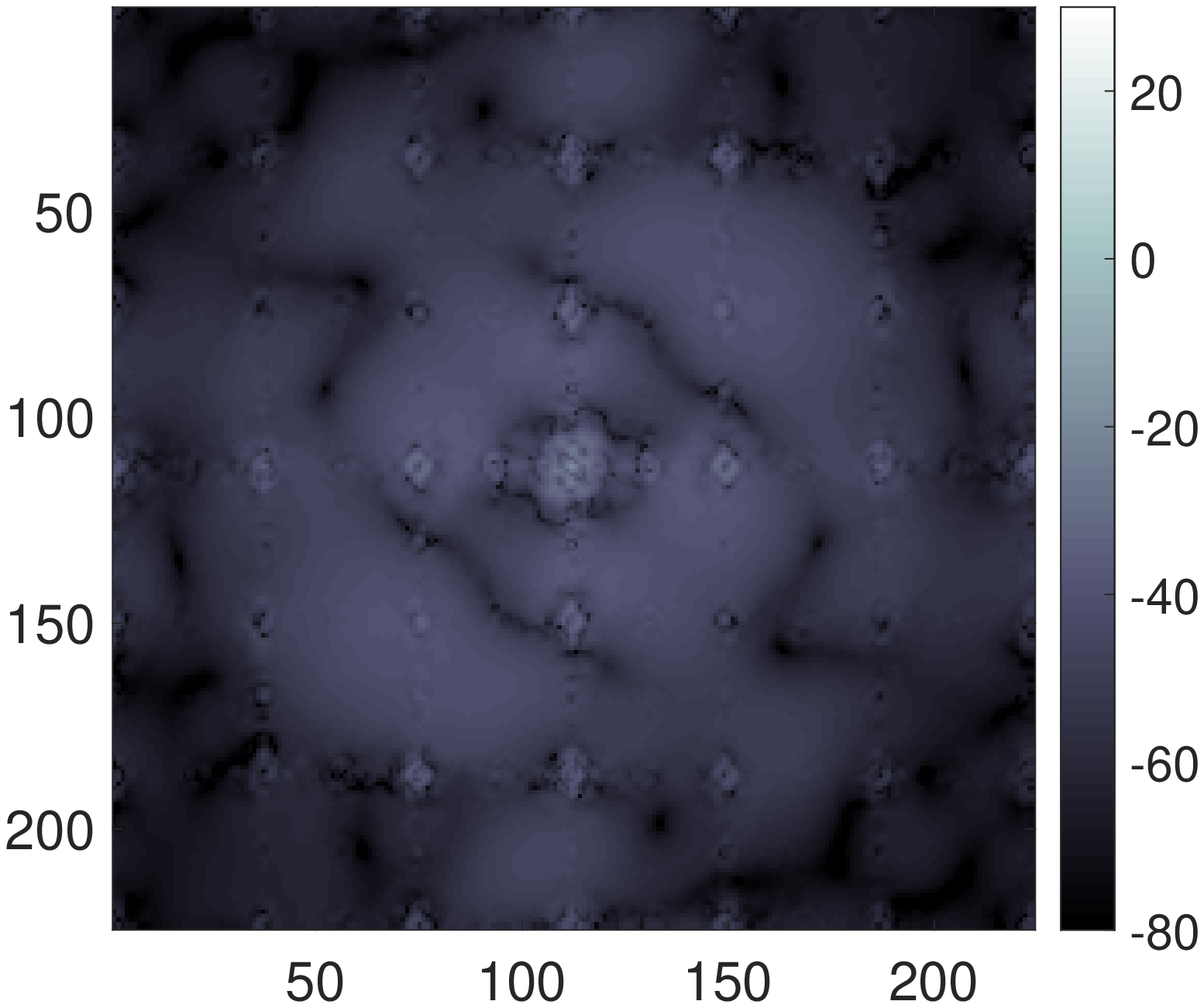} 
  }\\
  \vspace*{-0.02mm}
  \subfloat[MatConvNet-VGG-m 15.5\%]{
  \centering
  \includegraphics[trim={1.5cm 0cm 1.5cm 0cm},clip,width=0.47\columnwidth]{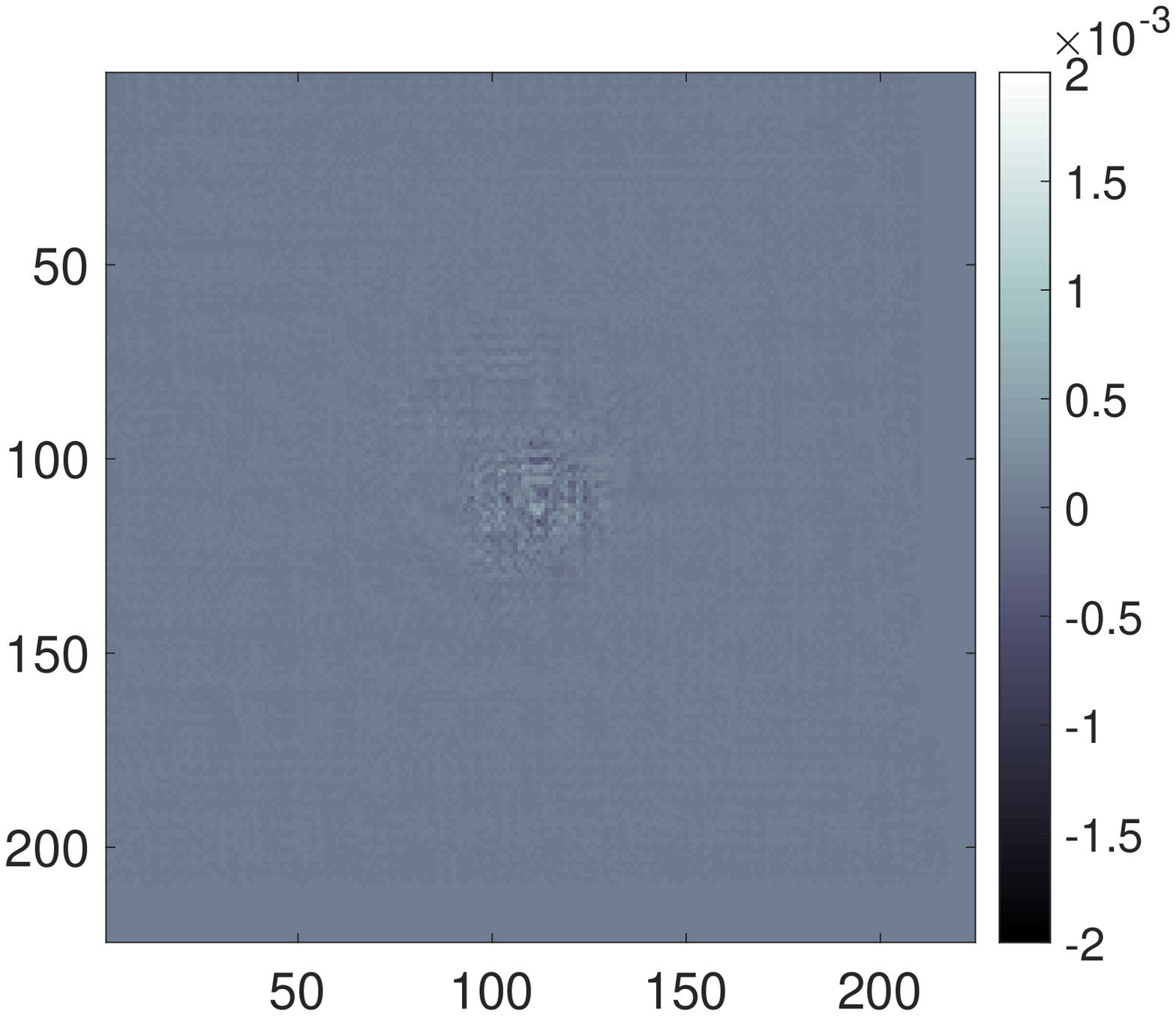}
  \includegraphics[trim={1.5cm 0cm 1.5cm 0cm},clip,width=0.47\columnwidth]{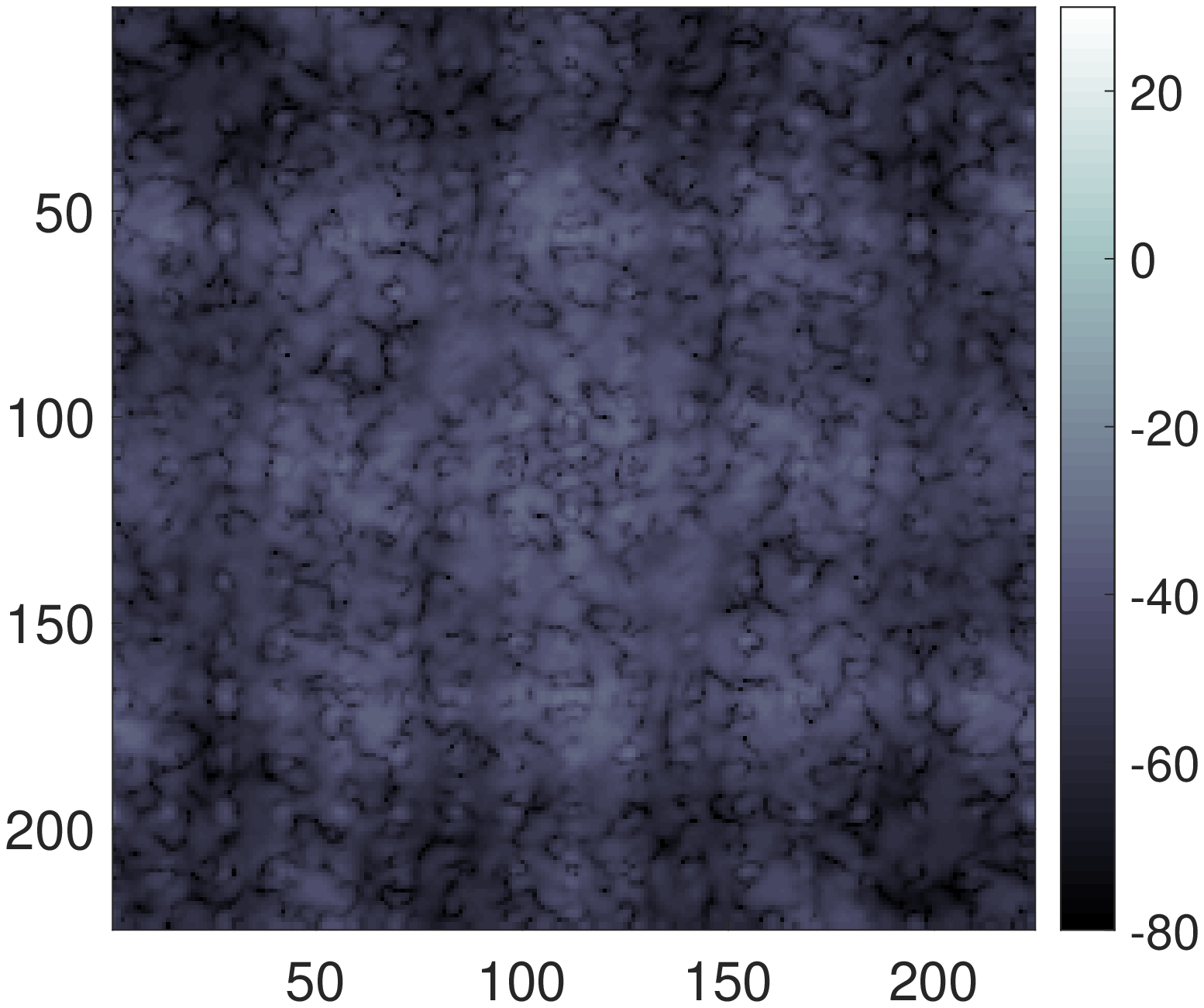} 
  }\\
\end{minipage}

\begin{minipage}[t]{0.5\linewidth}
  \subfloat[MatConvNet VGG-s 15.8\%]{
  \centering
  \includegraphics[trim={1.5cm 0cm 1.5cm 0cm},clip,width=0.47\columnwidth]{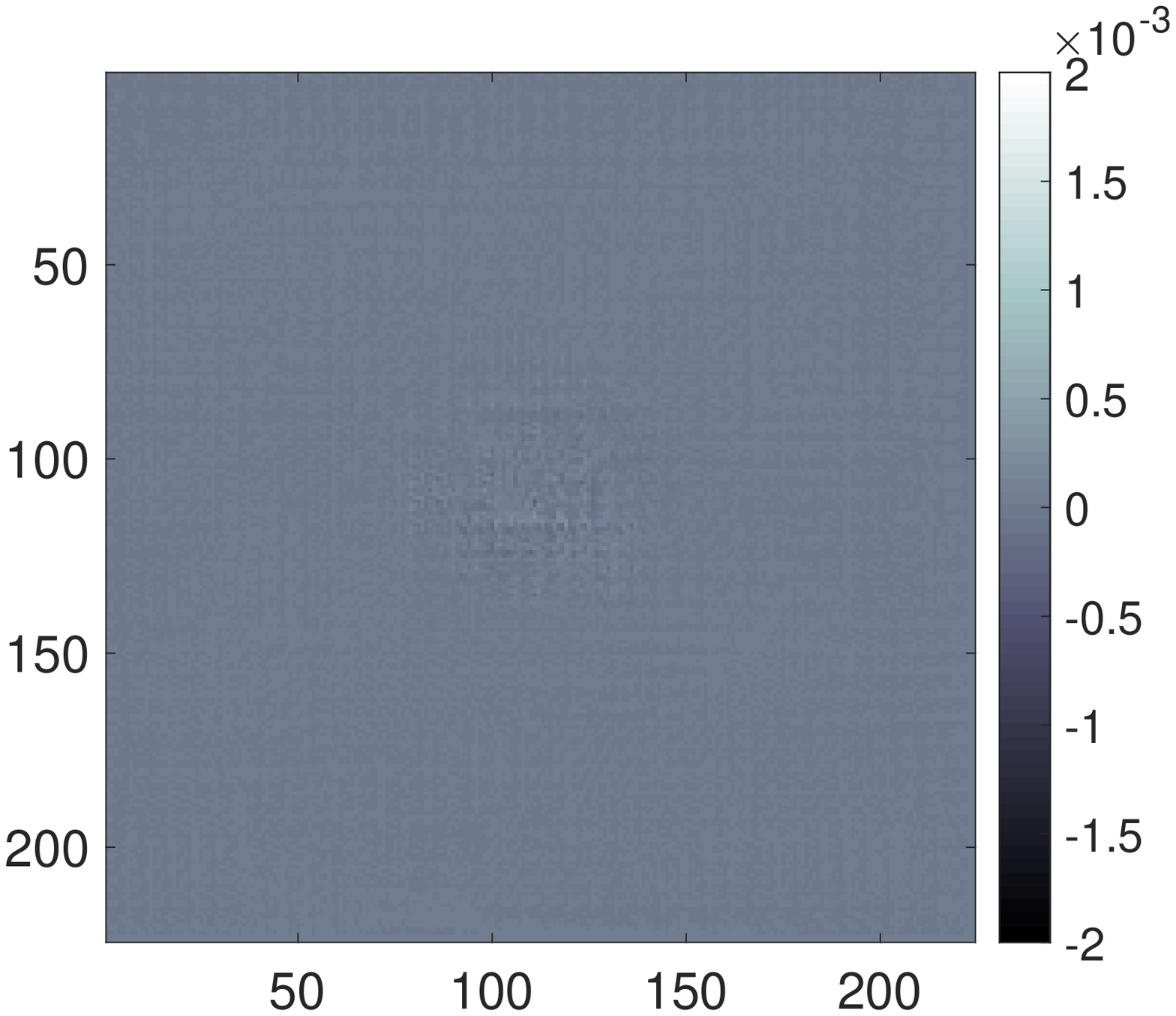}
  \includegraphics[trim={1.5cm 0cm 1.5cm 0cm},clip,width=0.47\columnwidth]{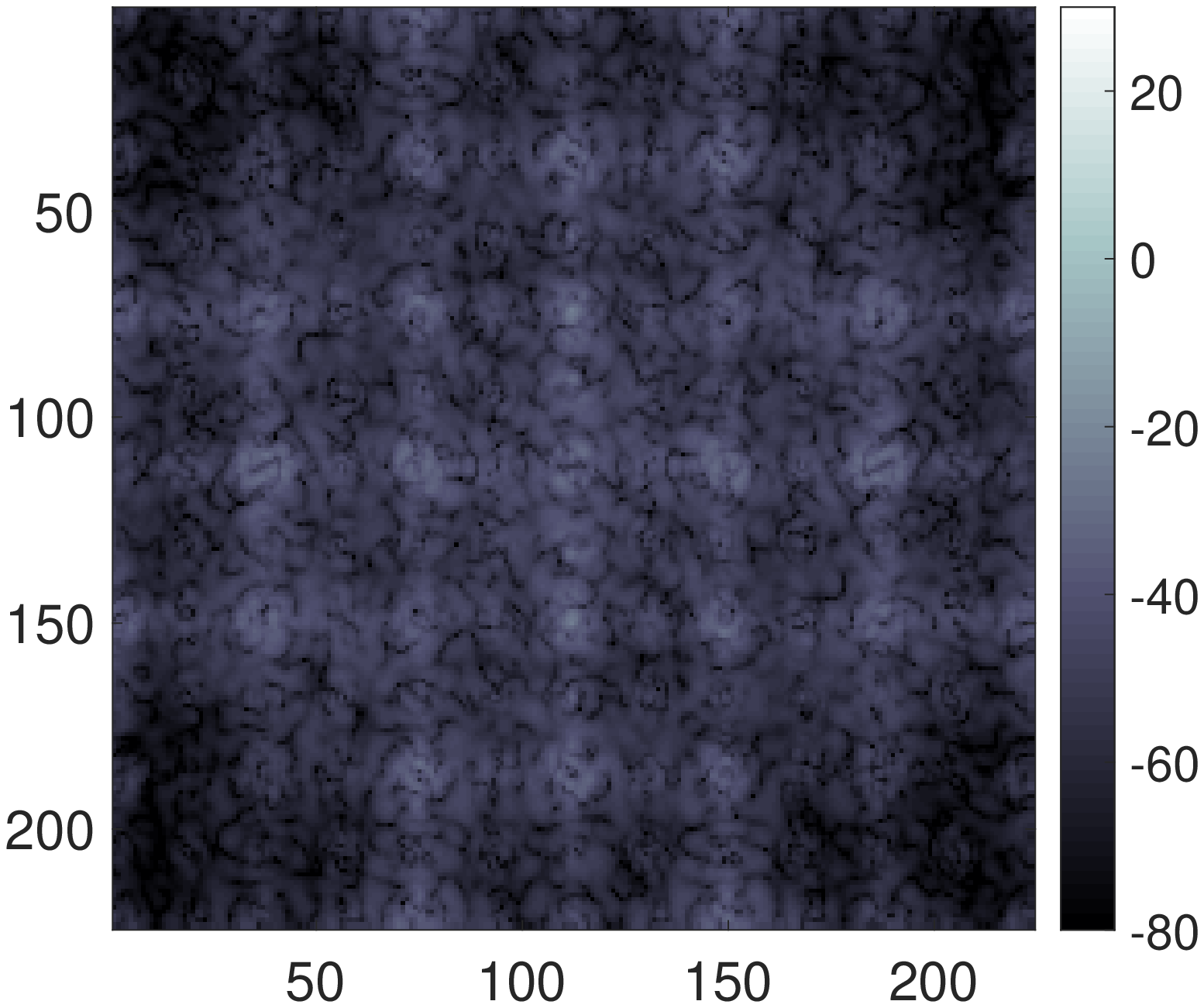}
  }\\
  \vspace*{-0.02mm}
  \subfloat[VGG-m-2048~\cite{chatfield14} 15.8\%]{
  \centering
  \includegraphics[trim={1.5cm 0cm 1.5cm 0cm},clip,width=0.47\columnwidth]{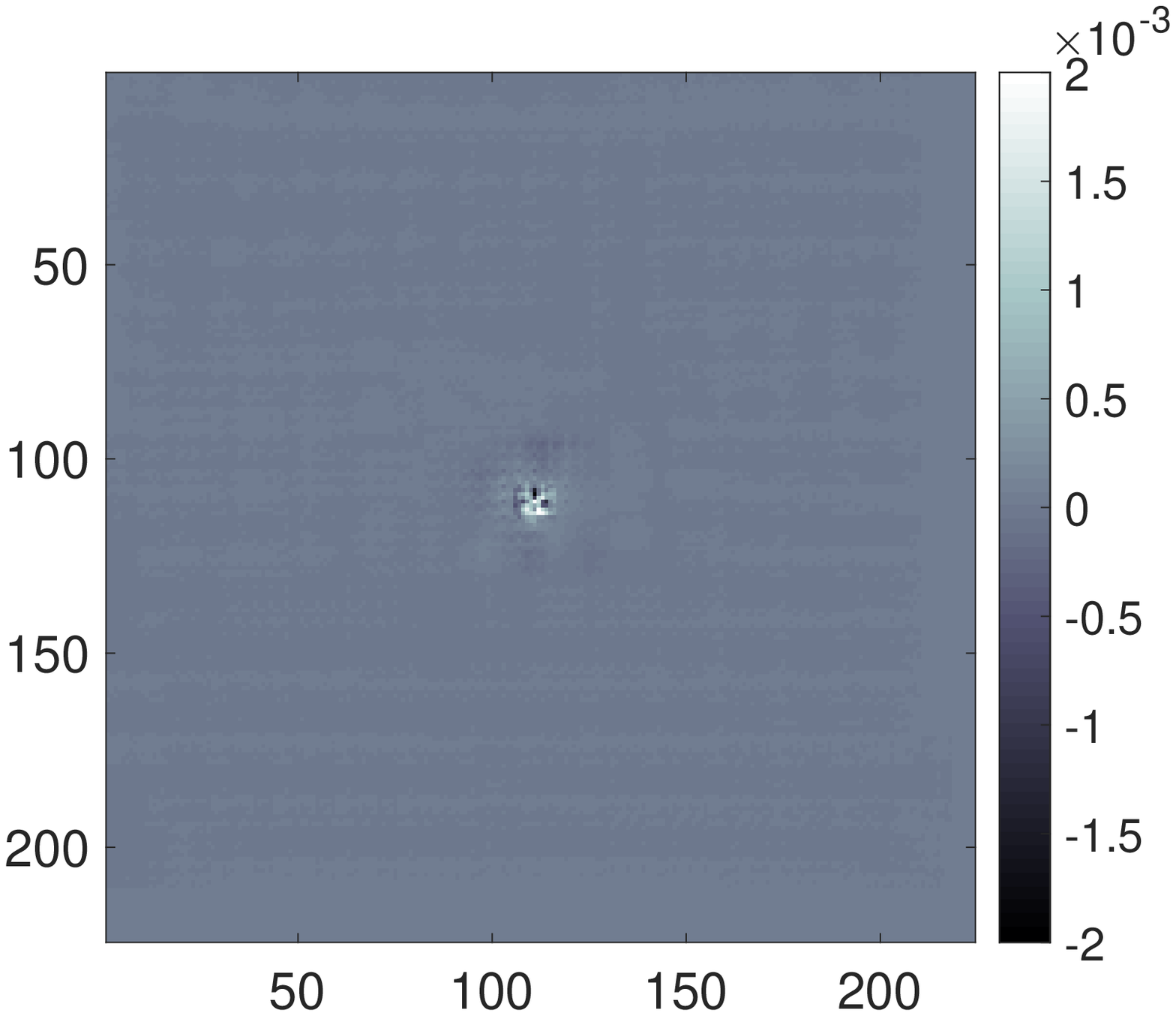}
  \includegraphics[trim={1.5cm 0cm 1.5cm 0cm},clip,width=0.47\columnwidth]{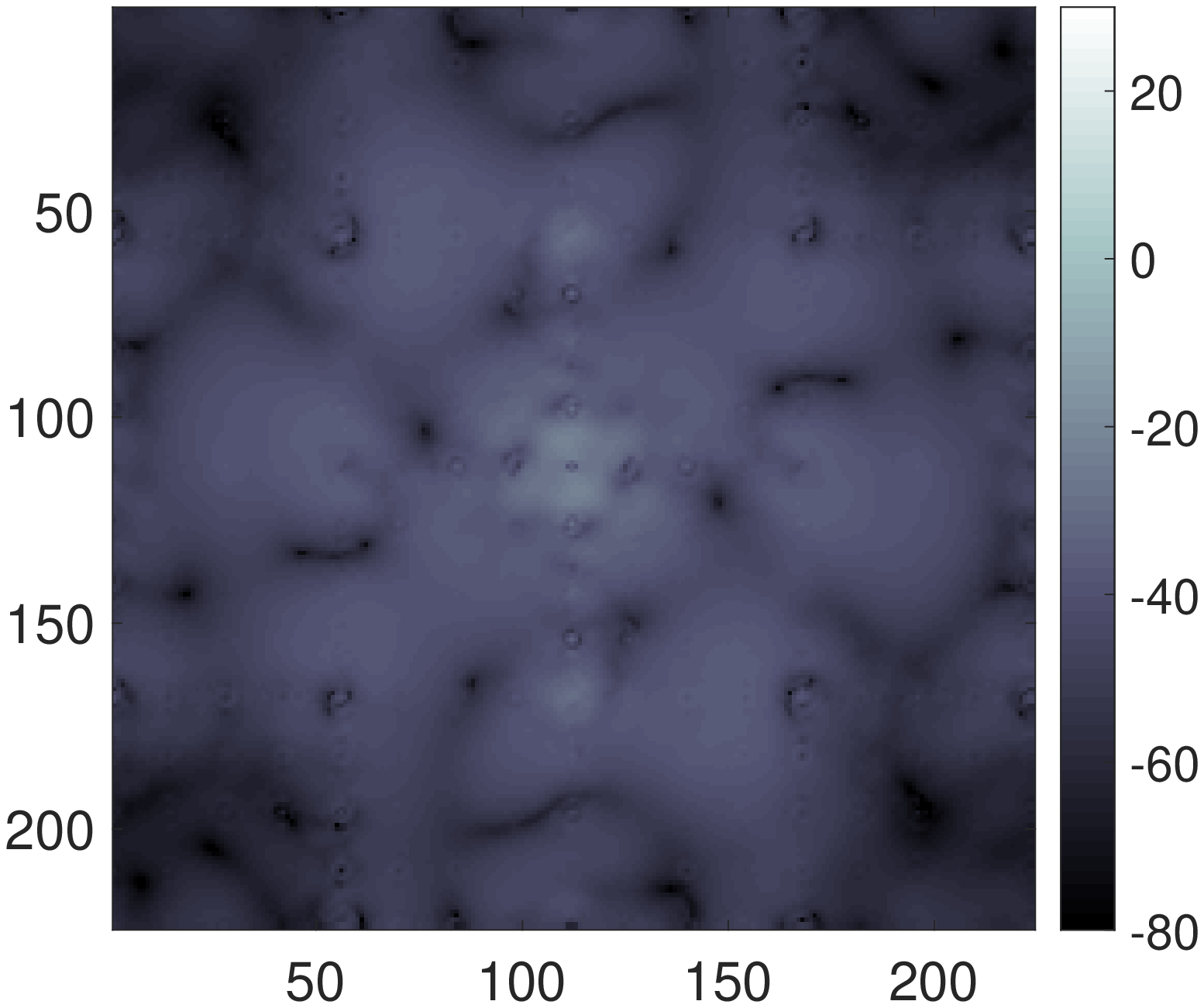}
  }\\
  \vspace*{-0.02mm}
  \subfloat[VGG-m~\cite{chatfield14} 15.9\%]{
  \centering
  \includegraphics[trim={1.5cm 0cm 1.5cm 0cm},clip,width=0.47\columnwidth]{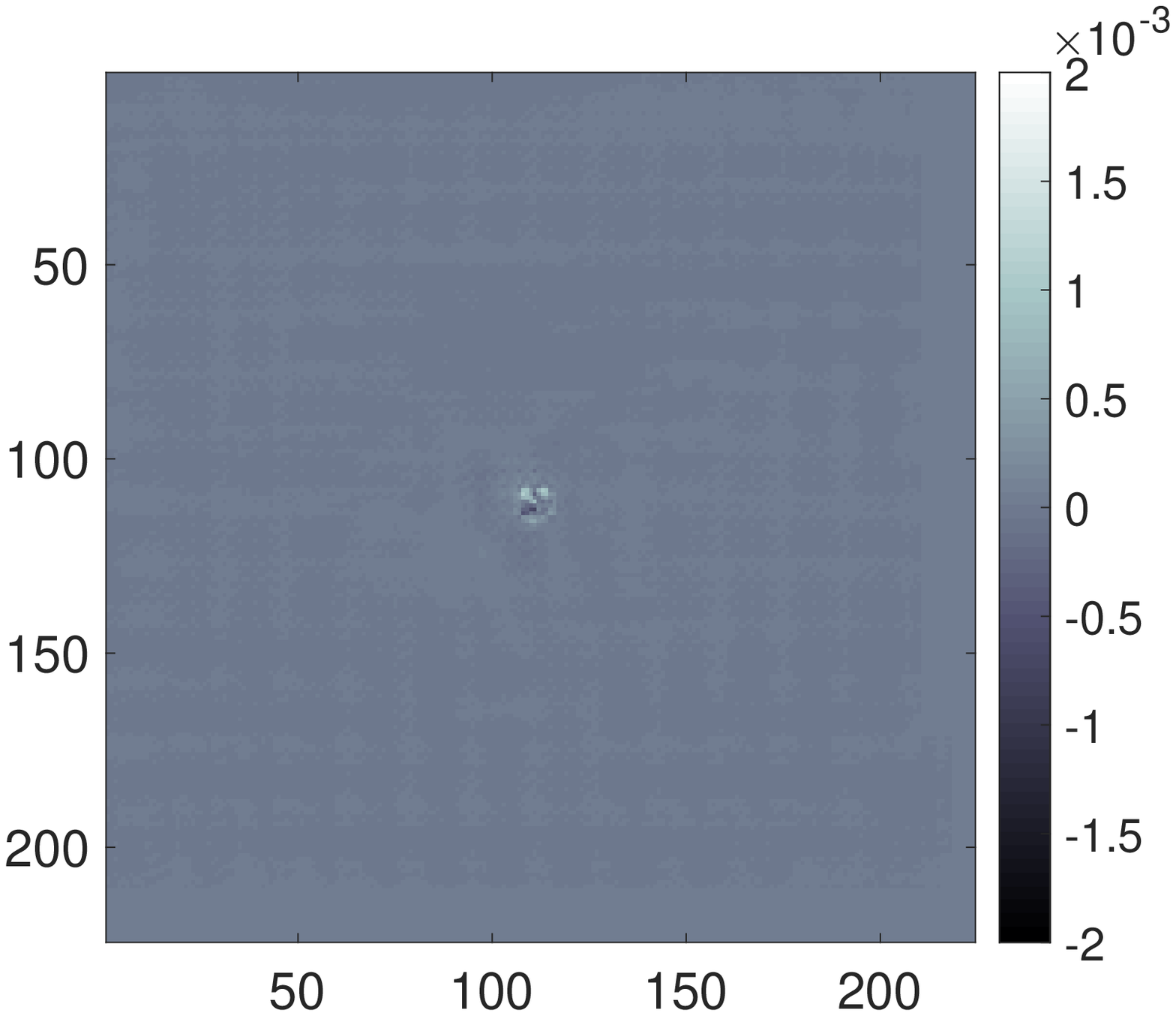}
  \includegraphics[trim={1.5cm 0cm 1.5cm 0cm},clip,width=0.47\columnwidth]{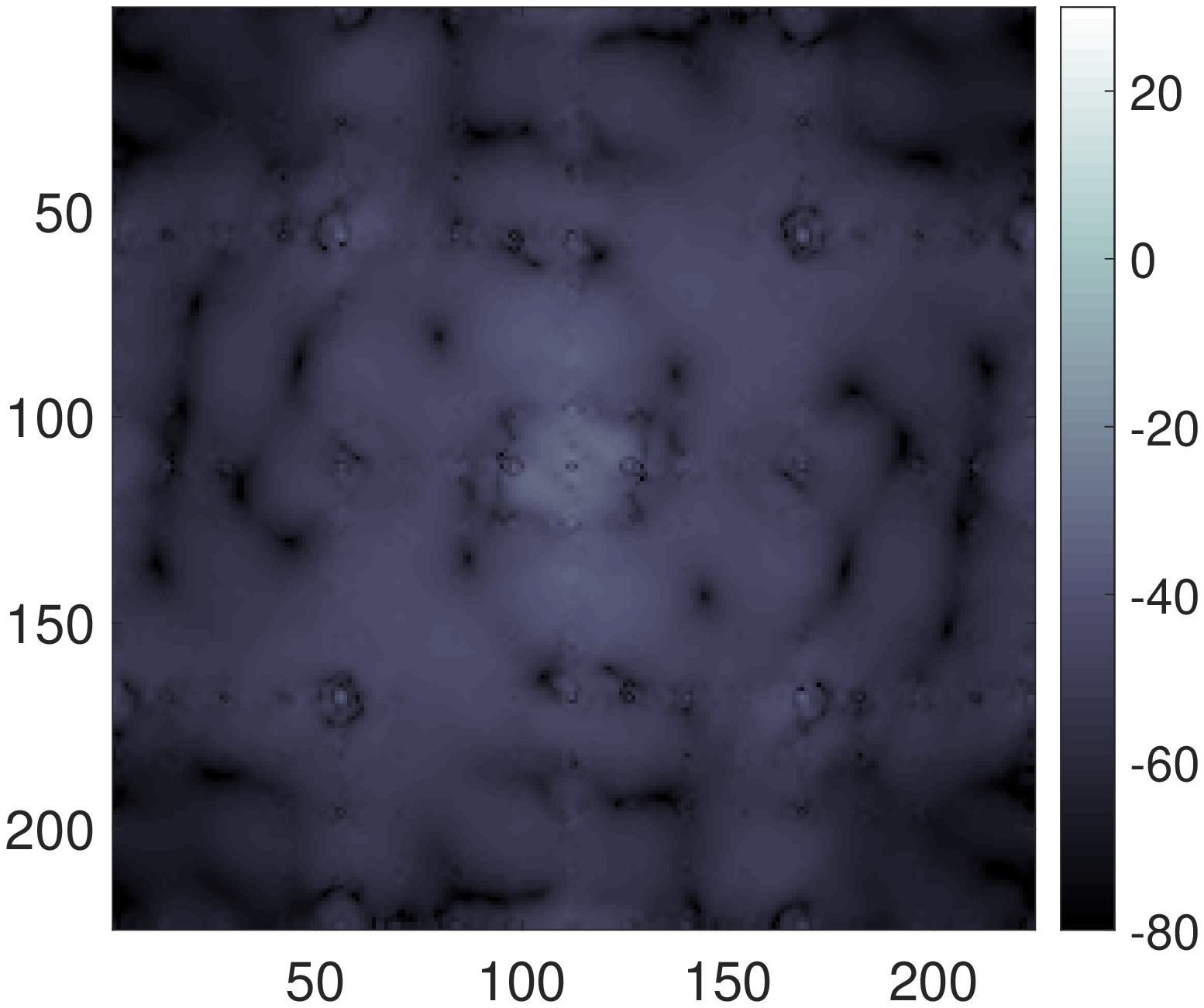} 
  }\\
  \vspace*{-0.02mm}
  \subfloat[VGG-m-1024~\cite{chatfield14} 16.1\%]{
  \centering
  \includegraphics[trim={1.5cm 0cm 1.5cm 0cm},clip,width=0.47\columnwidth]{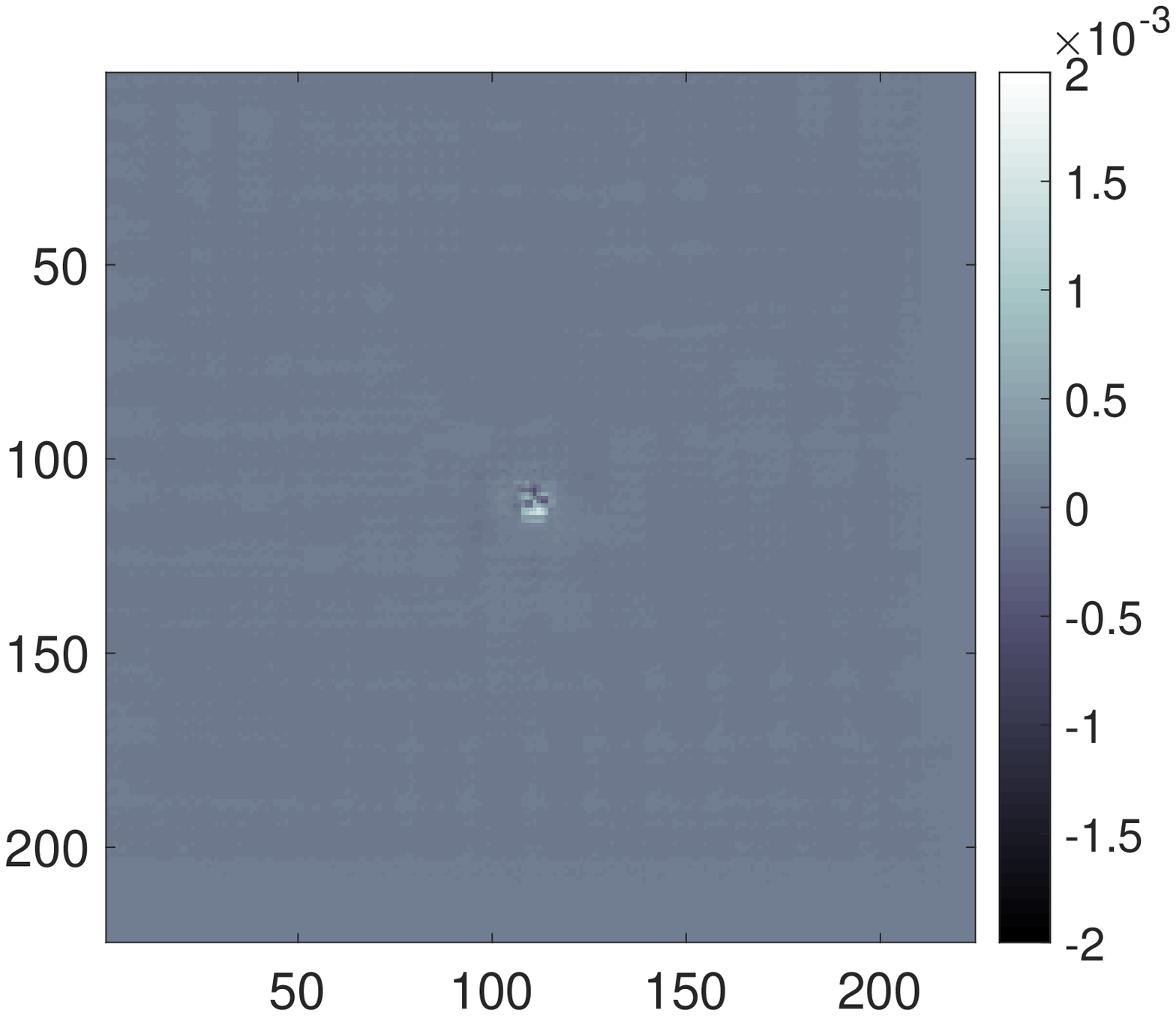}
  \includegraphics[trim={1.5cm 0cm 1.5cm 0cm},clip,width=0.47\columnwidth]{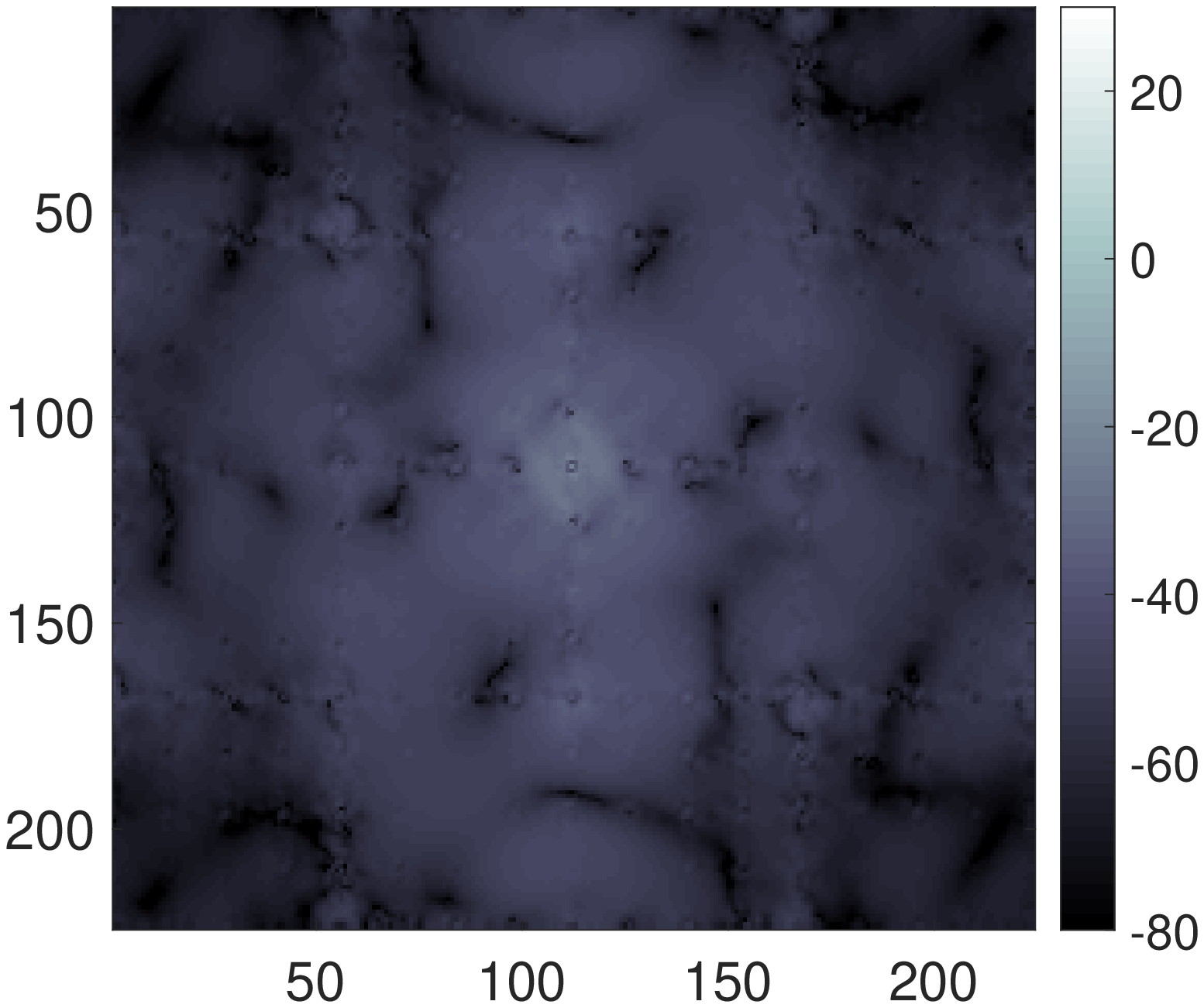} 
  }\\
  \vspace*{-0.02mm}
  \subfloat[VGG-m-128~\cite{chatfield14} 18.4\%]{
  \centering
  \includegraphics[trim={1.5cm 0cm 1.5cm 0cm},clip,width=0.47\columnwidth]{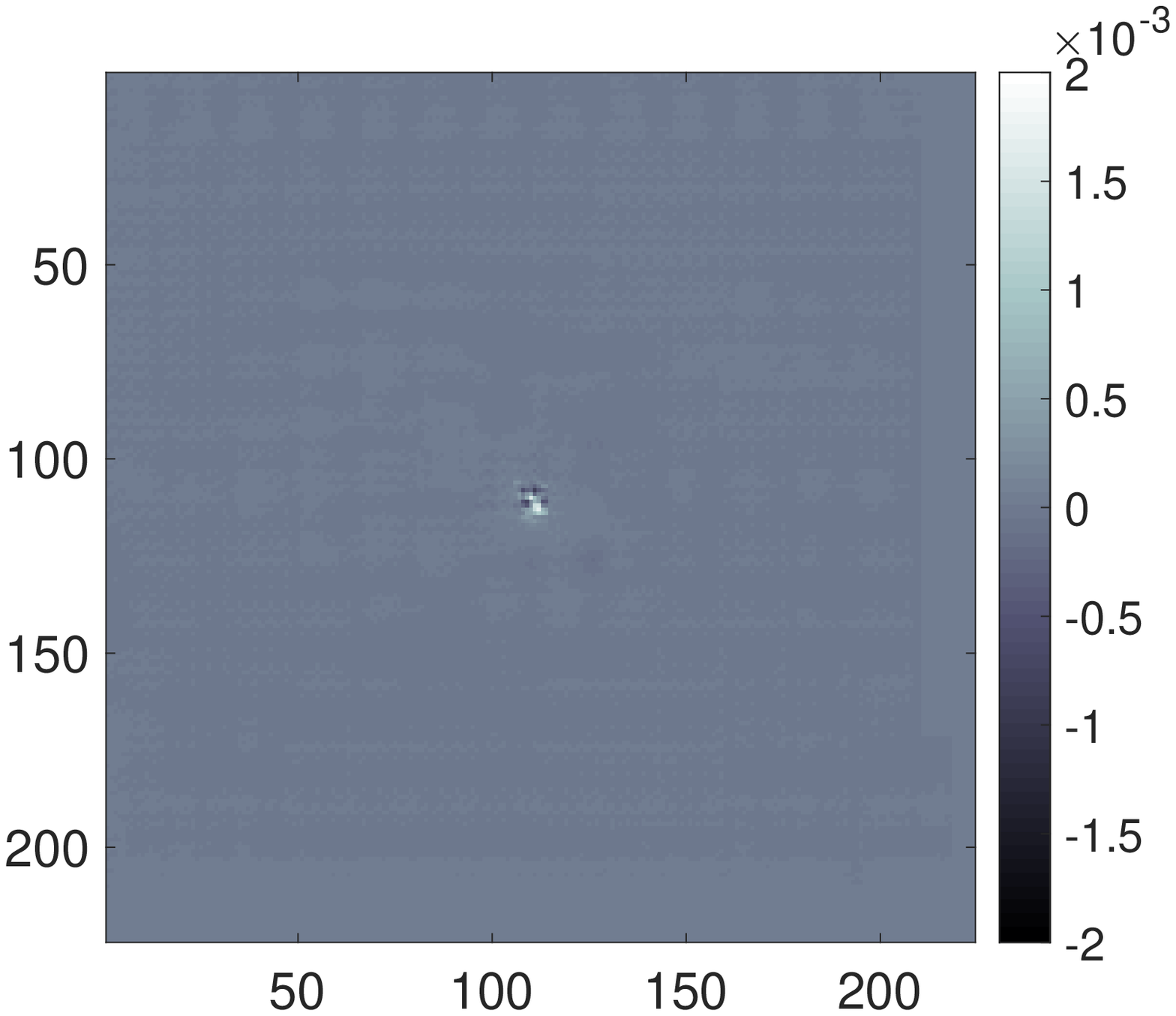}
  \includegraphics[trim={1.5cm 0cm 1.5cm 0cm},clip,width=0.47\columnwidth]{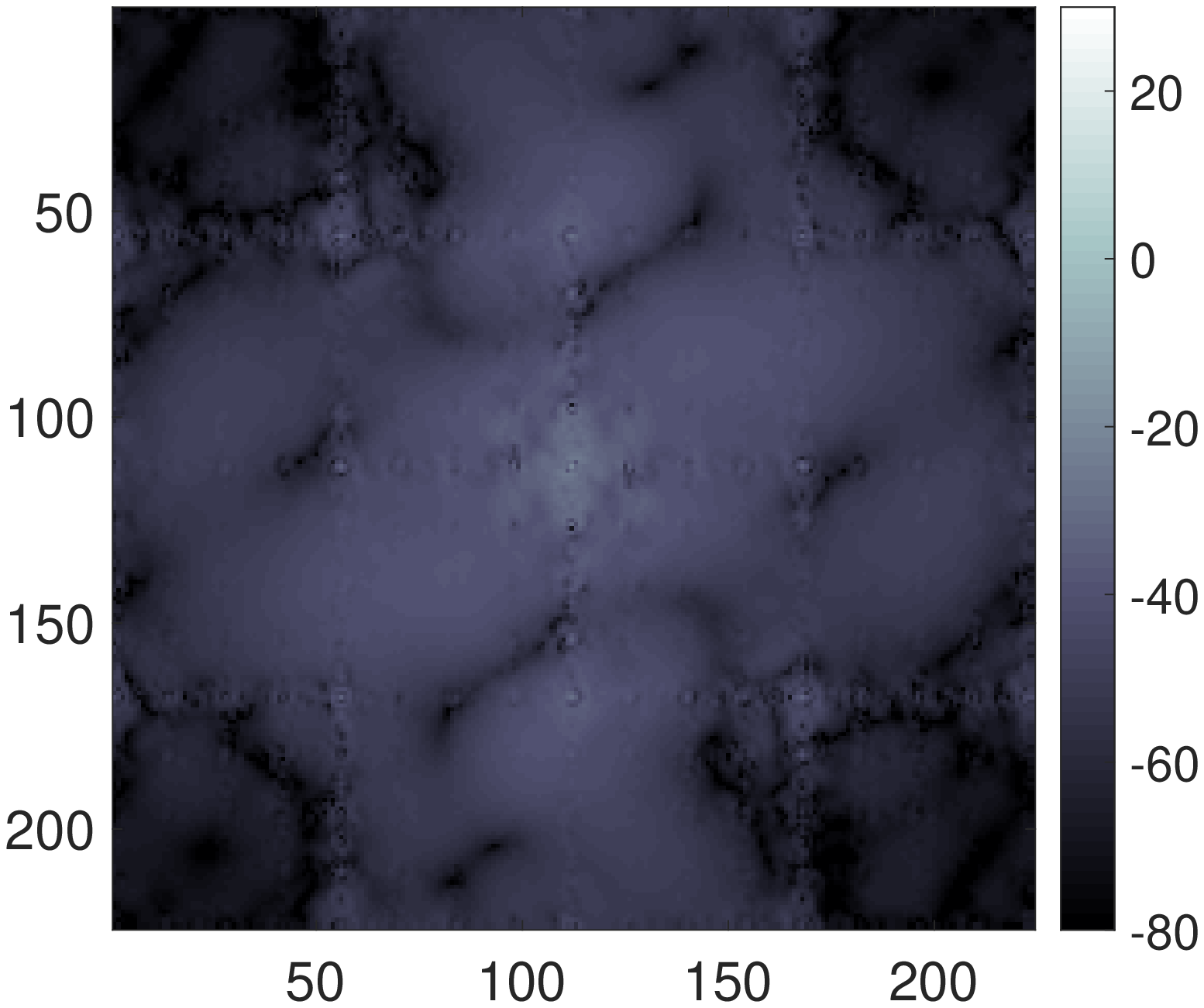} 
  }\\
  \vspace*{-0.02mm}
  \subfloat[VGG-f~\cite{chatfield14} 18.8\%]{
  \centering
  \includegraphics[trim={1.5cm 0cm 1.5cm 0cm},clip,width=0.47\columnwidth]{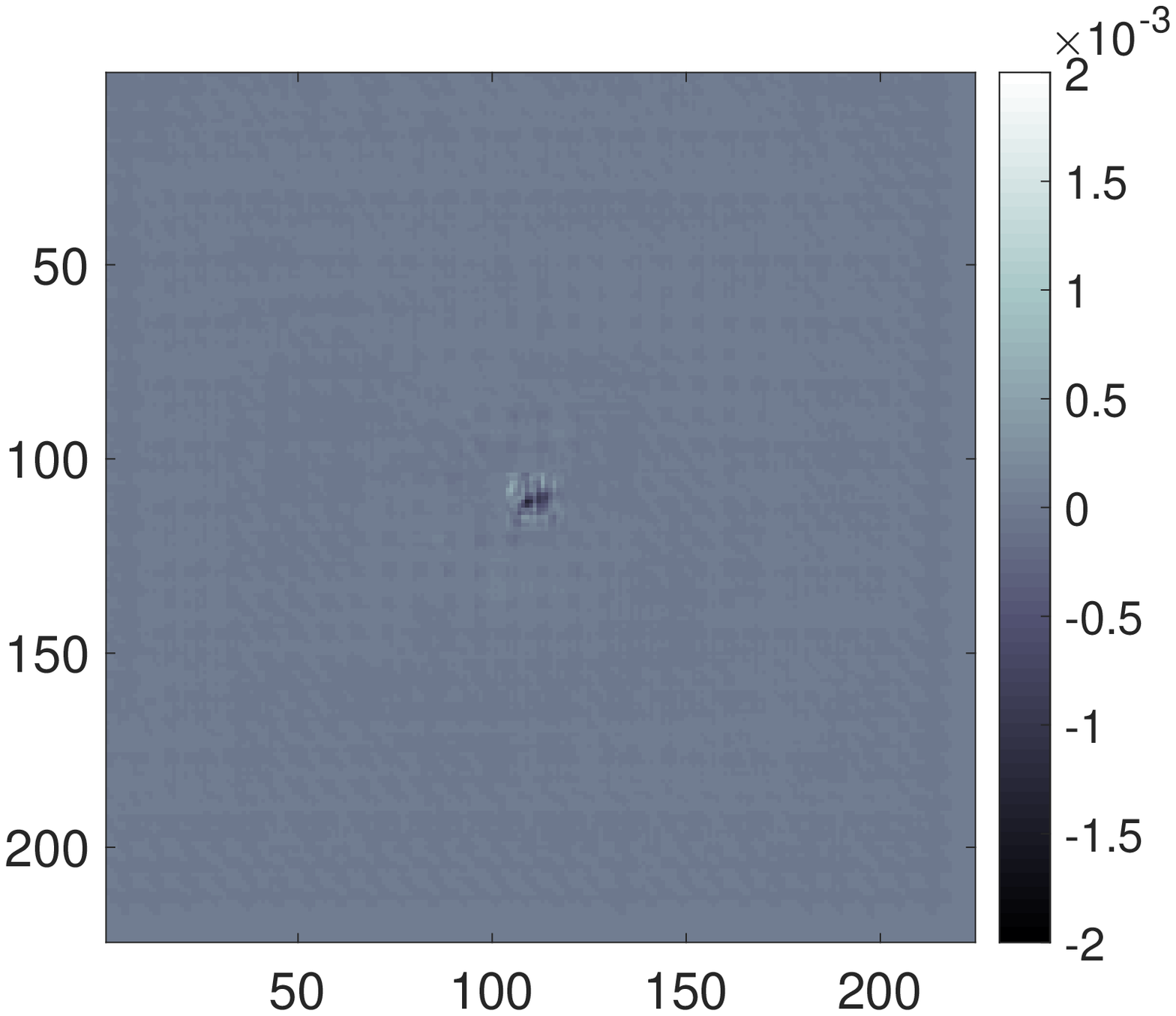}
  \includegraphics[trim={1.5cm 0cm 1.5cm 0cm},clip,width=0.47\columnwidth]{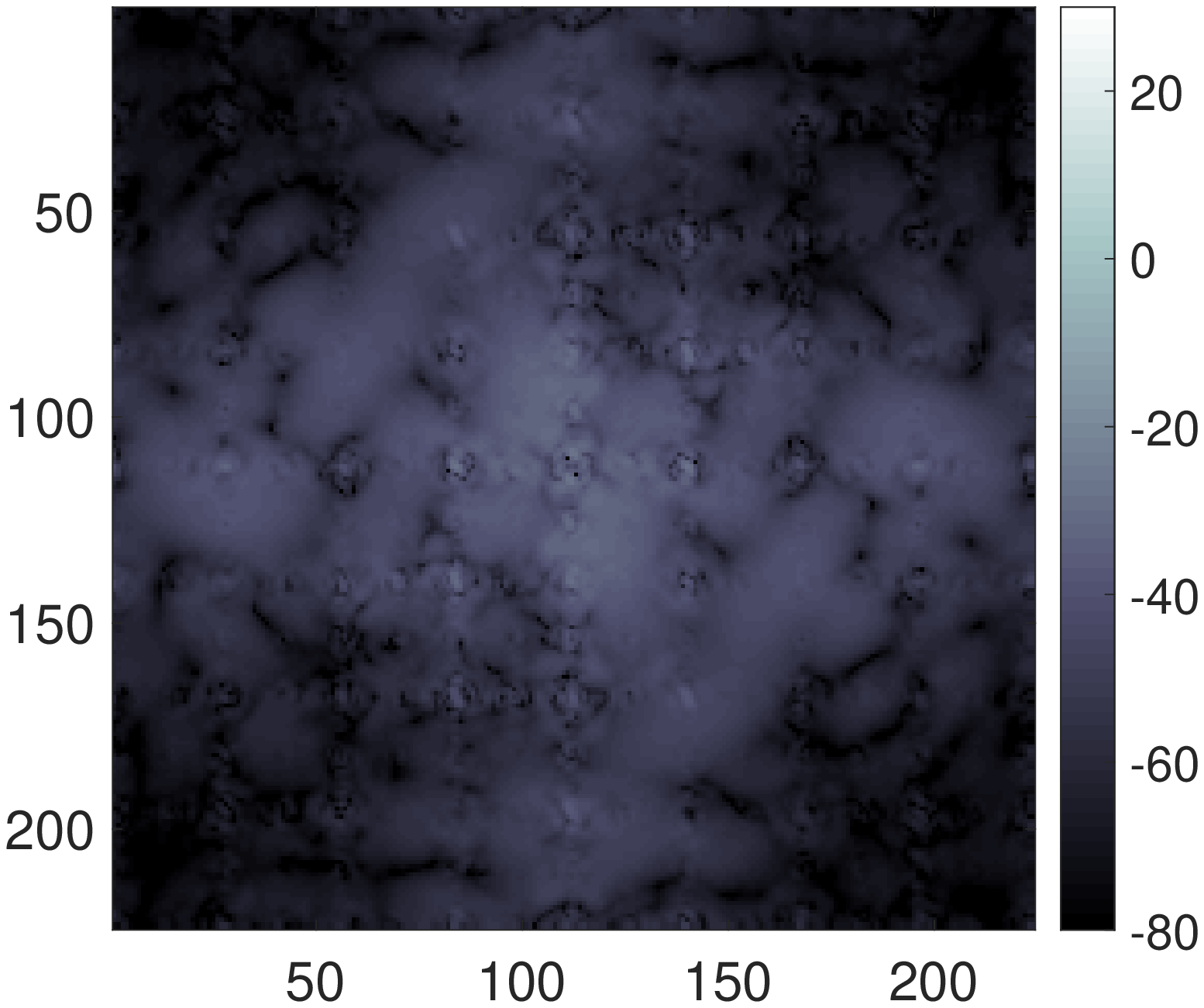} 
  }\\
  \vspace*{-0.02mm}
  \subfloat[MatConvNet-VGG-f 19.1\%]{
  \centering
  \includegraphics[trim={1.5cm 0cm 1.5cm 0cm},clip,width=0.47\columnwidth]{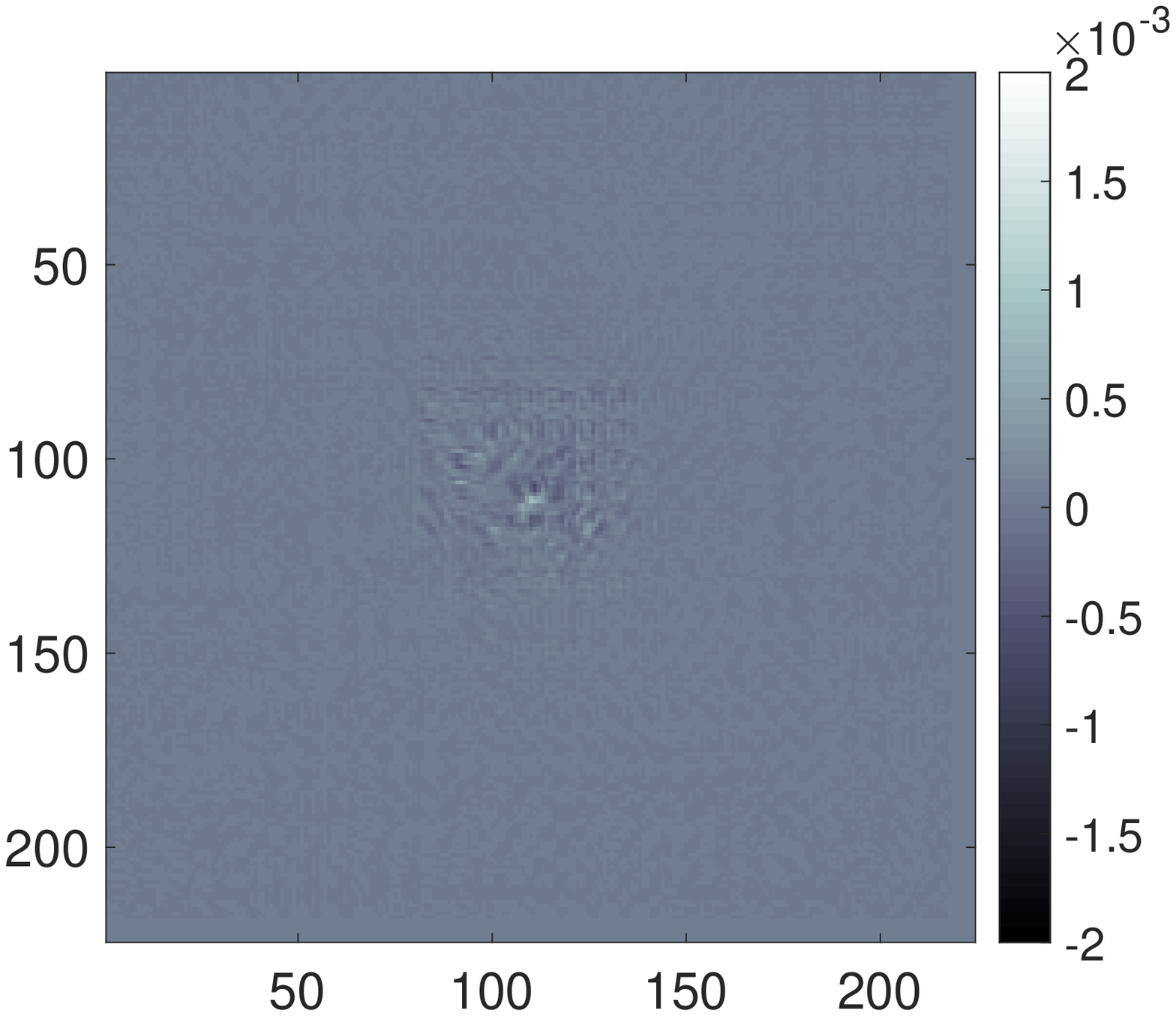}
  \includegraphics[trim={1.5cm 0cm 1.5cm 0cm},clip,width=0.47\columnwidth]{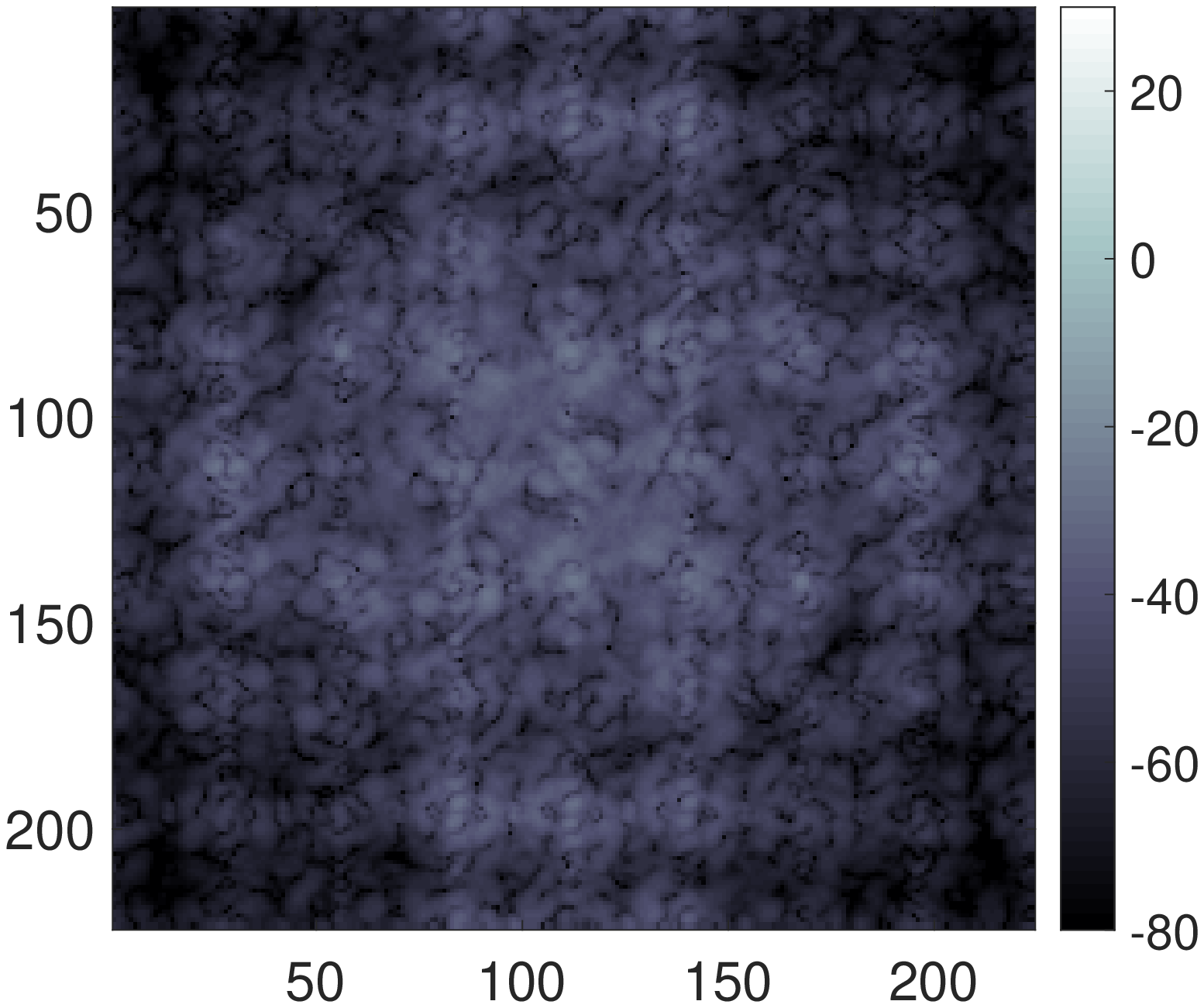} 
  }\\
  \vspace*{-0.02mm}
  \subfloat[MatConvNet-Alex 19.2\%]{
  \centering
  \includegraphics[trim={1.5cm 0cm 1.5cm 0cm},clip,width=0.47\columnwidth]{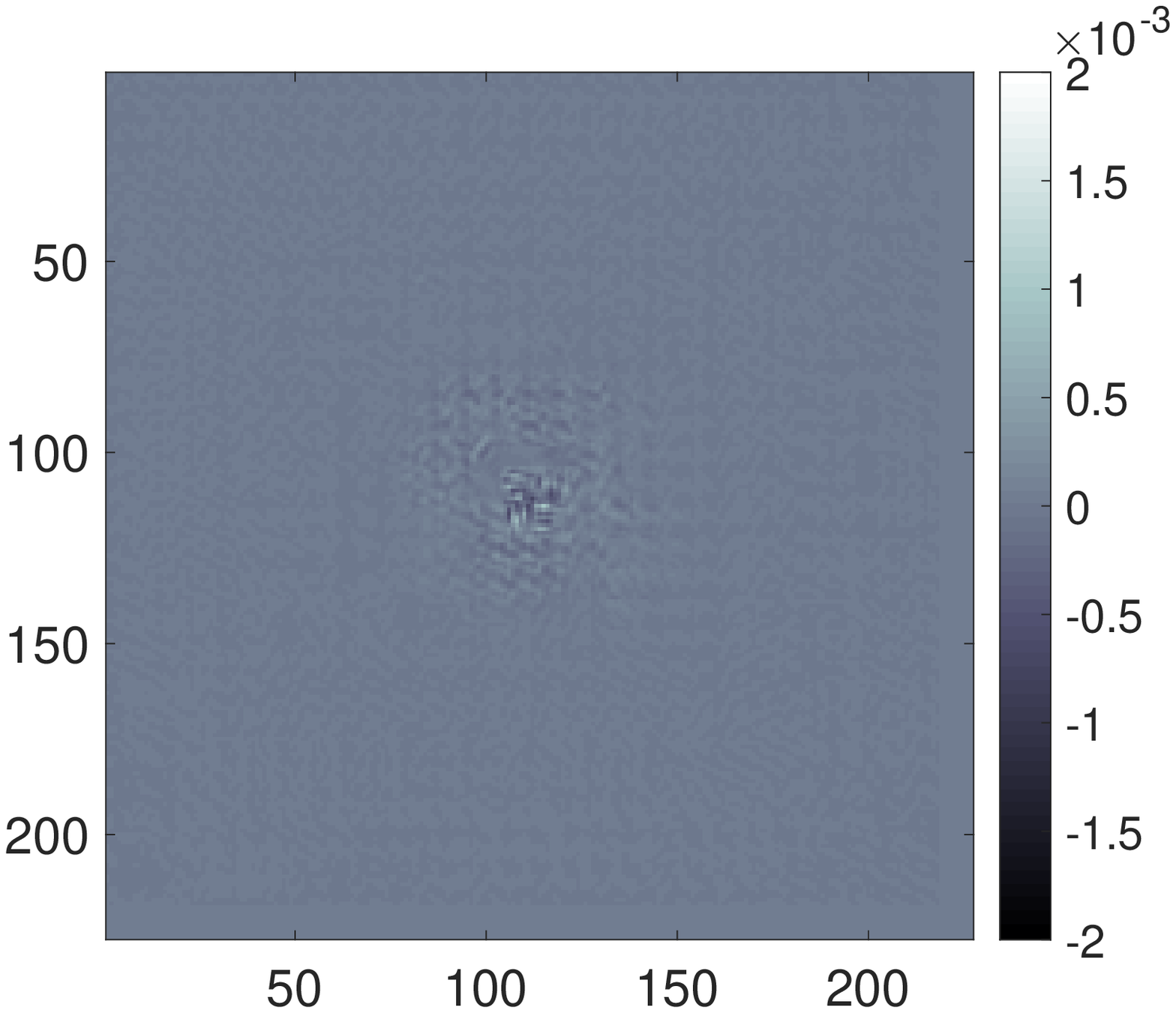}
  \includegraphics[trim={1.5cm 0cm 1.5cm 0cm},clip,width=0.47\columnwidth]{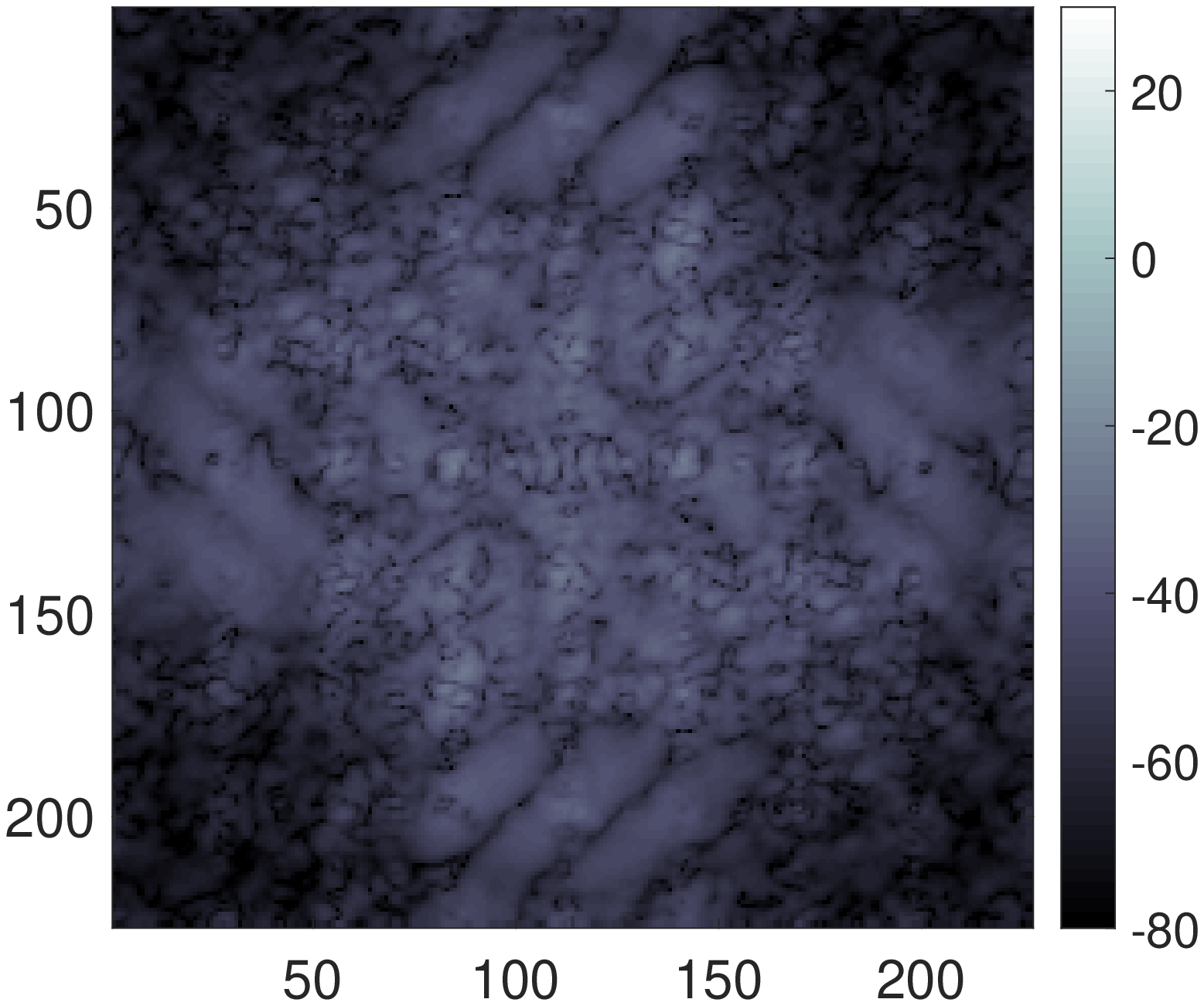} 
  }\\
  \vspace*{-0.02mm}
  \subfloat[Caffe-Alex~\cite{krizhevsky2012imagenet} 19.6\%]{
  \centering
  \includegraphics[trim={1.5cm 0cm 1.5cm 0cm},clip,width=0.47\columnwidth]{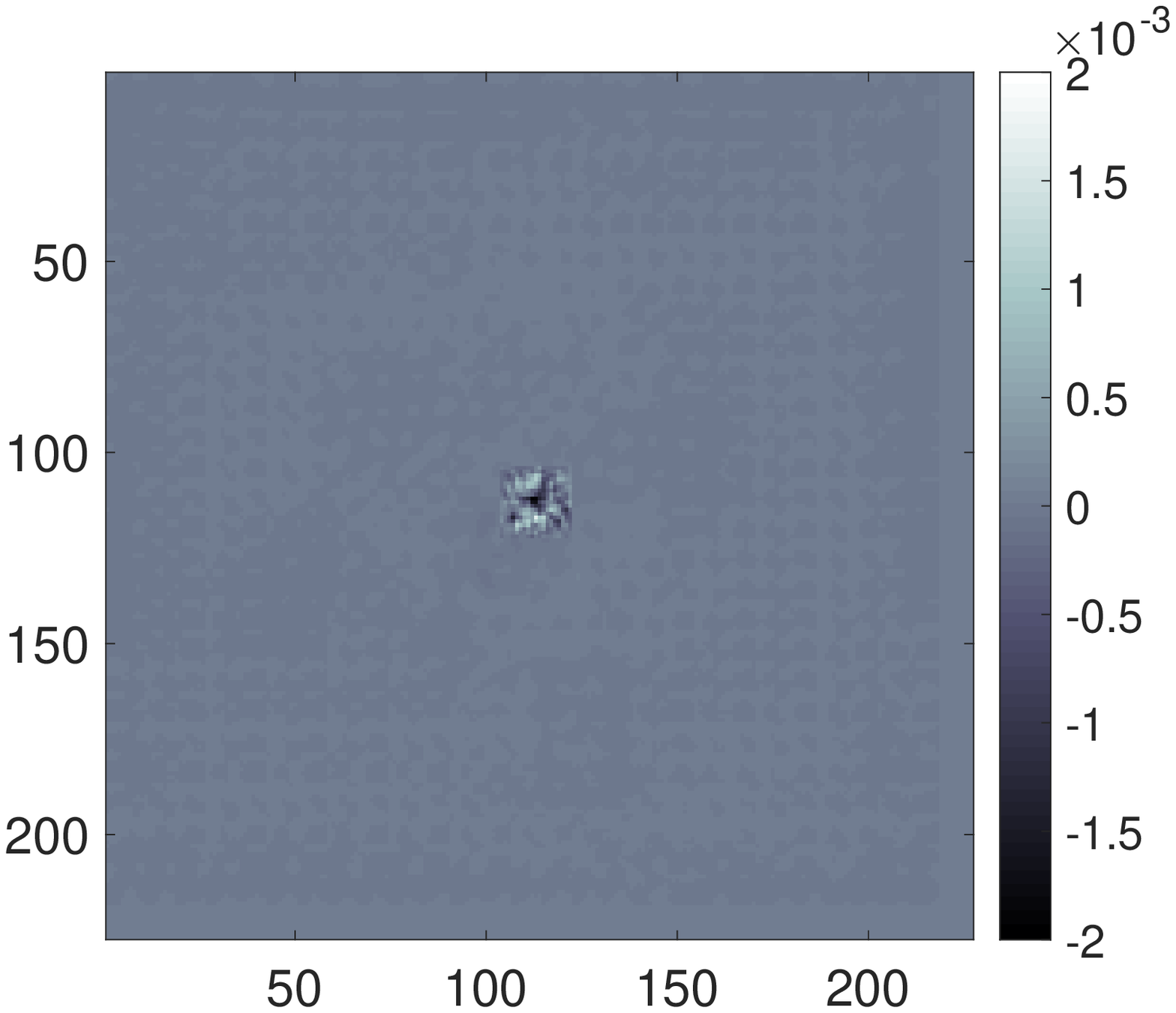}
  \includegraphics[trim={1.5cm 0cm 1.5cm 0cm},clip,width=0.47\columnwidth]{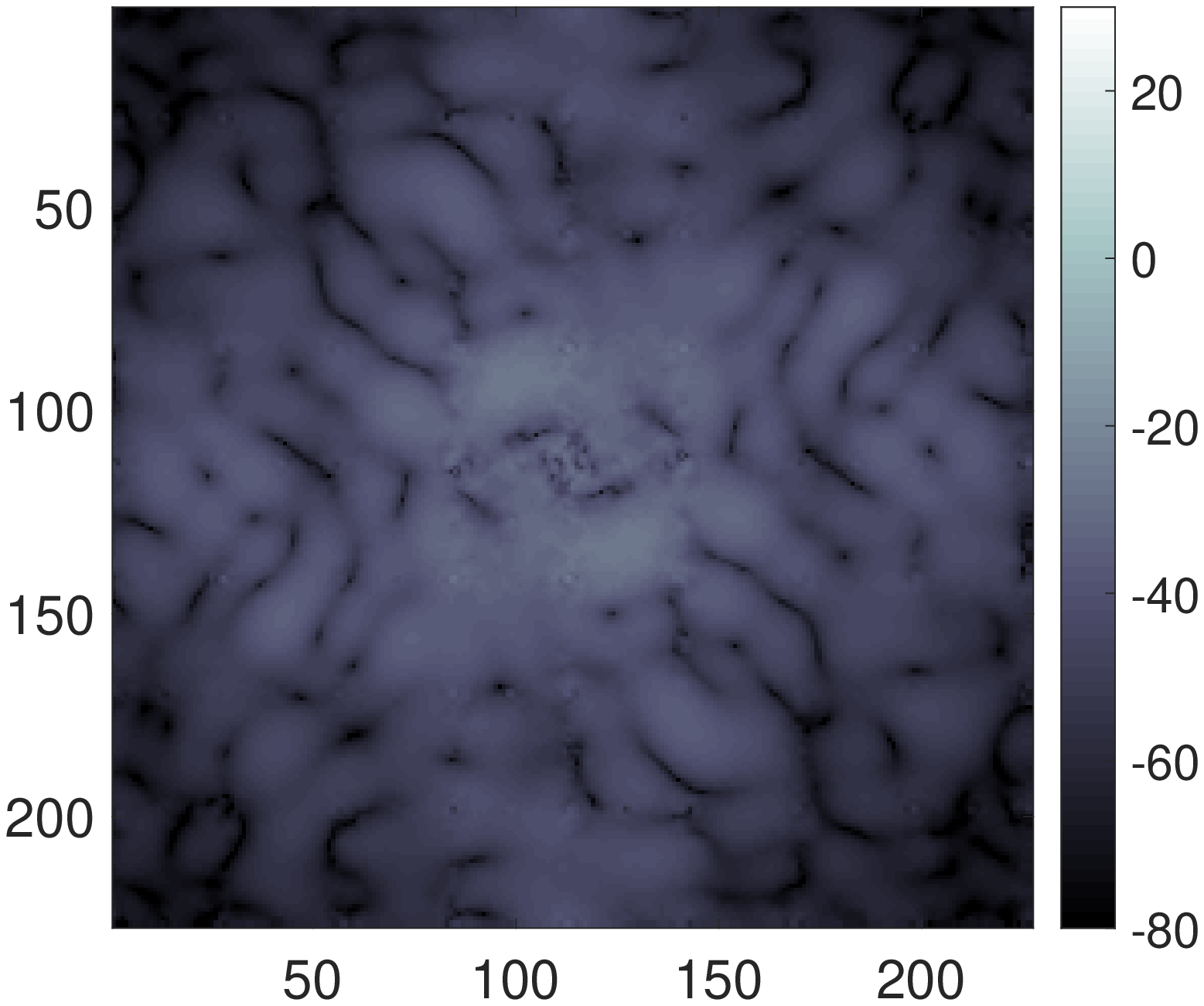} 
  }\\
\end{minipage}
}
\adjustbox{valign=t}{
\begin{minipage}[t]{1.\linewidth}
\caption{Data derivative $dz/dx$ (left) and frequency response (right) of $18$ CNNs pre-trained on ImageNet data, where $x$ was an impulse image in the red channel. The top-5 error rate on the ILSVRC 2012 validation data is listed for each model.}
\label{model_zoo}
\end{minipage}}
\end{figure}

\begin{figure*}
\centering
\subfloat[Red Channel]{\includegraphics[trim={0cm 0cm 0cm 0cm},clip,width=0.65\columnwidth]{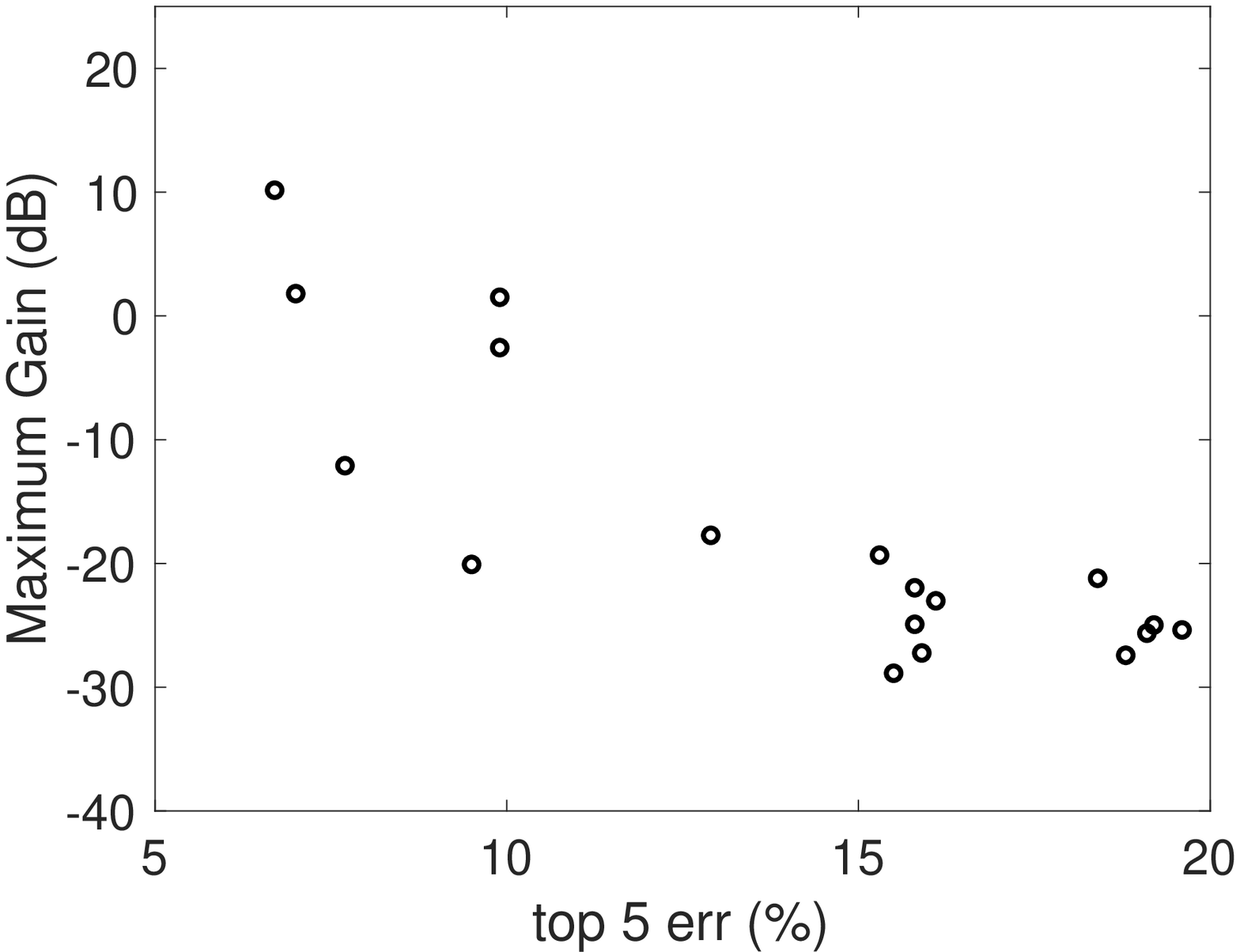}}
\subfloat[Green Channel]{\includegraphics[trim={0cm 0cm 0cm 0cm},clip,width=0.65\columnwidth]{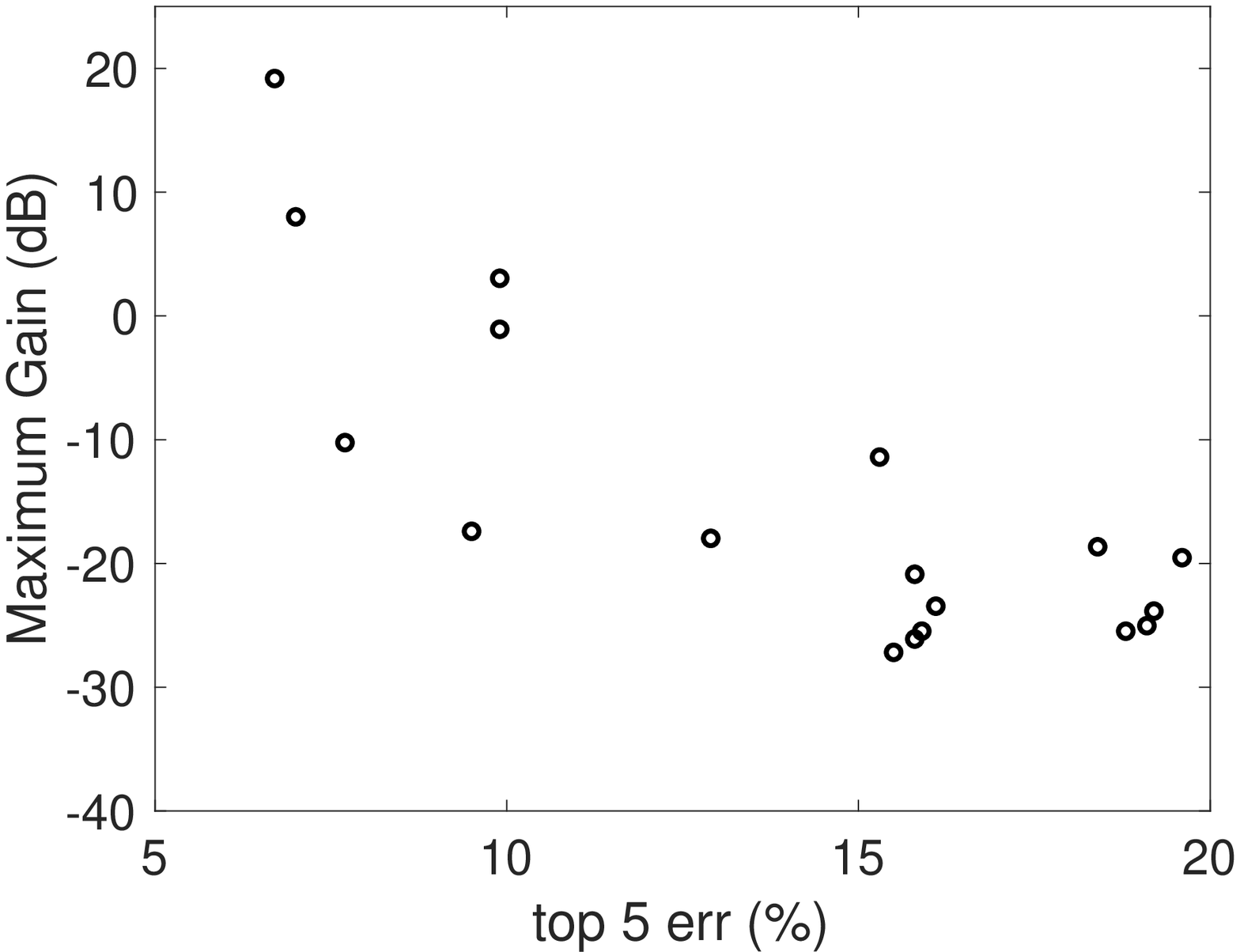}}
\subfloat[Blue Channel]{\includegraphics[trim={0cm 0cm 0cm 0cm},clip,width=0.65\columnwidth]{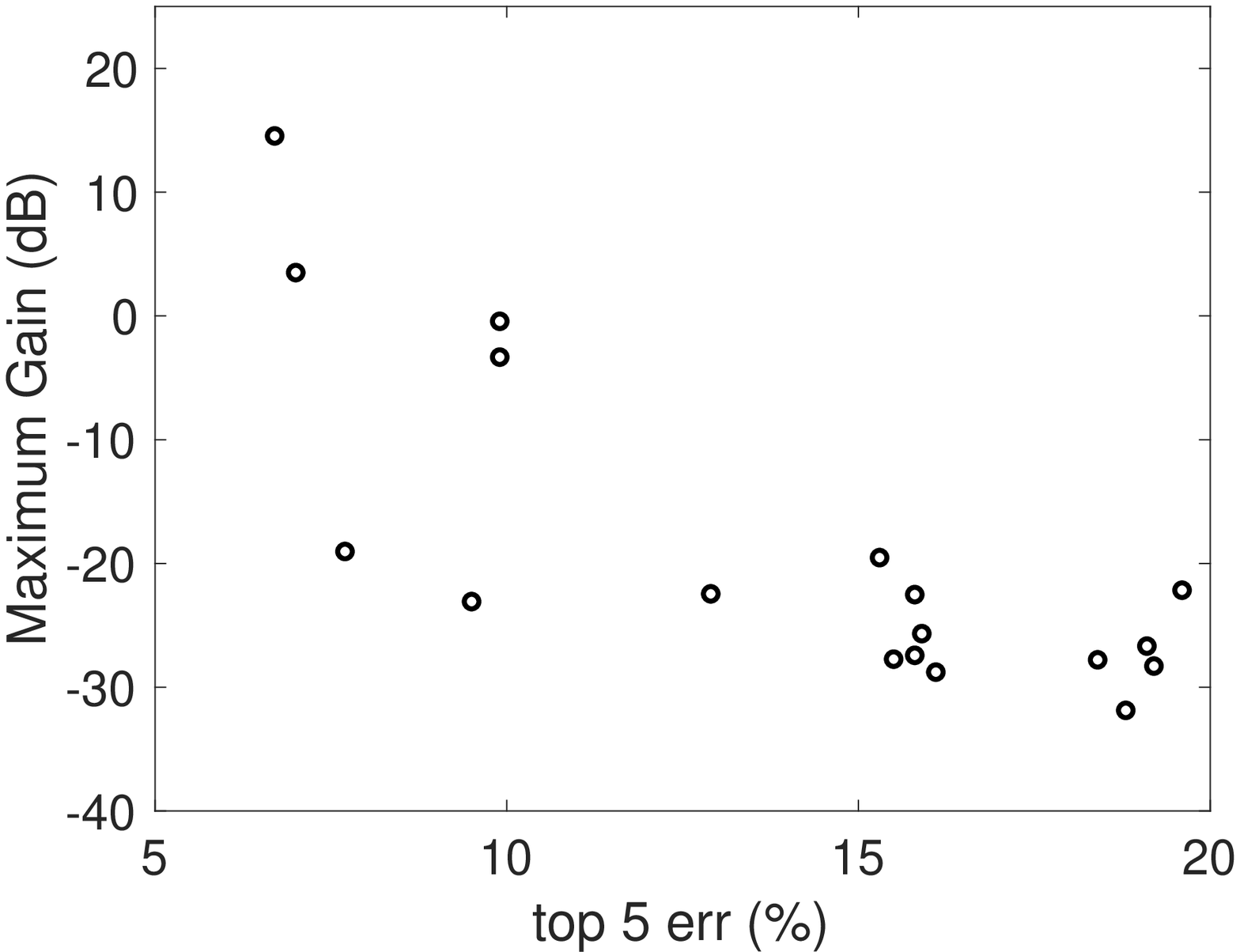}}
\caption{Maximum Gain for channel of an impulse image input to the ImageNet models of Figure~\ref{model_zoo} versus their top-5 error rate on the ILSVRC 2012 validation set.}
\label{max_gains}
\end{figure*}

\subsection{Results: pretrained models}

We used $18$ CNNs trained on ImageNet from MatConvNet\footnote{http://www.vlfeat.org/matconvnet/pretrained/} to compare their frequency response. Figure~\ref{model_zoo} shows the data derivative and frequency response obtained for the impulse image red channel and the top-5 error rate achieved by each model on the ImageNet Large Scale Visual Recognition Challenge (ILSVRC) 2012 validation data set. We observe that the best performing models are also the most sensitive to perturbations in the input so that their frequency response tends to have higher gain. Figure~\ref{max_gains} shows the Maximum Gain versus the error rate for each input channel in the impulse image. The best performing models tend to have maximum gains above $\sim -20$ dB whereas the worst performing models tend to have Maximum Gains below this. This trend persists if setting $p_i=\text{\emph{score}}$ because all models have comparable, low scores from their forward pass over the impulse image.

\section{Diagnose Training with Maximum Gain}

Given the previous results for pre-trained CNNs,
we examined the behavior the frequency response \emph{during} training in an effort to answer empirically the following questions:
\begin{itemize}
  \item Does the Maximum Gain increase as the CNN learns?
  \item Is Maximum Gain a useful quantity in deciding when the CNN has \emph{converged}?
\end{itemize}
To this end we trained CNNs on two public data sets using publicly available code\footnote{http://www.vlfeat.org/matconvnet/training/}
that performs Stochastic Gradient Descent (SGD) using backpropagation. We then measured the Maximum Gain from the impulse image induced frequency response
for snapshots of the same CNN saved after each training epoch.

We followed this approach across a set of experiments that examine the impact of learning rate on performance, as this is a known source of problems when not set appropriately. This enabled us to explore the potential for using Maximum Gain to detect such problems by comparing it to training and validation loss curves, which are normally used to probe the learning process.

\subsection{Training on MNIST}
\begin{figure*}
\centering
\subfloat[Maximum Gain]{
\includegraphics[trim={1cm 0cm 1cm 1cm},clip,width=0.61\columnwidth]{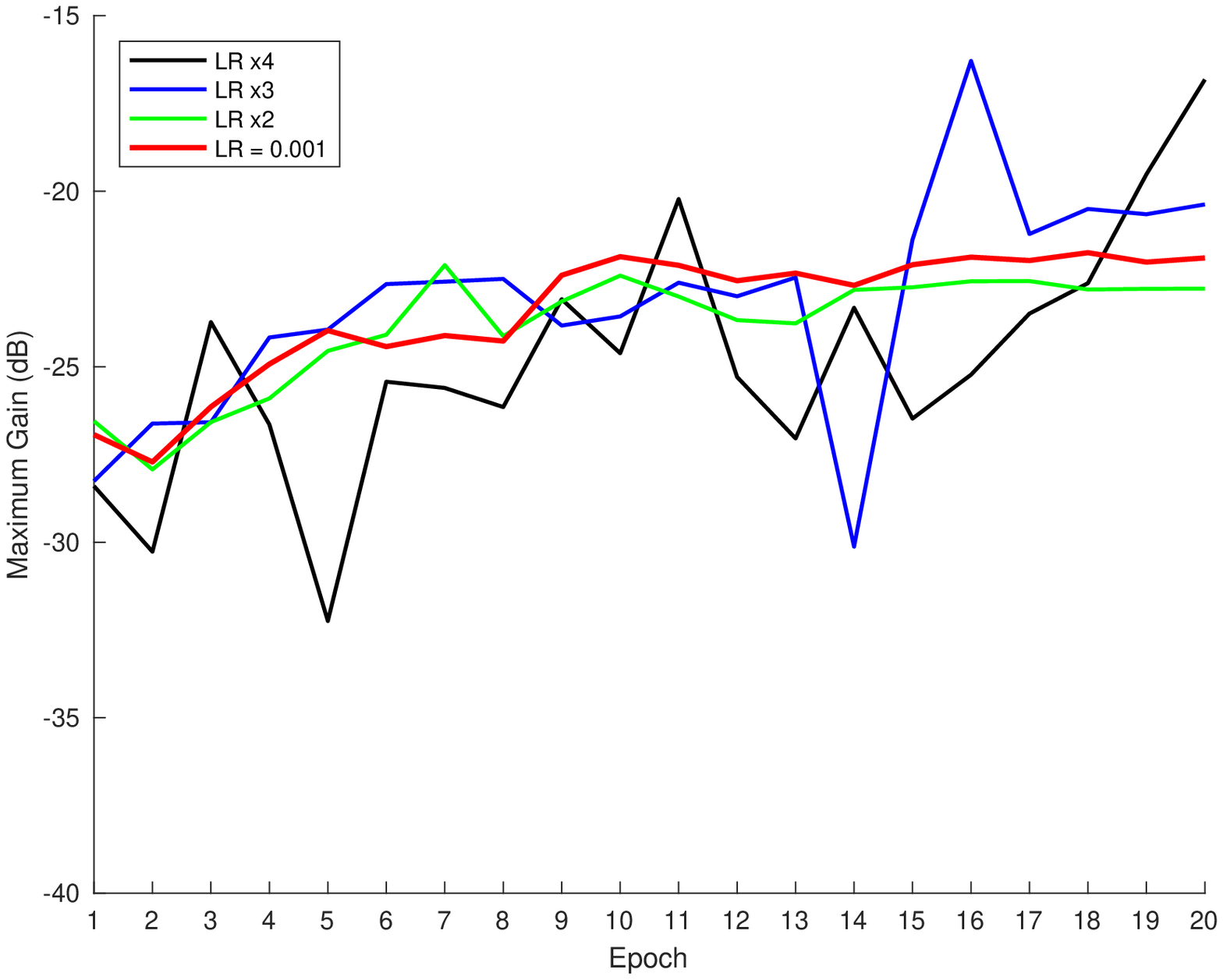}
}
\subfloat[Train \& Validation Loss]{
\includegraphics[trim={1cm 0cm 1cm 1cm},clip,width=0.61\columnwidth]{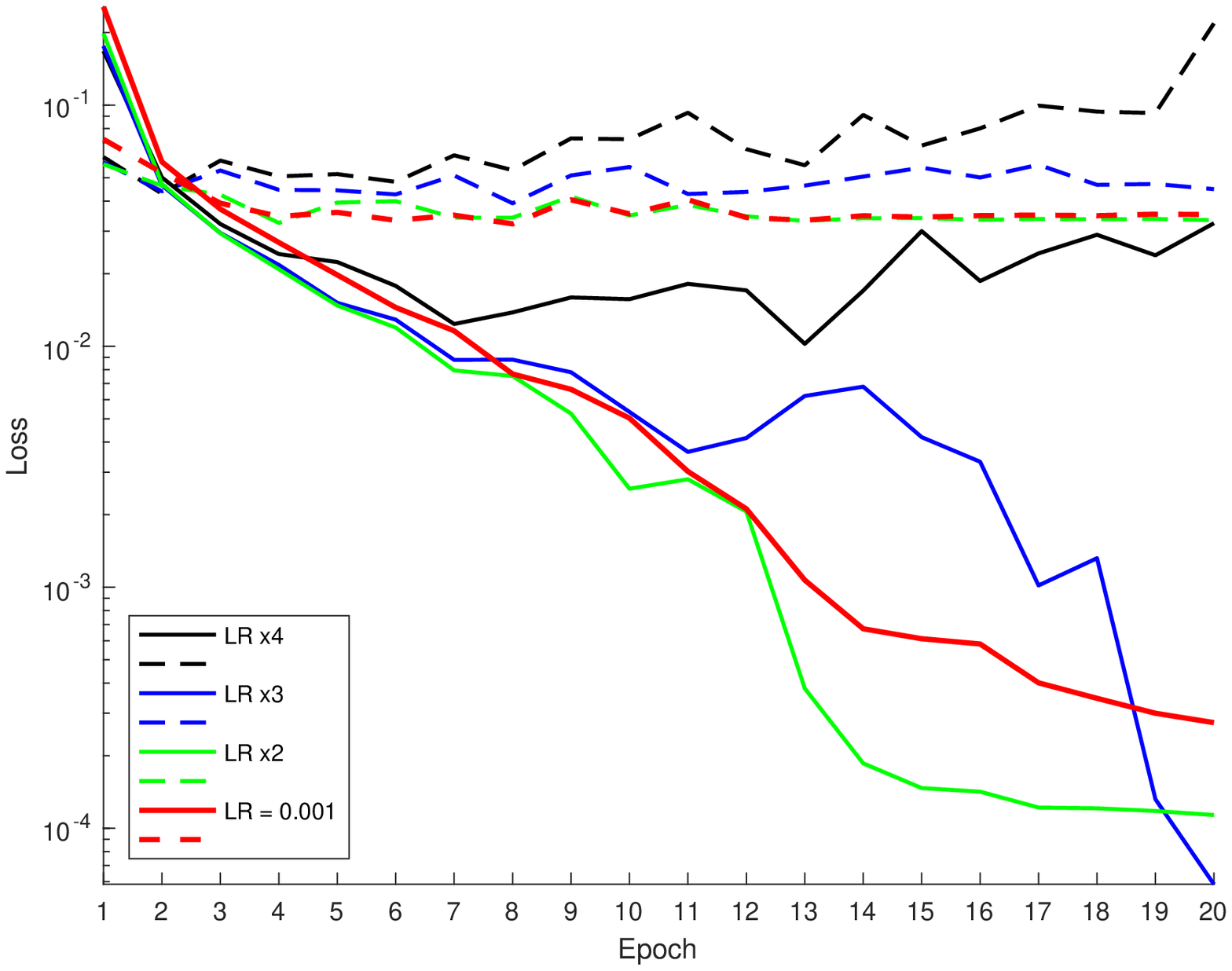}
}
\subfloat[Train \& Validation Error Rate]{
\includegraphics[trim={1cm 0cm 1cm 1cm},clip,width=0.61\columnwidth]{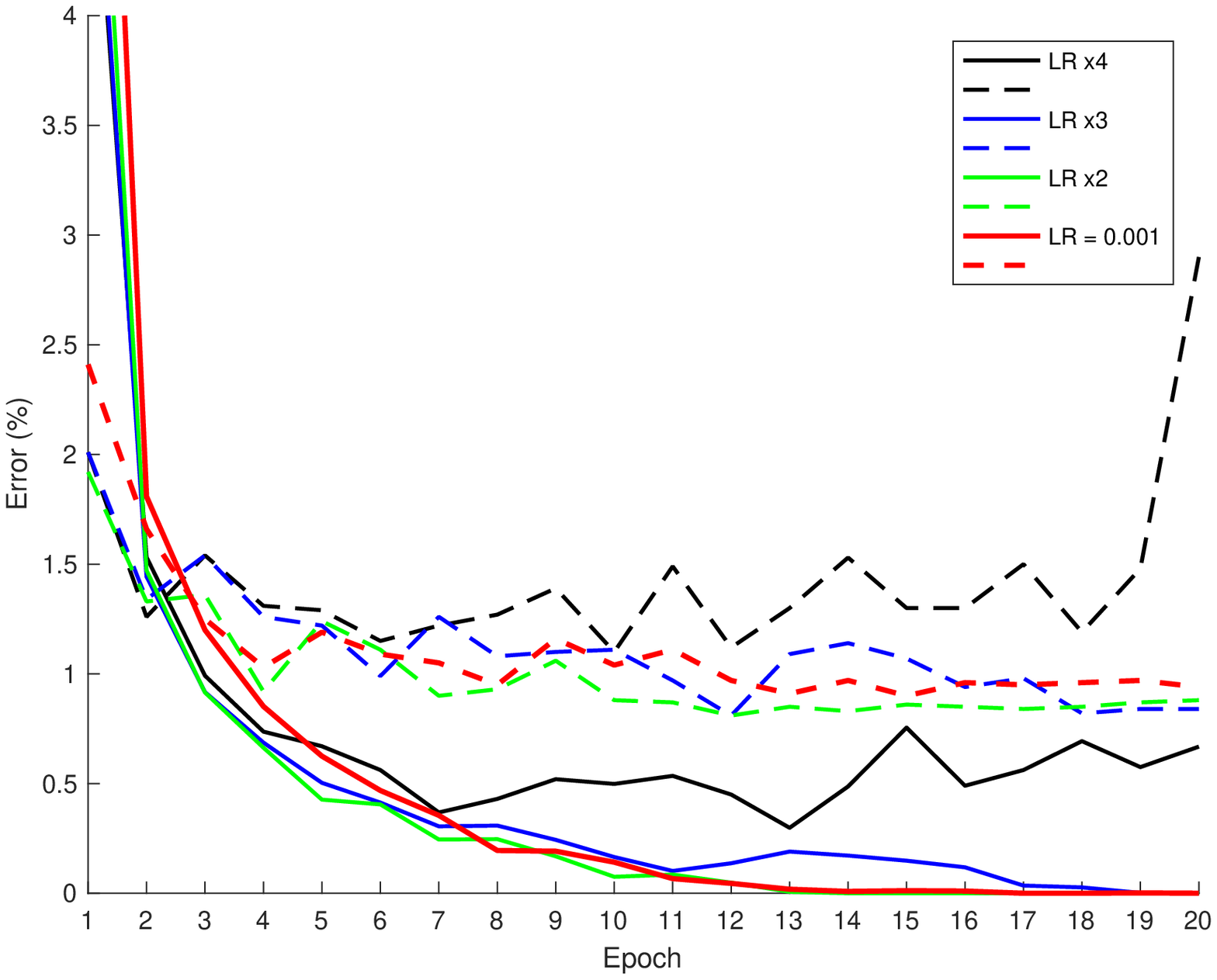}
}
\caption{Training LeNet on MNIST: separate training runs illustrate the effects of increasing a learning rate of 0.001 (red) by $\times2$ (green), $\times3$ (blue) or $\times4$ (black) on the maximum gain as well as on the training/validation (solid/dashed lines) losses and error rates.}
\label{mnist_lr_higher}
\end{figure*}

\begin{figure*}
\centering
\subfloat[Maximum Gain]{
\includegraphics[trim={1cm 0cm 1cm 1cm},clip,width=0.61\columnwidth]{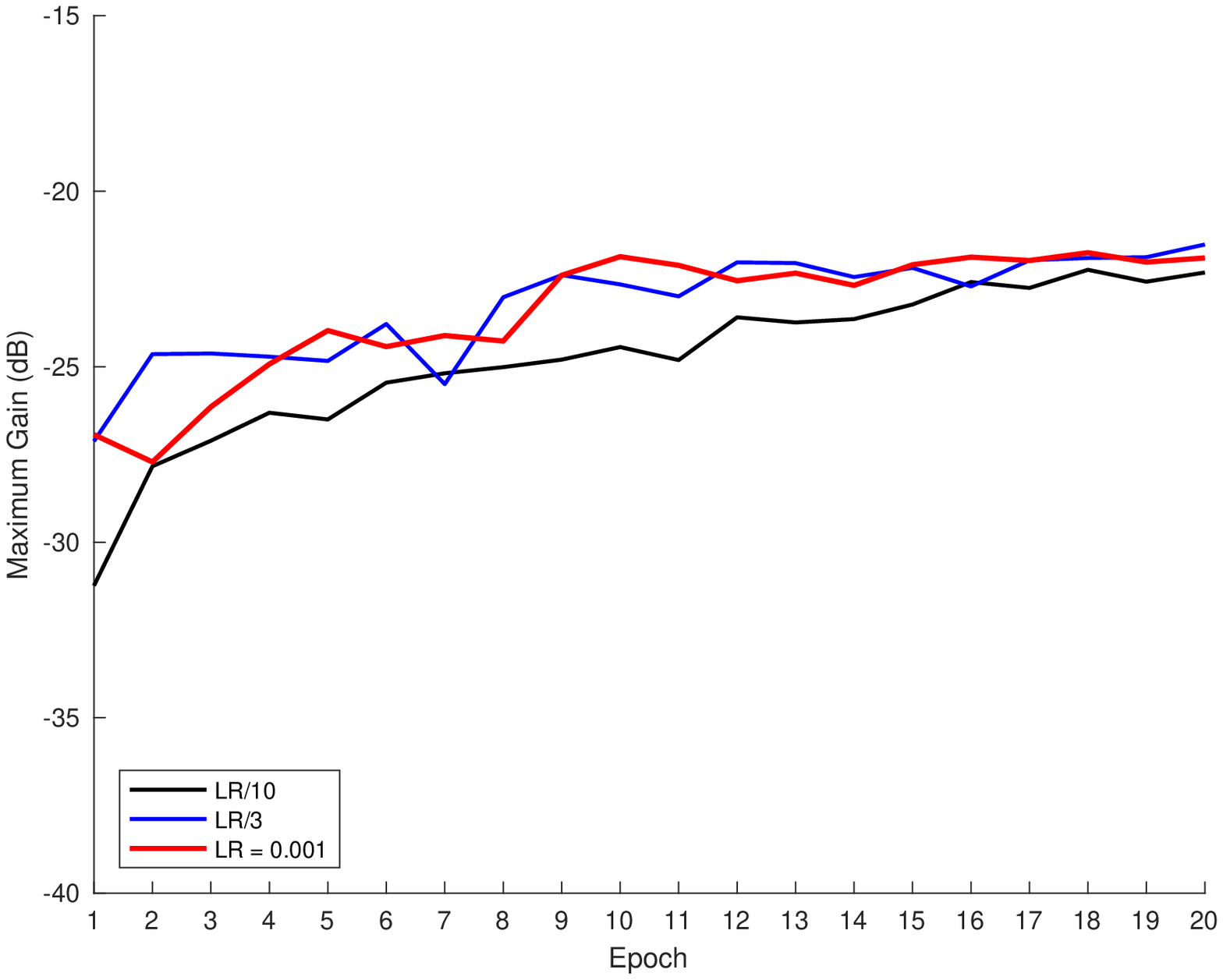}
}
\subfloat[Train \& Validation Loss]{
\includegraphics[trim={1cm 0cm 1cm 1cm},clip,width=0.61\columnwidth]{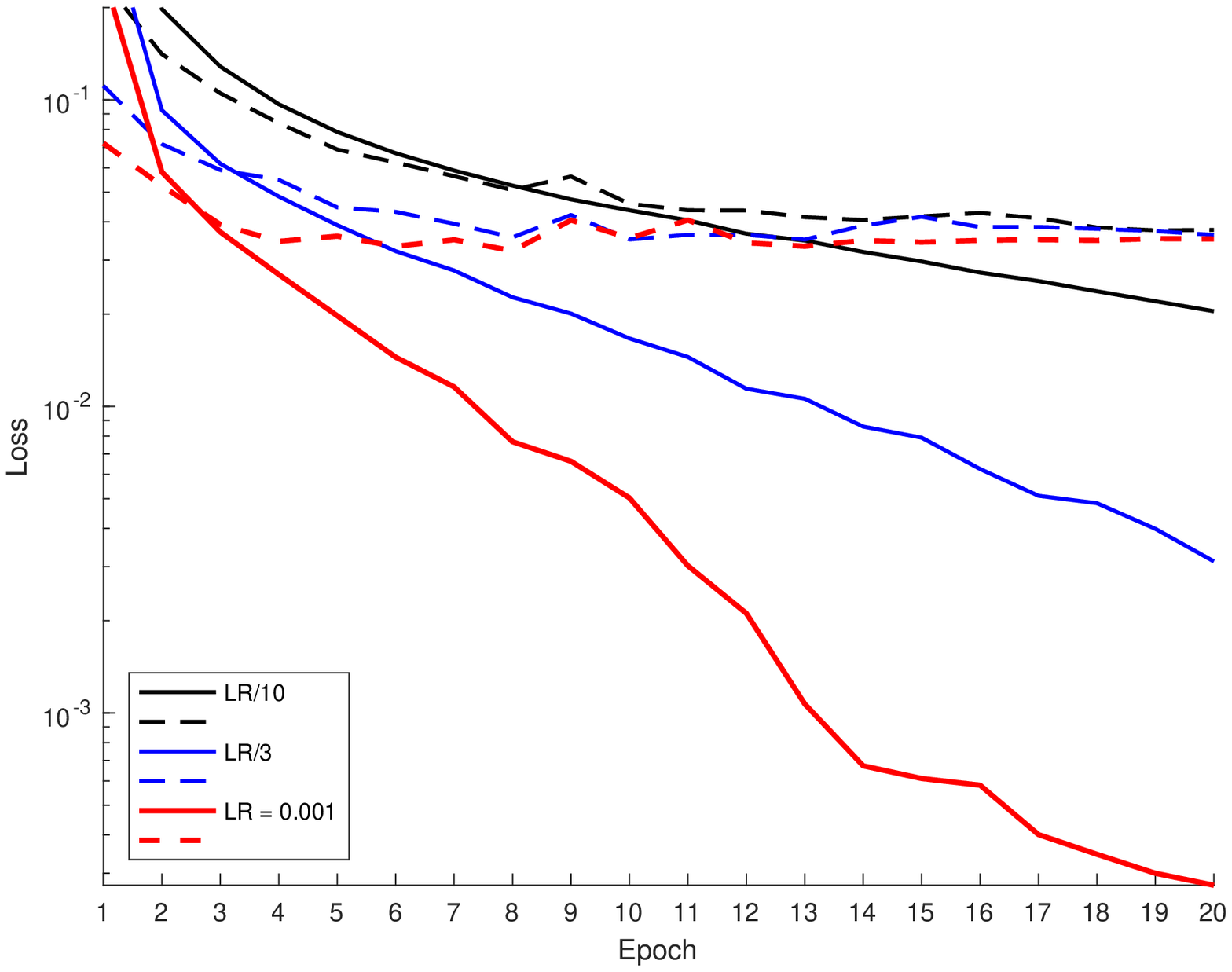}
}
\subfloat[Train \& Validation Error Rate]{
\includegraphics[trim={1cm 0cm 1cm 1cm},clip,width=0.61\columnwidth]{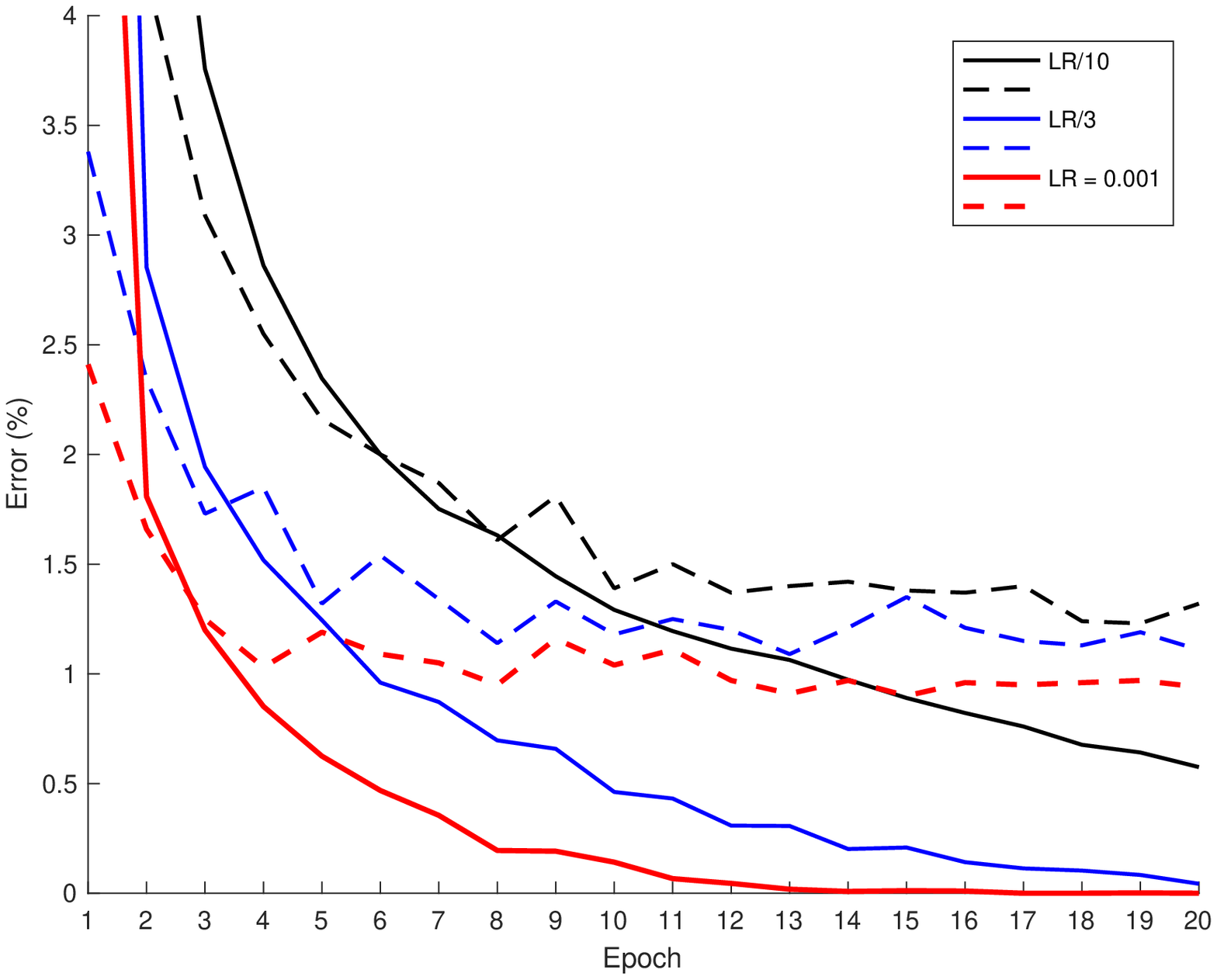}
}
\caption{Training LeNet on MNIST: separate training runs illustrate the effects of reducing a learning rate of 0.001 (red) by $\times\frac{1}{3}$ (blue) or $\times\frac{1}{10}$ (black) on the maximum gain as well as on the training/validation (solid/dashed lines) losses and error rates.}
\label{mnist_lr_lower}
\end{figure*}

We examined the behavior of Maximum Gain for an impulse image while training on the MNIST handwritten digit image data set~\cite{lecun1998mnist}, which contains $10$ classes and consists of $60,000$/$10,000$ training/validation grayscale images that are $28\times28$ pixels in size. For these experiments we chose to use the simple and well understood LeNet architecture~\cite{lecun98gradient}, which does not require a complex learning rate schedule to train properly. The CNN was made of three \emph{conv} layers and one \emph{fully-connected} layer, interspersed with non-linearities in the form of two \emph{max pool} layers and a \emph{relu} layer.

We began by training the CNN using a default learning rate of $0.001$ over $20$ epochs, the results of which are the red curves in Figures~\ref{mnist_lr_higher} and~\ref{mnist_lr_lower}. Here the Maximum Gain rises initially, then remains flat after $\sim10$ epochs. The corresponding training loss decreases over all epochs as the fit to the training data improves, while the validation loss decreases slowly up to epoch $\sim12$ after which it too becomes flat. The corresponding error rates have similar behavior. Here the Maximum Gain provides a similar diagnostic information as the validation loss.

In Figure~\ref{mnist_lr_higher} the learning rate was increased by $\times2$, $\times3$ and $\times4$, which progressively increased the variability of the Maximum Gain curves. In the case of the largest learning rate, training and validation losses begin to rise beyond epoch $7$, which indicates that the CNN is not learning, and the Maximum Gain curve no longer exhibits smoothly increasing or constant behavior, as was the case for a learning rate of $0.001$.

The learning rate was next reduced by factors $1/3$ and $1/10$ in Figure~\ref{mnist_lr_lower}. Maximum Gain is not as sensitive to these changes, however, for the lowest learning rate, it slowly rises over $20$ epochs. This indicates that the network would continue to learn if trained for longer. Given the fixed number of epochs, these Maximum Gain curves indicate that the two higher learning rates are more appropriate. This is confirmed by the validation errors, which provide the additional insight that after $20$ epochs the learning rate of $0.001$ performs better than the learning rate of $0.001/3$.

\subsection{Training on CIFAR-10}
\begin{figure*}
\centering
\subfloat[Maximum Gain]{
\includegraphics[trim={1cm 0cm 1cm 1cm},clip,width=0.6\columnwidth]{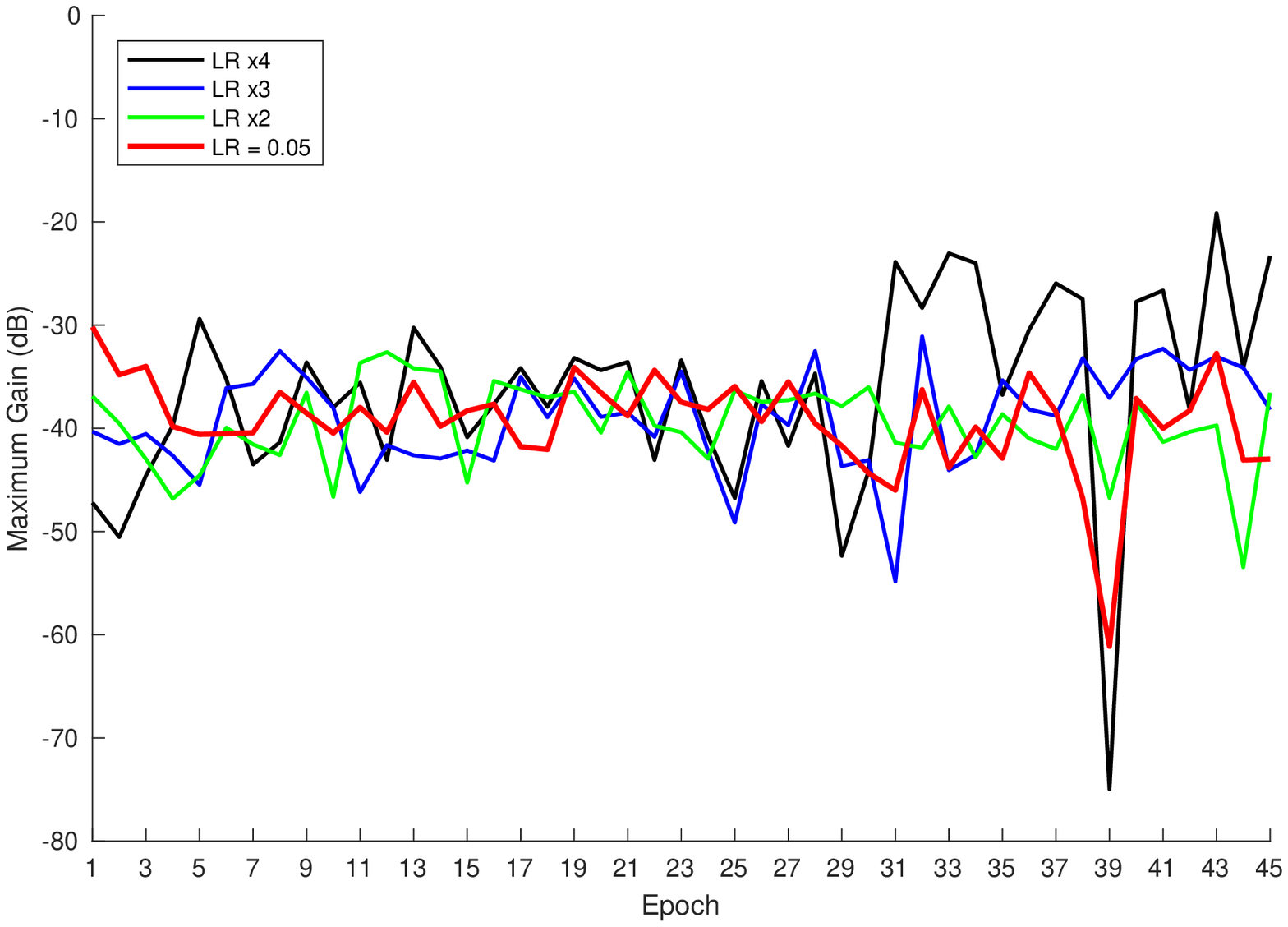}
}
\subfloat[Train \& Validation Loss]{
\includegraphics[trim={1cm 0cm 1cm 1cm},clip,width=0.6\columnwidth]{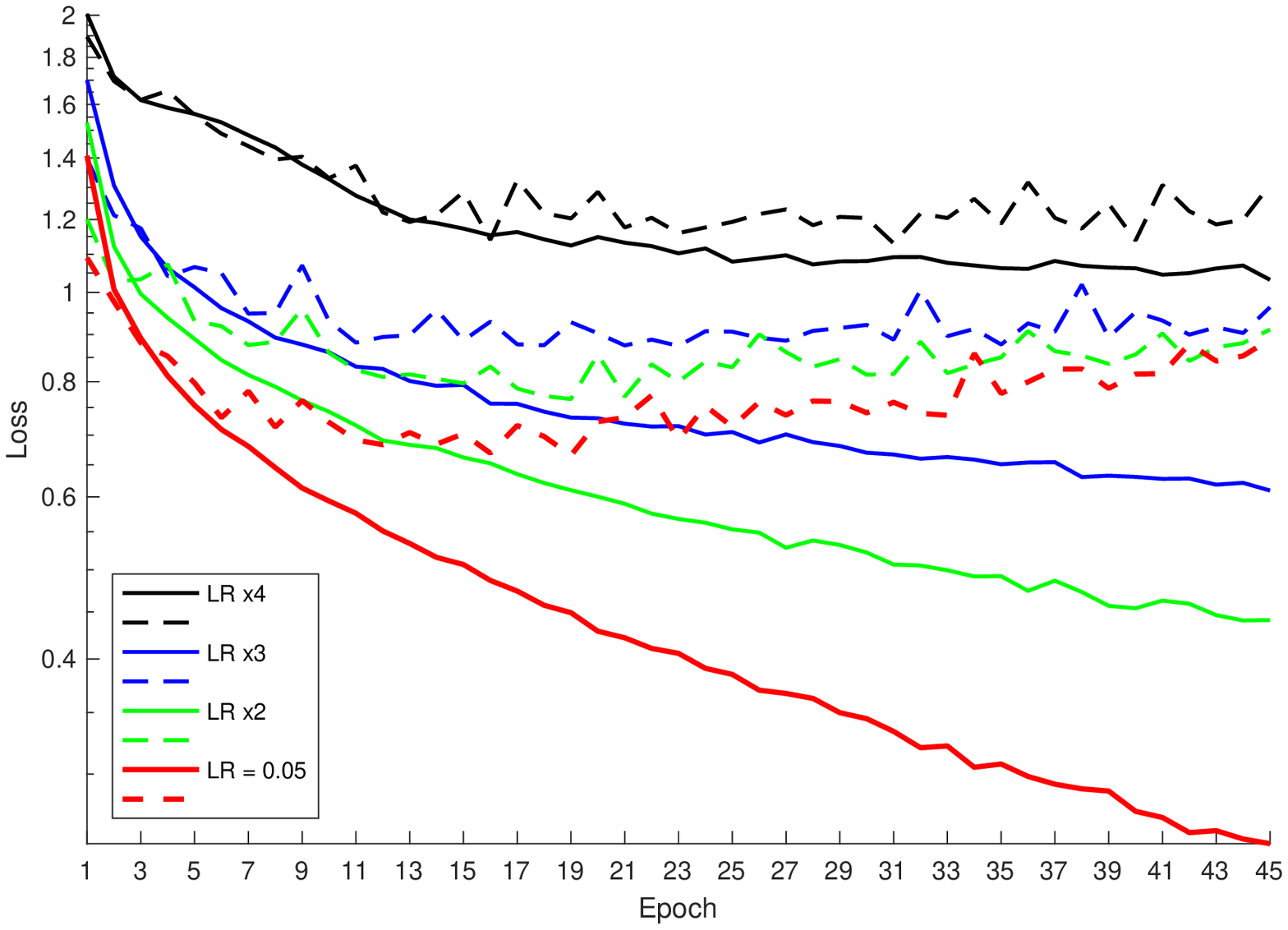}
}
\subfloat[Train \& Validation Error Rate]{
\includegraphics[trim={1cm 0cm 1cm 1cm},clip,width=0.6\columnwidth]{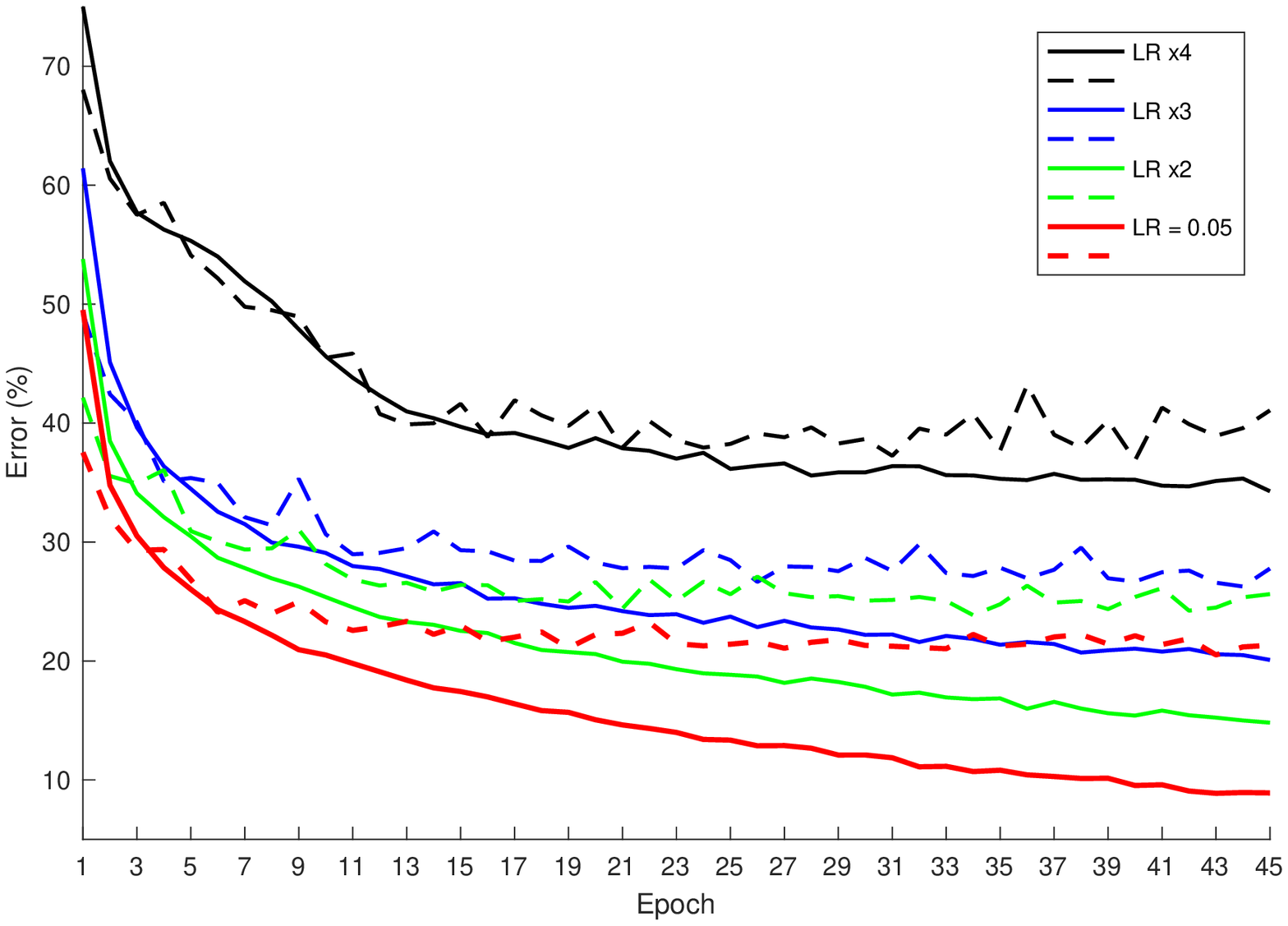}
}
\caption{Training LeNet on CIFAR-10: separate training runs illustrate the effects of increasing a learning rate of 0.05 (red) by $\times2$ (green), $\times3$ (blue) or $\times4$ (black) on the maximum gain as well as on the training/validation (solid/dashed lines) losses and error rates.}
\label{cifar10_lr_higher}
\end{figure*}

\begin{figure*}
\centering
\subfloat[Maximum Gain]{
\includegraphics[trim={1cm 0cm 1cm 1cm},clip,width=0.6\columnwidth]{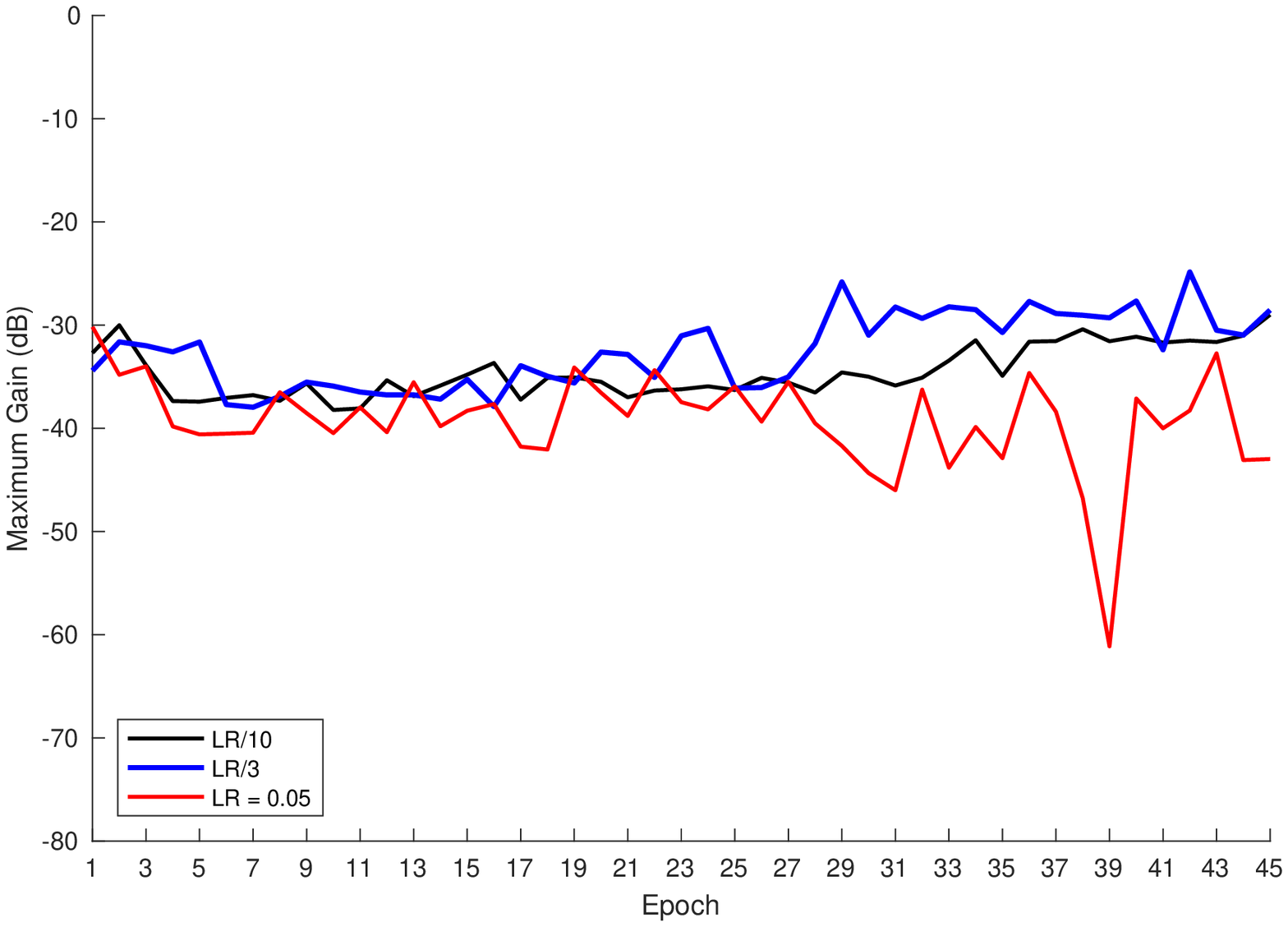}
}
\subfloat[Train \& Validation Loss]{
\includegraphics[trim={1cm 0cm 1cm 1cm},clip,width=0.6\columnwidth]{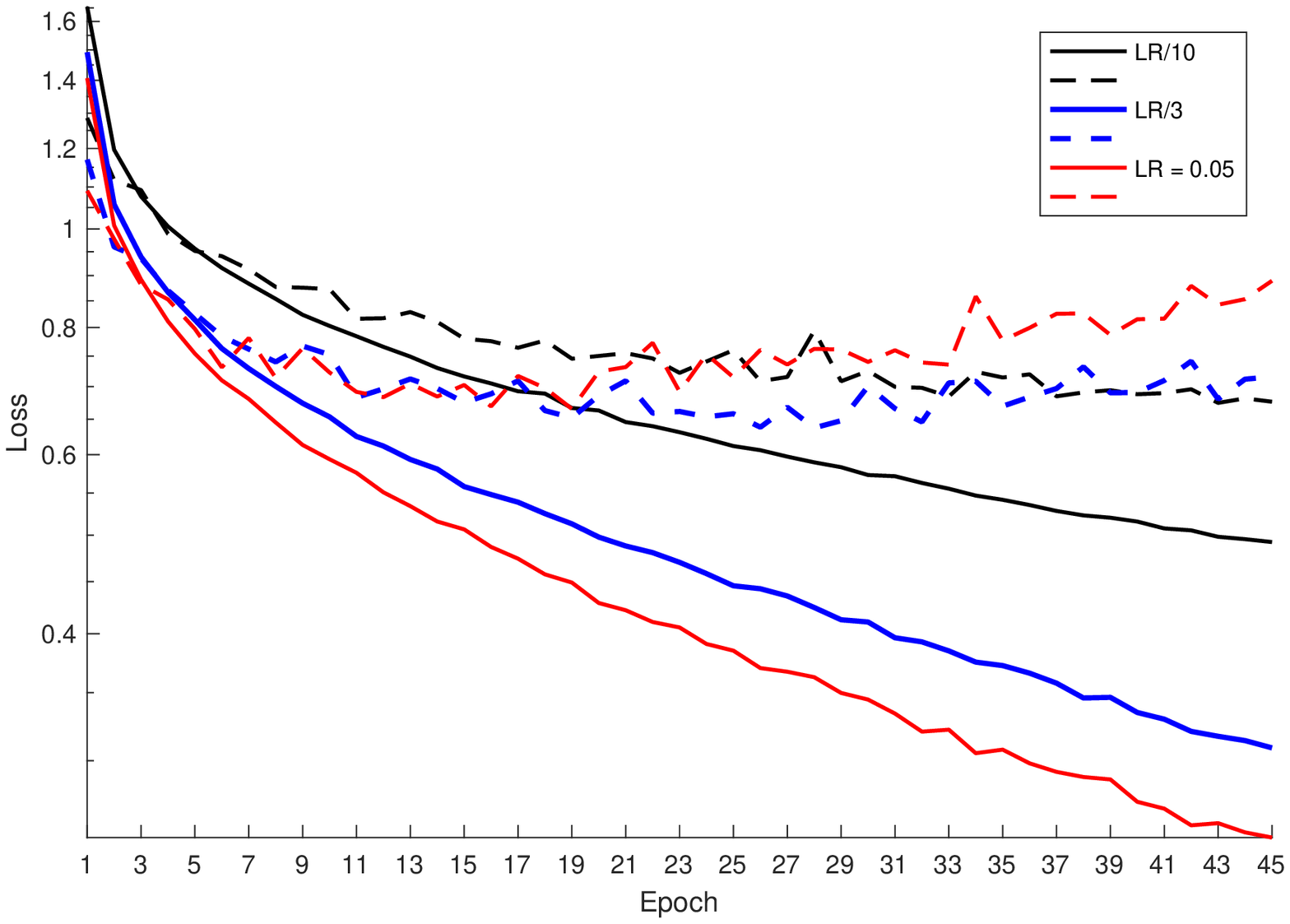}
}
\subfloat[Train \& Validation Error Rate]{
\includegraphics[trim={1cm 0cm 1cm 1cm},clip,width=0.6\columnwidth]{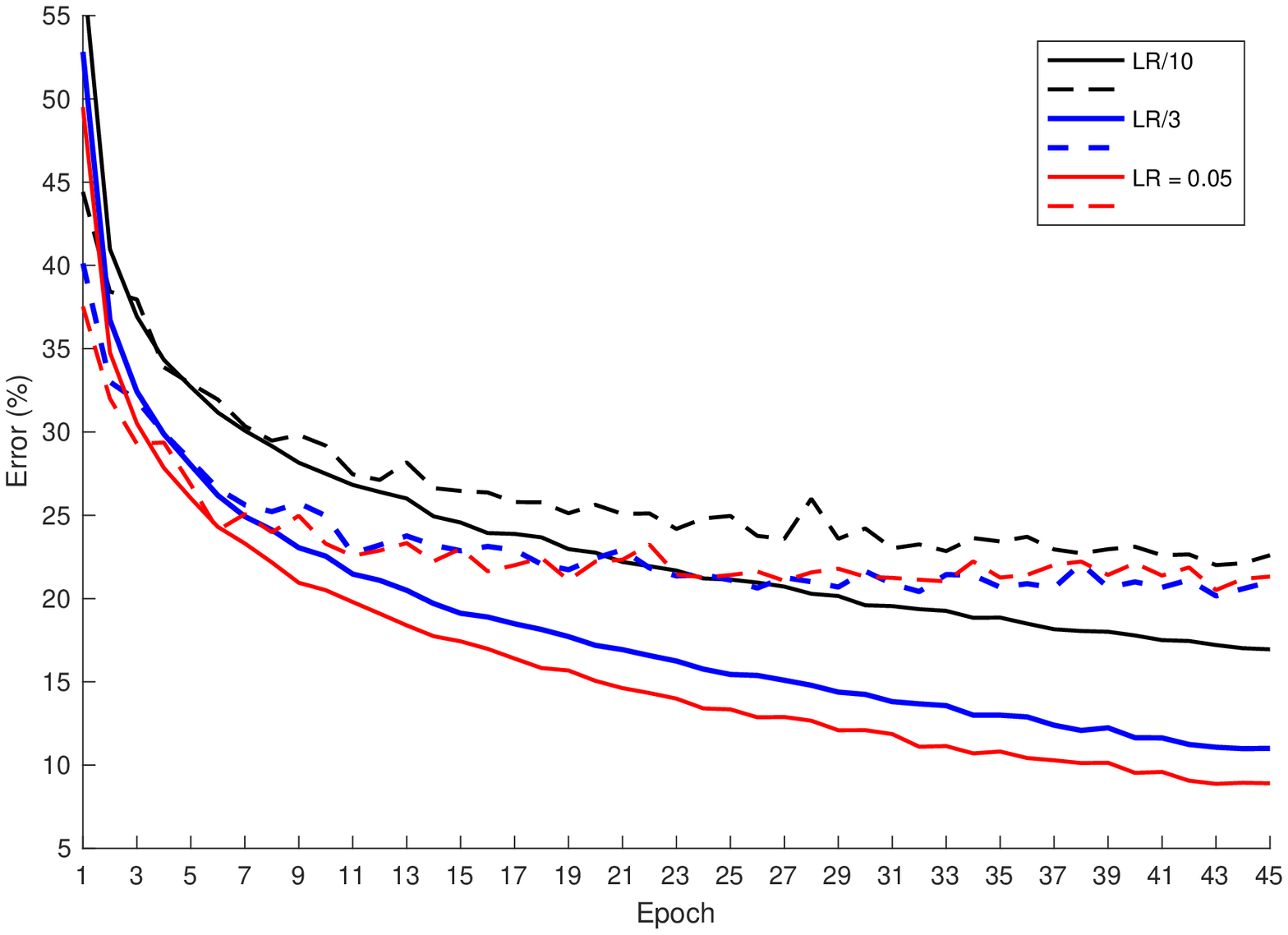}
}
\caption{Training LeNet on CIFAR-10: separate training runs illustrate the effects of reducing a learning rate of 0.05 (red) by $\times\frac{1}{3}$ (blue) or $\times\frac{1}{10}$ (black) on the maximum gain as well as on the training (solid lines) and validation (dashed lines) losses and error rates.}
\label{cifar10_lr_lower}
\end{figure*}

We next measured the Maximum Gain during training on the CIFAR-10 image data set~\cite{krizhevsky2009learning}, which contains $10$ classes and consists of $50,000$/$10,000$ training/validation color images that are $32\times32$ pixels in size. The Maximum Gain obtained for each input color channel was averaged in order to only have one value per model.
As this is a more difficult data set than MNIST, a larger LeNet architecture was deployed, with four \emph{conv} layers, two \emph{average pool} layers and one \emph{fully-connected} layer together with two \emph{max pool} layers and four \emph{relu} layers. Results from a second architecture~\cite{lin2013network} are compared in the discussion below (\ref{nin_resuls}).

The CNN was initially trained with a learning rate of $0.05$ over $45$ epochs
and this corresponds to the red curves in Figures~\ref{cifar10_lr_higher} and~\ref{cifar10_lr_lower}. The Maximum Gain drops at first, then remains steady up to epoch $\sim 27$, where it becomes more variable.
This corresponds to a rise in validation loss coupled with decreasing training loss
and shows that the Maximum Gain curve is able to detect the problem of strong overfitting. 

Figure~\ref{cifar10_lr_higher} shows that increasing the learning rate
leads to additional variability in the Maximum Gain curve at all epochs.
This corresponds to higher training and validation losses and error rates
due to a degradation of the learning process.
The effect of reducing rate by factors $1/3$ and $1/10$ is shown in Figure~\ref{cifar10_lr_lower}.
The Maximum Gain curves become smooth and increase between epochs $\sim7$ and $45$.
In particular, the black curve for the lowest learning rate of $0.005$ suggests that the CNN has not yet converged.
The blue curve, on the other hand, which corresponds to a learning rate of $0.05/3$, also has higher variability at later epochs where the model overfits.

\subsection{Discussion}

\subsubsection{Using Maximum Gain to select the learning rate and detect overfitting}

Based on the preceding experiments, we summarize the key aspects of the learning process that Maximum Gain curves can diagnose:
\begin{itemize}
  \item High variability across all epochs indicates that the learning rate is set too high.
  \item Smooth gradual rise across all epochs indicates that the learning rate is too low or that training for longer would noticeably improve performance.
  \item A rise followed by a smooth flattening suggests that the model has begun to converge.
  \item Smooth, flat behavior followed by high variability indicates when strong overfitting occurs
  and that it is time to stop training.
\end{itemize}
A caveat of this is that one must determine empirically what constitutes `high variability' for a particular CNN architecture and data set by trialing a few learning rates.
In terms of detecting model overfitting, however, what is `high' or `low' variability can be determined from the same curve.

\subsubsection{Impact of model architecture}\label{nin_resuls}

\begin{figure*}
\centering
\subfloat[Maximum Gain]{
\includegraphics[trim={1cm 0cm 1cm 1cm},clip,width=0.6\columnwidth]{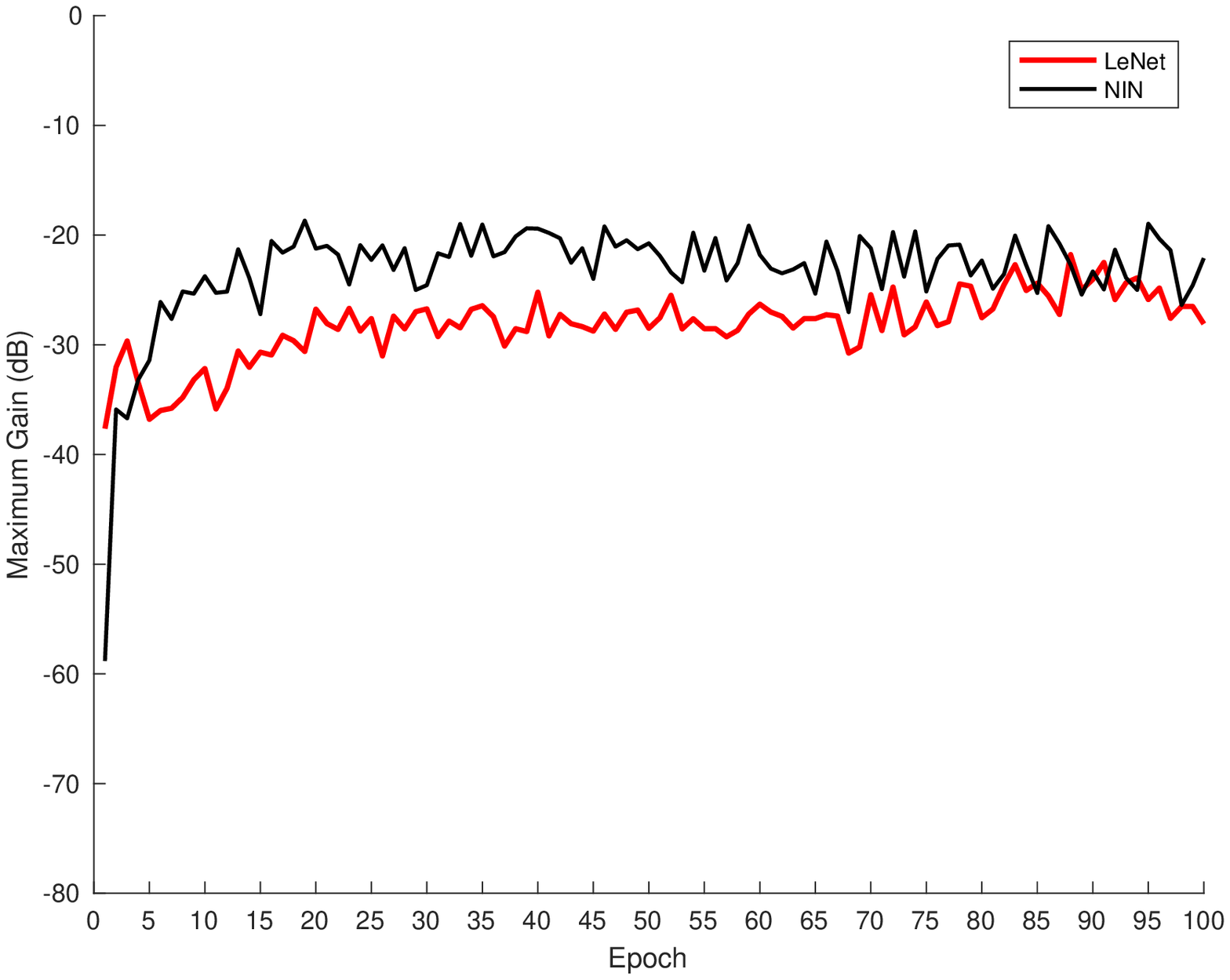}
}
\subfloat[Train \& Validation Loss]{
\includegraphics[trim={1cm 0cm 1cm 1cm},clip,width=0.6\columnwidth]{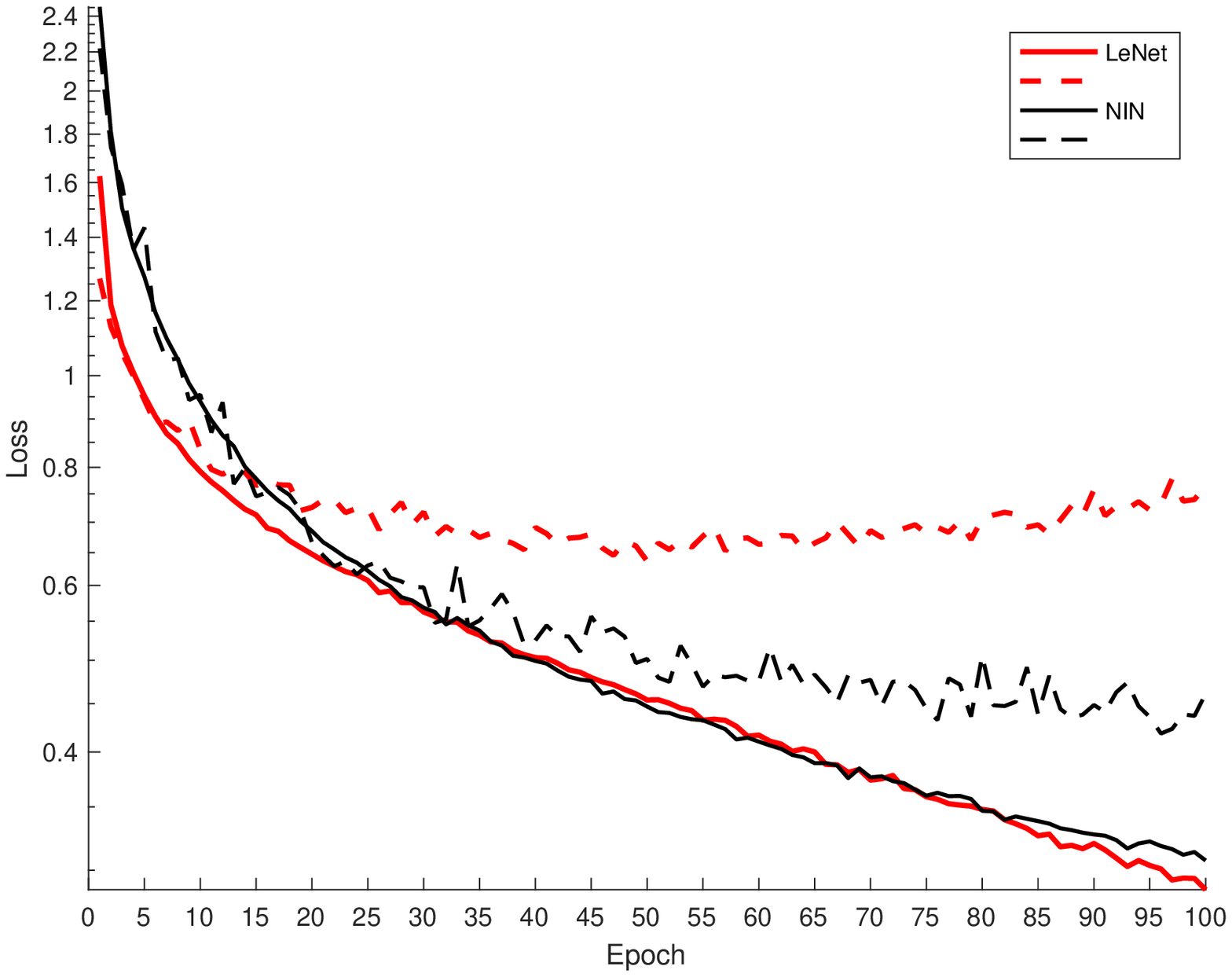}
}
\subfloat[Train \& Validation Error Rate]{
\includegraphics[trim={1cm 0cm 1cm 1cm},clip,width=0.6\columnwidth]{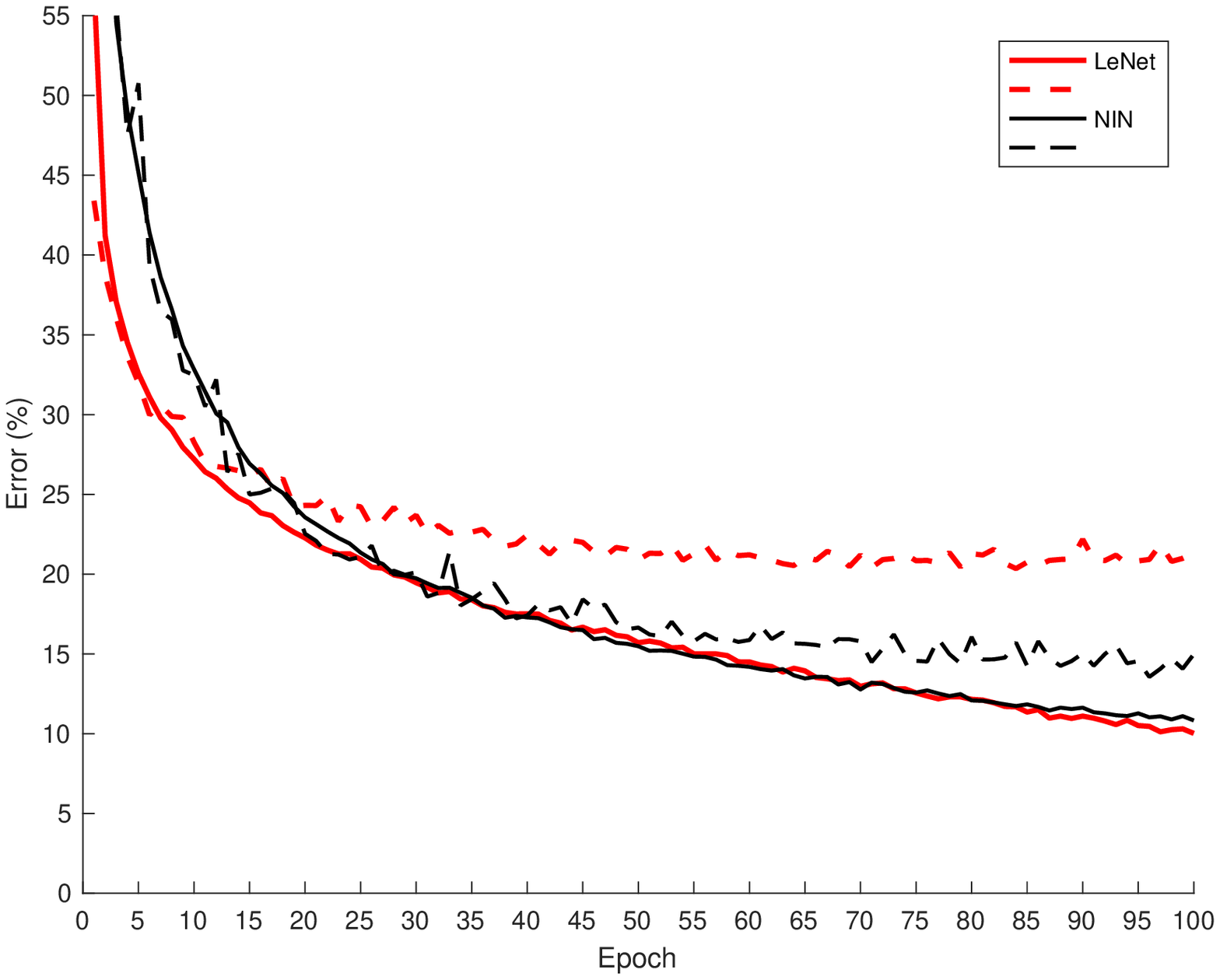}
}
\caption{Training on CIFAR-10 with learning rate of 0.005: comparison of the maximum gain as well as the training (solid lines) and validation (dashed lines) losses and error rates for the LeNet (red) or NIN (black) architectures.}
\label{cifar10_nin}
\end{figure*}

\begin{figure*}
\centering
\subfloat[Maximum Gain]{
\includegraphics[trim={1cm 0cm 1cm 1cm},clip,width=0.6\columnwidth]{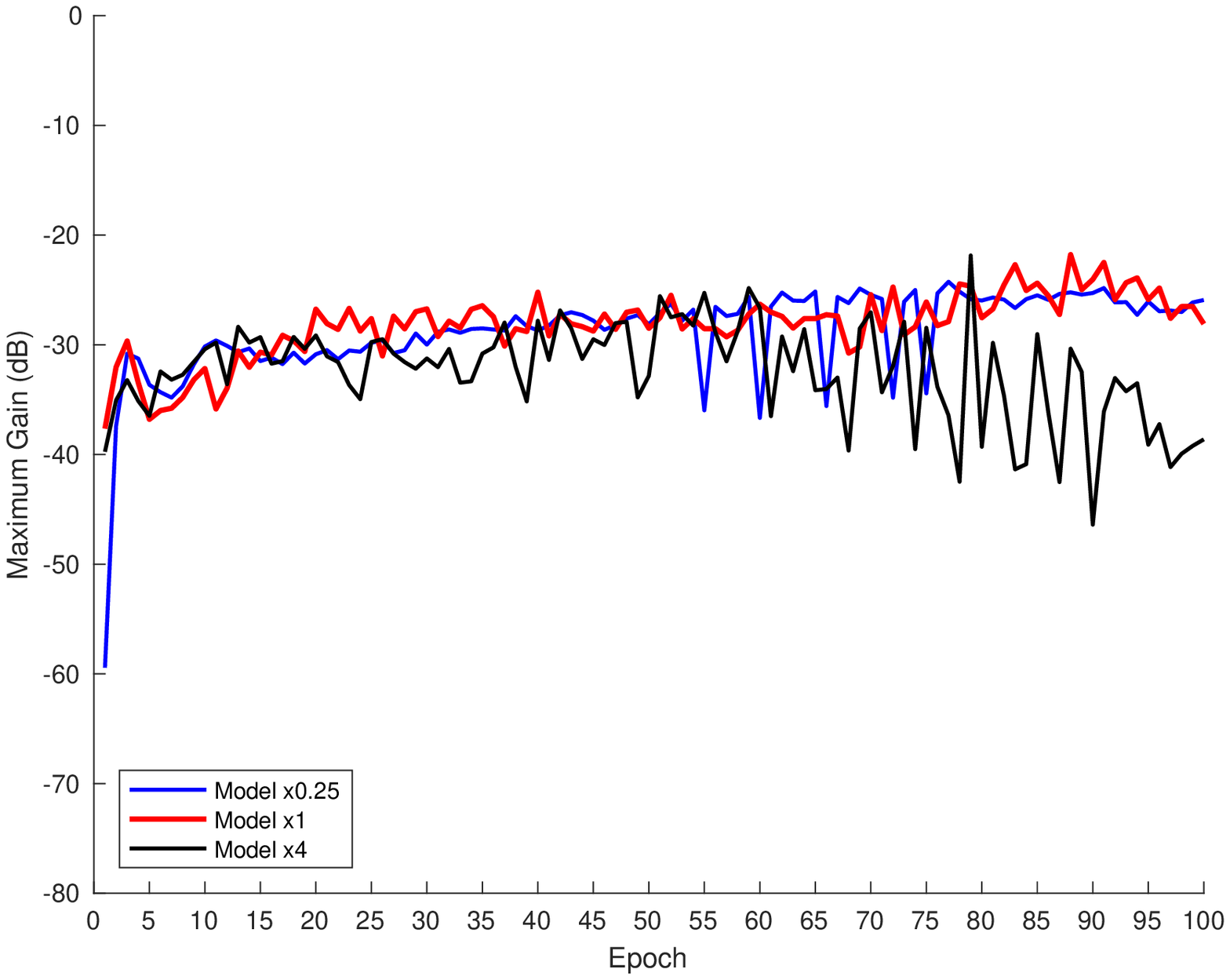}
}
\subfloat[Train \& Validation Loss]{
\includegraphics[trim={1cm 0cm 1cm 1cm},clip,width=0.6\columnwidth]{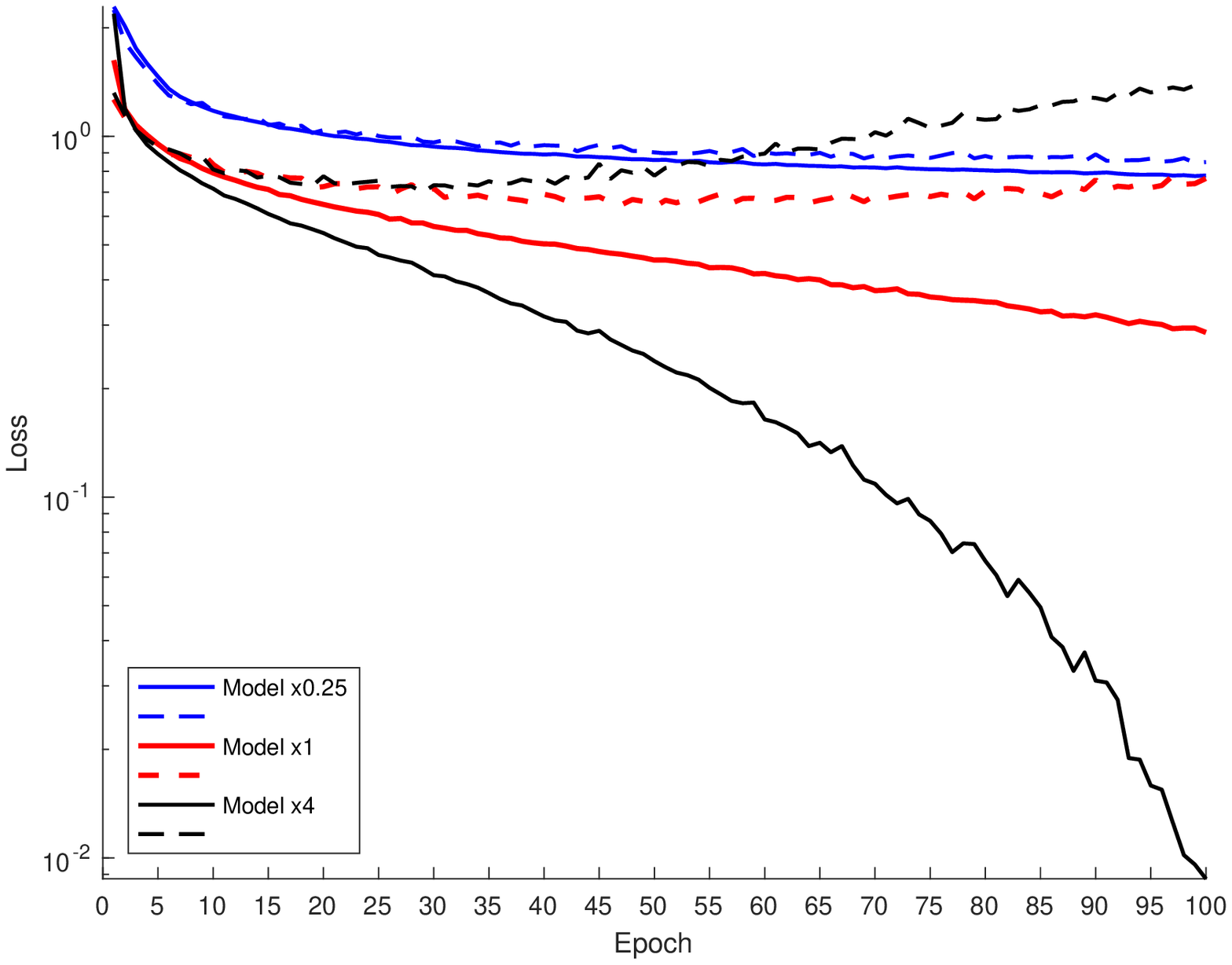}
}
\subfloat[Train \& Validation Error Rate]{
\includegraphics[trim={1cm 0cm 1cm 1cm},clip,width=0.6\columnwidth]{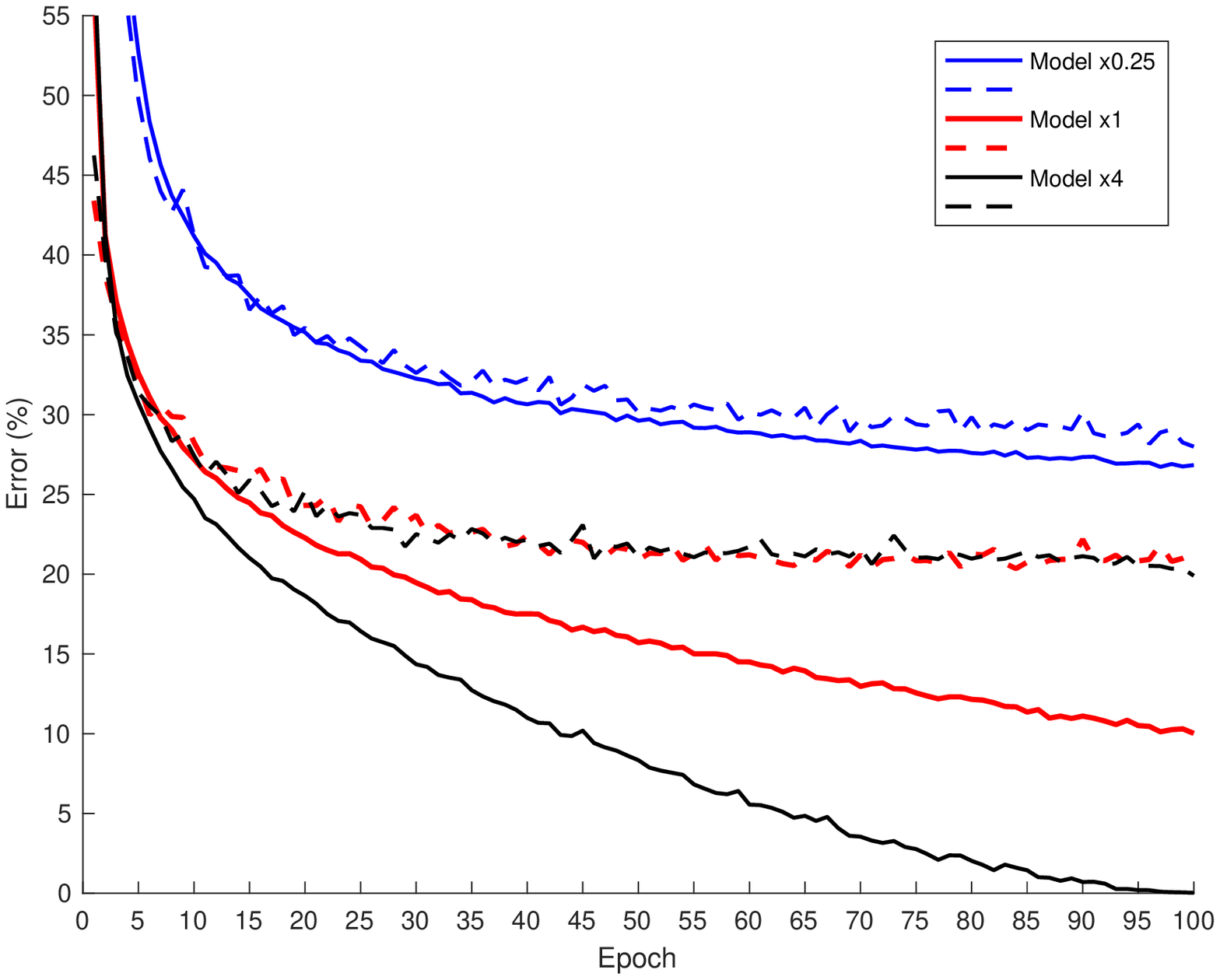}
}
\caption{Training LeNet on CIFAR-10 with learning rate of 0.005: scaling the original number of filters (red) in \emph{conv} layers by $\times 4$ (black) and by $\times 1/4$ (blue). Training/validation losses and error rates are shown as solid/dashed lines.}
\label{cifar10_modelscale}
\end{figure*}

We trained a Network in Network (NIN)~\cite{lin2013network} on CIFAR-10 using a learning rate of $0.005$ for $100$ epochs. Figure~\ref{cifar10_nin} shows that this achieves lower validation error than LeNet, which experiences strong overfitting beyond epoch $70$. The LeNet Maximum Gain curve is flat up to this point,
after which it becomes more variable and comparable to that of NIN, which is flat. This demonstrates that absolute value of Maximum Gain is not a reliable indicator of performance.

The impact of model capacity in shown in Figure~\ref{cifar10_modelscale}, where we reduced and increased the number of filters in each LeNet \emph{conv} layer by $\times 1/4$ and $\times4$, respectively. The smaller model underfits, but its Maximum Gain curve is similar to the original, albeit with some additional variability. Hence, Maximum Gain does not detect underfitting, however, this is already highlighted by the high training error. The larger model overfits from epoch $15$ and this is reflected in its Maximum Gain curve,
which becomes increasingly variable.

\section{Conclusion}

In this work we have defined the transfer function of a CNN by using its input data derivative,
which is obtained for a given input image via backpropagation. We characterized the resulting frequency response in terms of its peak value, which we called the Maximum Gain. Comparing CNNs with different architectures that were trained for image classification on ImageNet, we found that better performing models tend to have higher Maximum Gain for an impulse image. Intrigued by this observation, we examined the behavior Maximum Gain during training and found that, when the learning process works well, it rises with training epoch and then saturates.  

Experiments on MNIST and CIFAR-10 showed that Maximum Gain can provide similar information about the learning process as the validation loss. We propose that Maximum Gain could serve as a stand-in replacement for validation loss, which would be useful in situations where validation data are not available. This can happen in practice, for example during a final training stage in which the validation set has been combined with the training set to have more training data. Furthermore, the method is fast because it operates on a single impulse image. As caveats, we note that while Maximum Gain does detect overfitting, it does not reliably detect underfitting, nor the degree of overfitting, which is typically gauged from the gap between validation and training losses or errors. Moreover, it does not provide an absolute measure of performance on new data, which is provided by the validation error. Future work will explore further the impact of CNN architecture on Maximum Gain, as well as its application to new data sets.


\section*{Acknowledgment}
Discussions with Dr Sebastien Wong of University of South Australia and Dr Hugh Kennedy of DST Group are gratefully acknowledged.



%

\bibliographystyle{IEEEtran}
\bibliography{impulse}


\end{document}